\documentclass[10pt,twocolumn,letterpaper]{article}

\usepackage{wacv}     % To produce the REVIEW version for the datasets track
\usepackage{booktabs}    % Professional-quality horizontal rules.
\usepackage{array}       % Extends the tabular environment with custom column
\usepackage{colortbl}    % Enables row, column, and cell coloring via
\usepackage{multirow}    % Allows table cells to span multiple rows using
\usepackage{longtable}   % Supports tables that automatically span multiple pages.

\usepackage{diagbox}     % Creates diagonal table header cells for row/column labels.

\definecolor{tealx}{HTML}{0E7C7B}

\definecolor{rosex}{HTML}{9D174D}

\definecolor{soft}{HTML}{000000}

\definecolor{rule}{HTML}{C9D2EA}

\definecolor{ink}{HTML}{1A2238}        % near-black navy for text
\definecolor{accent}{HTML}{2D5BFF}     % electric blue
\definecolor{accentdk}{HTML}{1E3A8A}   % deep blue
\definecolor{amber}{HTML}{B45309}      % amber for reviewer-lens
\definecolor{paper}{HTML}{FBFBFD}

\arrayrulecolor{rule}

\usepackage{tikz}
\usepackage{pgfplots}
\pgfplotsset{compat=1.18}

\usetikzlibrary{calc}

\usepackage{subcaption}

\usepackage{booktabs}      % \toprule, \midrule, \bottomrules
\usepackage{tabularx}      % tabularx environment
\usepackage{array}         % custom column formatting
\usepackage[table]{xcolor} % \rowcolor
\usepackage{graphicx}      % \includegraphics

\definecolor{citegreen}{RGB}{0,128,80}

\definecolor{catgray}{RGB}{235,235,235}

\usepackage{microtype}

\definecolor{wacvblue}{rgb}{0.21,0.49,0.74}
\definecolor{citegreen}{RGB}{0,128,80}

\usepackage{hyperref}

\hypersetup{
    colorlinks=true,
    breaklinks=true,
    citecolor=citegreen,
    linkcolor=citegreen,
    urlcolor=blue,
    pdfborder={0 0 0}
}

\usepackage{xr-hyper}
\def\wacvPaperID{2518} % *** Enter the WACV Paper ID here
\def\confName{WACV}
\def\confYear{2027}

\title{
\LARGE\bfseries
\raisebox{-0.10cm}{\includegraphics[height=1.2cm]{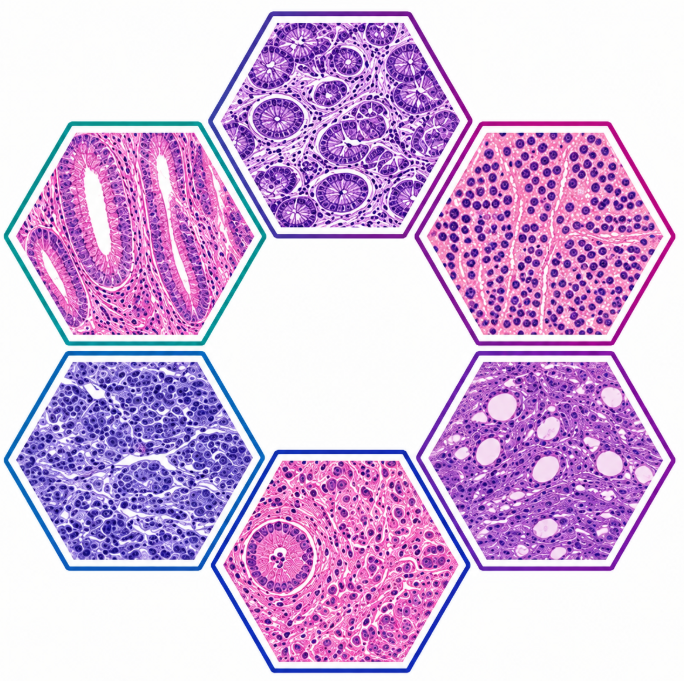}}\hspace{0.4em}
MOSAIC: Towards a Universal Latent \\ Space for Computational Pathology Foundation Models
}

\author{
Dibaloke Chanda \qquad Praveen Madiraju\\
Department of Computer Science\\
Marquette University\\
Milwaukee, WI, USA
}

\begin{document}
\maketitle

\begin{abstract}
Pathology foundation models (PFMs) differ substantially in architecture, pretraining objective, training data and representation dimensionality, yet it remains unclear whether they learn fundamentally different feature spaces or alternative coordinate systems of a shared underlying representation. We present \textbf{MOSAIC}, a systematic study of latent-space similarity, alignment and interoperability across $12$ patch-level and $6$ slide-level PFMs. We first quantify representational agreement using complementary similarity measures and analyze how architecture, pretraining objective, network depth and magnification influence latent geometry. We then evaluate multiple alignment strategies for mapping independently learned representations into a common latent space. The resulting representations are evaluated through cross-model agreement, retrieval, geometry preservation, and downstream multiple-instance learning across molecular and morphological prediction tasks. MOSAIC provides a unified framework for studying whether heterogeneous PFMs can move beyond representational similarity toward a shared and functionally transferable pathology representation space.  Our code, benchmarks and additional results are available  (anonymously) at
\href{https://researchsubmissions66.github.io/MOSAIC/}
{https://researchsubmissions66.github.io/MOSAIC/}.
\end{abstract}
\section{Introduction}
\label{sec:intro}
Pathology foundation models (PFMs) provide transferable representations learned from large collections of whole-slide images, image--text pairs, and multimodal clinical data~\citep{chen2024towards,xu2024whole,vorontsov2023virchow,lu2024visual,ding2025multimodal}. However, these models differ substantially in architecture, pretraining objective, training data and representation dimensionality. This raises a fundamental question: \emph{Do independently trained PFMs learn fundamentally different feature spaces or different coordinate systems of a shared underlying representation?}

Prior work has shown that independently trained neural networks can exhibit substantial representational convergence~\citep{li2015convergent,huh2024platonic}, while recent studies suggest that pretraining objectives strongly influence latent geometry~\citep{ciernik2024objective}. At the same time, latent-space alignment and model-stitching methods have shown that representations from different models can sometimes be translated or made interoperable~\citep{moschella2022relative,maiorca2023latent,bansal2021revisiting,achara2026multi}. In CPath, recent studies have begun to characterize similarity and redundancy across PFMs, but whether heterogeneous pathology encoders can be explicitly aligned into a shared and functionally useful space remains largely unexplored.

We introduce \textbf{MOSAIC} (\textbf{M}orphological \textbf{O}rganization and \textbf{S}hared \textbf{A}lignment for \textbf{I}nteroperable \textbf{C}omputational Pathology Foundation Models), a systematic study of representation similarity, alignment, and interoperability across $12$ pathology patch encoders~(UNI2-H~\cite{chen2024towards}, Prov-GigaPath~\cite{xu2024whole}, Virchow~\cite{vorontsov2023virchow}, Virchow2~\cite{zimmermann2024virchow2}, H-optimus-0~\cite{filiot2025distilling}, GPFM~\cite{ma2024towards}, CTransPath~\cite{wang2022transformer}, CONCH~\cite{lu2024visual}, CONCH v1.5~\cite{lu2024visual}, KEEP~\cite{zhou2026knowledge}, MUSK~\cite{xiang2025vision}, ResNet50~\cite{he2016deep}) and $6$ slide encoders~(CHIEF~\cite{wang2024pathology}, Madeleine~\cite{jaume2024multistain}, PRISM~\cite{shaikovski2024prism}, GigaPath~\cite{xu2024whole}, TITAN~\cite{ding2025multimodal}, Feather~\cite{shao2025multiple}). Additional details regarding these PFMs are provided in the supplementary material (Appendix \textbf{A}). We analyze how architecture and pretraining objective shape latent geometry, evaluate multiple strategies for mapping independently learned representations into a common space and test whether alignment preserves downstream molecular and morphological information. Finally, we examine cross-model transfer by asking whether predictors trained with one encoder can operate on aligned representations from another without retraining.

\subsection{Research Questions}

MOSAIC is organized around five questions that examine whether independently trained PFMs exhibit compatible latent structure, whether those spaces can be aligned, and whether such alignment is useful beyond geometric analysis.

\textit{\textbf{RQ1: How similar are the representation spaces of different pathology encoders?}}

Prior work has shown that independently trained neural networks can exhibit substantial representational convergence, even when their latent coordinates differ~\citep{li2015convergent,huh2024platonic,li2026platonic2}. A range of similarity measures, including canonical correlation analysis (CCA) based methods, centered kernel alignment (CKA) and representational similarity analysis, have consequently been developed to quantify such correspondence~\citep{raghu2017svcca,morcos2018insights,kornblith2019similarity,kriegeskorte2008representational}. We examine whether this convergence also emerges across heterogeneous PFMs.

\textit{\textbf{RQ2: What determines latent geometry: model architecture or pretraining objective?}}

Recent evidence suggests that representational similarity is strongly shaped by the learning objective and may remain consistent across datasets when models share similar pretraining objectives~\citep{ciernik2024objective,huh2024platonic,lu2026indra}. We therefore investigate whether pathology encoders cluster primarily according to architectural family or according to their supervision paradigm, including self-supervised, vision--language and supervised learning.

\textit{\textbf{RQ3: Can independently learned representation spaces be aligned into a shared latent space?}}

A growing body of work shows that independently learned latent spaces can be translated or aligned through relatively simple transformations, enabling communication between otherwise incompatible models~\citep{moschella2022relative,maiorca2023latent,cannistraci2024bricks,norelli2023asif}. More recent work extends this idea to simultaneous multi-model and geometry-preserving alignment~\citep{achara2026multi,yu2026geometry}. We test whether analogous alignment is possible across diverse PFMs.

\textit{\textbf{RQ4: Does alignment produce representations that remain useful for downstream pathology tasks?}}

Geometric alignment alone does not guarantee that task-relevant information is preserved. Prior work on latent-space translation, representation matching and model stitching has therefore evaluated aligned representations through downstream functional performance~\citep{maiorca2023latent,bansal2021revisiting,csiszarik2021similarity,achara2026multi}. We similarly assess whether the shared pathology space preserves discriminative information across molecular and morphological prediction tasks.

\textit{\textbf{RQ5: Can knowledge learned using one encoder transfer to another through the shared space?}}

A stronger notion of alignment is \emph{interoperability}: representations produced by different encoders should support the transfer of downstream models without retraining. Previous studies have demonstrated zero-shot latent communication, model stitching and backward-compatible representations across independently trained networks~\citep{moschella2022relative,bansal2021revisiting,csiszarik2021similarity,shen2020towards}. We test whether a predictor learned using one pathology encoder can operate on another encoder after mapping both into the shared space.

\subsection{Related Studies in CPath}

Recent work has begun to examine PFMs beyond downstream performance. Mishra et al.~\citep{mishra2026comparing} analyzed six PFMs using representational similarity analysis (RSA) across four TCGA cohorts, revealing substantial but heterogeneous representational similarity and no consistent evidence that shared pretraining paradigms yield more similar latent spaces. Flannery et al.~\citep{flannery2025information} studied redundancy and complementarity across patch-level and slide-level PFMs using CKA, singular vector canonical correlation analysis~(SVCCA), Procrustes distance, neighborhood overlap and proposed correlation-guided feature pruning to fuse complementary representations.

MOSAIC moves beyond similarity analysis and feature fusion by explicitly aligning heterogeneous PFMs into a shared latent space. We evaluate whether this shared geometry preserves downstream information and supports cross-model transfer, thereby testing whether independently trained pathology encoders can become functionally interoperable.

\section{Methodology}

\subsection{Datasets and Downstream Tasks}
We evaluate our analysis against \textbf{MOSAIC-Bench} (Table.~\ref{tab:downstream_tasks}), a unified benchmark designed to systematically assess representational similarity, cross-model alignment, shared-space transfer, and downstream utility across heterogeneous PFMs.

\begin{table}[h]
\centering
\caption{\textbf{MOSAIC-Bench cohorts and downstream tasks.}
All evaluations use patient-level splits to prevent data leakage.}
\label{tab:downstream_tasks}

\scriptsize
\setlength{\tabcolsep}{3.2pt}
\renewcommand{\arraystretch}{1.15}

\begin{tabularx}{\columnwidth}{
    l
    l
    l
    >{\raggedright\arraybackslash}X
}
\toprule
\textbf{Dataset} &
\textbf{Cohort} &
\textbf{Task Type} &
\textbf{Prediction Target} \\
\midrule

\multirow{4}{*}{TCGA}
& \multirow{2}{*}{TCGA-NSCLC}
& Histologic subtyping
& LUAD vs.\ LUSC \\
&
& Pathologic staging
& Tumor stage \\

\cmidrule(lr){2-4}

& \multirow{2}{*}{TCGA-BRCA}
& Histologic subtyping
& IDC vs.\ ILC \\
&
& Pathologic staging
& Tumor stage \\

\midrule

\multirow{3}{*}{CPTAC}
& CPTAC-LUAD
& Mutation prediction
& \textit{TP53}, \textit{KRAS}, \textit{STK11} \\

& CPTAC-BRCA
& Mutation prediction
& \textit{PIK3CA}, \textit{GATA3}, \textit{MAP3K1} \\

& CPTAC-COAD
& Mutation prediction
& \textit{TP53}, \textit{KRAS}, \textit{PIK3CA} \\

\bottomrule
\end{tabularx}
\end{table}
MOSAIC-Bench comprises whole-slide images (WSIs) from TCGA~\cite{weinstein2013cancer} ($2169$ WSIs) and CPTAC~\cite{edwards2015cptac} ($2162$ WSIs) cohorts spanning breast, lung and colon cancer. TCGA cohorts are used for histologic subtyping and pathologic staging, including LUAD vs.\ LUSC in NSCLC and IDC vs.\ ILC in breast cancer. CPTAC cohorts are used for molecular prediction, covering recurrent mutations in \textit{TP53}, \textit{KRAS}, \textit{STK11}, \textit{PIK3CA}, \textit{GATA3}, and \textit{MAP3K1}. All downstream evaluations use patient-level splits to prevent data leakage. Further details regarding the downstream tasks are provided in the supplementary material (Appendix \textbf{B}).

\subsection{Cross-Encoder Representation Similarity}\label{subsec:encoder_sim}

Let \(f_i\) denote the \(i\)-th pathology encoder and
\(\mathbf{X}^{(i)} \in \mathbb{R}^{N\times d_i}\) its representation
matrix for \(N\) matched samples, where \(d_i\) is the corresponding
feature dimensionality. For each compatible encoder pair
\((f_i,f_j)\), we compute the following:

\begin{equation}
S_{ij}^{(m)}
=
m\!\left(
\mathbf{X}^{(i)},
\mathbf{X}^{(j)}
\right)
\end{equation}

where \(m(\cdot,\cdot)\) denotes a representation-similarity measure.
Features are centered across samples before comparison.

We use seven complementary metrics spanning kernel, subspace,
geometric, relational, and statistical notions of similarity:

\begin{itemize}
    \item \textbf{Linear CKA} compares the geometry induced by the
    linear Gram matrices of two representations and is invariant to
    orthogonal transformations and isotropic scaling
    ~\citep{kornblith2019similarity}.

    \item \textbf{Kernel CKA} extends CKA to nonlinear kernels,
    allowing comparison of nonlinear relationships between
    representation spaces~\citep{kornblith2019similarity}.

    \item \textbf{SVCCA} identifies shared correlated subspaces by
    first reducing each representation with SVD and then applying
    canonical correlation analysis~\citep{raghu2017svcca}.

    \item \textbf{PWCCA} extends CCA by weighting canonical
    directions according to their contribution to the original
    representation~\citep{morcos2018insights}.

    \item \textbf{Procrustes Similarity} measures how closely two
    representations can be matched under an optimal orthogonal
    transformation, capturing geometric correspondence between
    feature spaces.

    \item \textbf{Cosine RSA} compares the relational structure of
    two representations through their pairwise cosine-distance
    matrices~\citep{kriegeskorte2008representational}.

    \item \textbf{Distance Correlation} measures general statistical
    dependence between representation spaces without requiring equal
    dimensionality or a linear relationship~\citep{szekely2007measuring}.
\end{itemize}

For each metric, we construct an encoder-by-encoder similarity matrix
\(\mathbf{S}^{(m)}\), providing complementary views of representational
convergence across PFMs. Full mathematical
definitions and implementation details for all similarity measures are
provided in the supplementary material (Appendix \textbf{C}).

\subsection{Multi-Encoder Shared-Space Alignment}

Given \(M\) encoders (indexed by $i$) with matched representation matrices
\(\{\mathbf{X}^{(i)}\}_{i=1}^{M}\), where
\(\mathbf{X}^{(i)} \in \mathbb{R}^{N\times d_i}\), we seek
encoder-specific mappings \(T_i\) into a common \(d\)-dimensional
latent space:

\begin{equation}
\mathbf{Z}^{(i)}
=
T_i\!\left(\mathbf{X}^{(i)}\right)
\in
\mathbb{R}^{N\times d}
\end{equation}

Here, $\mathbf{Z}^{(i)}$ denotes the aligned representation of encoder $i$ in the shared $d$-dimensional latent space. We therefore learn an encoder-specific transformation for each representation space,
such that:

\begin{equation}
\mathbf{X}^{(1)}
\xrightarrow{\,T_1\,}
\mathbf{Z}^{(1)},
\qquad
\mathbf{X}^{(2)}
\xrightarrow{\,T_2\,}
\mathbf{Z}^{(2)}
\end{equation}

where both $\mathbf{Z}^{(1)}$ and $\mathbf{Z}^{(2)}$ lie in the same shared latent space. We compare five
complementary alignment strategies spanning correlation-based,
geometric, shared-subspace, and nonlinear formulations
~\citep{maiorca2023latent,achara2026multi}:

\begin{itemize}

    \item \textbf{Generalized CCA (GCCA)} uses the MAX-VAR formulation
    to learn encoder-specific linear projections that map all
    representation spaces toward a common latent representation
    ~\citep{kettenring1971canonical,benton2019deep}.

    \item \textbf{Multiset CCA (MCCA)} uses the SUMCOR formulation to
    maximize the summed pairwise correlations among projected encoder
    representations, directly encouraging agreement across all model
    pairs~\citep{kettenring1971canonical}.

    \item \textbf{Generalized Procrustes Alignment (GPA)} first reduces
    each representation to the target dimensionality using PCA and then
    aligns the resulting spaces through orthogonal transformations and
    global scaling, largely preserving their internal geometry
    ~\citep{gower1975generalized}.

    \item \textbf{Joint PCA} constructs a shared linear subspace by
    applying PCA to the concatenated matched representations and then
    learns an encoder-specific ridge-regression mapping from each
    original feature space into the common coordinates.

    \item \textbf{Shared Autoencoder} learns encoder-specific nonlinear
    encoders and decoders coupled through a shared bottleneck. The model
    jointly optimizes reconstruction and cross-model alignment losses,
    encouraging matched samples to converge while preserving
    encoder-specific information.

\end{itemize}

All alignment models are fit using training samples only and are applied
unchanged to validation and test representations to prevent information
leakage. Full mathematical formulations, optimization objectives, and
implementation details are provided in the supplementary material (Appendix \textbf{D}).

\subsection{Shared-Space and Downstream Evaluation}

We evaluate each alignment method from both geometric and functional
perspectives to determine whether the learned shared space preserves
cross-model correspondence while retaining task-relevant information.  Further Details are provided in the supplementary material (Appendix \textbf{E}).

\begin{itemize}
\item \textbf{Cross-Model Agreement and Retrieval.}
For aligned representations
\(\mathbf{Z}^{(i)},\mathbf{Z}^{(j)}\in\mathbb{R}^{N\times d}\),
we measure the cosine similarity between representations of matched
samples:

\begin{equation}
S_{\mathrm{align}}(i,j)
=
\frac{1}{N}
\sum_{n=1}^{N}
\cos\!\left(
\mathbf{z}^{(i)}_n,
\mathbf{z}^{(j)}_n
\right)
\end{equation}

We additionally evaluate cross-model retrieval, where representations
from encoder \(i\) are used to query a database constructed from
encoder \(j\). We consider both identity retrieval, in which the target
is the corresponding representation of the same patch, and
label-based retrieval, in which representations sharing the same class
label are considered relevant. Retrieval quality is evaluated using
Recall@\(K\) (\(K\in\{1,5,10\}\)), mean average precision (mAP),
mean reciprocal rank (MRR) and normalized discounted cumulative gain
(NDCG).

\item\textbf{Geometry Preservation.}
Alignment should improve cross-model correspondence without destroying
the structure of the original representation spaces. We therefore
evaluate neighborhood preservation by measuring the overlap between the
\(k\)-nearest neighbors of each sample before and after projection into
the shared space. We additionally assess reconstruction fidelity by
mapping each shared representation back to its source encoder space and
reporting normalized mean squared error (NMSE), \(R^2\) and cosine
similarity between the original and reconstructed representations.

\item\textbf{Downstream Utility.}
We evaluate the functional utility of the learned shared representations
on the downstream tasks.
Under an identical patient-level MIL evaluation protocol, we compare
three input conditions: individual encoder representations, concatenated
multi-encoder representations, and aligned shared representations.
ABMIL and TransMIL are used as the primary downstream model, with the alignment
function fitted using training data only to prevent information leakage.
Performance is evaluated using area under the ROC curve (AUC), with patient-level train/test
separation maintained throughout.

\item \textbf{Cross-Model Transfer.}
We evaluate whether the shared representation enables direct translation
between encoder feature spaces. Given a representation
\(\mathbf{X}^{(i)}\) from source encoder \(i\), we first map it into the
shared latent space using \(E_i\) and then project it into the native
feature space of target encoder \(j\) using \(D_j\):

\begin{equation}
\mathbf{X}^{(i)}
\xrightarrow{E_i}
\mathbf{Z}
\xrightarrow{D_j}
\widetilde{\mathbf{X}}^{(j)},
\qquad i \neq j
\end{equation}

Here, \(\widetilde{\mathbf{X}}^{(j)}\) denotes the target-space
representation reconstructed from encoder \(i\). We compare it with the
corresponding native target representation \(\mathbf{X}^{(j)}\) to
measure how faithfully information can be transferred across encoders.

\end{itemize}

\subsection{Factors Governing Representation Geometry} 

\begin{figure*}
    \centering
    \includegraphics[width=0.7\linewidth]{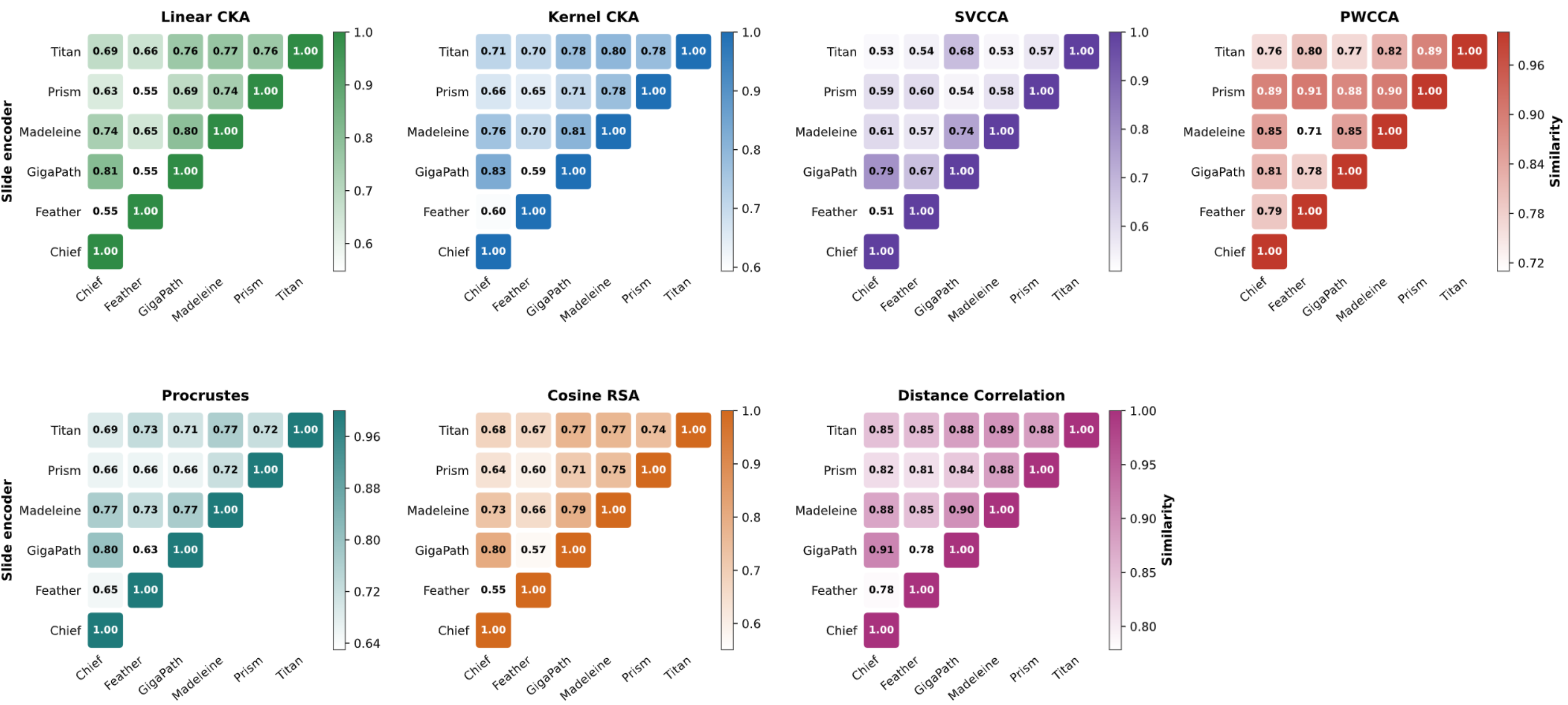}
    \caption{\textbf{Representational similarity among slide-level pathology encoders on CPTAC.}
Pairwise similarity between encoders is evaluated using representational similarity metrics mentioned in \textbf{Sec.}~\ref{subsec:encoder_sim}. Each panel corresponds to a different metric, with the lower triangular portion of the symmetric similarity matrix shown for clarity.}
    \label{fig:slide}
\end{figure*}
Beyond pairwise similarity, we investigate which model design choices
most strongly shape the representation geometry of pathology foundation
models. Prior work suggests that representational similarity can depend
strongly on the learning objective, architecture and network depth
~\citep{huh2024platonic,ciernik2024objective,lu2026indra}. We therefore
analyze similarity with respect to pretraining objective, architecture,
layer depth, and image magnification.

\begin{itemize}
\item \textbf{Architecture vs.\ Pretraining Objective.}
We examine whether representation geometry is influenced more by model
architecture or by the pretraining objective. Each encoder \(f_i\) is
annotated with an architectural family \(a_i\) and a pretraining family
\(o_i\), such as vision self-supervised, vision--language or supervised
learning.

Given an encoder similarity matrix \(\mathbf{S}^{(m)}\), where
\(S_{ij}^{(m)}\) measures the similarity between encoders \(f_i\) and
\(f_j\) under metric \(m\), we compare models that share the same design
factor with those that do not. For the pretraining objective, we define the following:

\begin{equation}\label{eq:6}
\Delta_{\mathrm{obj}}^{(m)}
=
\mathbb{E}\!\left[
S_{ij}^{(m)} \mid o_i=o_j
\right]
-
\mathbb{E}\!\left[
S_{ij}^{(m)} \mid o_i\neq o_j
\right]
\end{equation}

and similarly for architecture as follows:

\begin{equation}\label{eq:7}
\Delta_{\mathrm{arch}}^{(m)}
=
\mathbb{E}\!\left[
S_{ij}^{(m)} \mid a_i=a_j
\right]
-
\mathbb{E}\!\left[
S_{ij}^{(m)} \mid a_i\neq a_j
\right]
\end{equation}

Larger positive values indicate that models sharing the same pretraining
objective or architecture tend to learn more similar representations.
Comparing the two effects helps determine which factor is more strongly
associated with representation geometry.

\item \textbf{Layer-wise Similarity.}
To examine how representational similarity evolves with network depth, we
extract intermediate representations from each transformer block (or
stage for convolutional architectures) using the same set of image
patches. Let
\(\mathbf{X}^{(i,\ell)}\in\mathbb{R}^{N\times d_{i,\ell}}\)
denote the representation produced at layer \(\ell\) of encoder \(f_i\).
For every pair of layers \((\ell,r)\) from encoders \(f_i\) and \(f_j\),
we compute linear CKA as follows:

\begin{equation}
S_{ij}^{(\ell,r)}
=
\operatorname{CKA}\!\left(
\mathbf{X}^{(i,\ell)},
\mathbf{X}^{(j,r)}
\right)
\end{equation}

This produces a layer-by-layer similarity matrix describing how the two
representation spaces evolve across depth. Since models contain
different numbers of layers, layer positions are normalized to relative
depth in \([0,1]\). For each layer of \(f_i\), we additionally identify
the most similar layer of \(f_j\), producing an alignment trajectory
that reveals whether comparable representations emerge at similar
depths.

\item \textbf{Magnification-dependent Similarity.}
We further investigate whether cross-model agreement depends on tissue
resolution. For each magnification
\(s\in\{5\times,10\times,20\times\}\), we independently repeat the
representation-similarity analysis using the same compatible encoders,
slides, and sampling protocol. For metric \(m\), this produces a
magnification-specific similarity matrix defined as follows:

\begin{equation}
\mathbf{S}^{(m,s)}
=
\left[
S_{ij}^{(m,s)}
\right]_{i,j}
\end{equation}

We compare these matrices across magnifications to determine whether
cross-model agreement changes from coarse tissue organization to finer
cellular morphology. Because patch coordinate grids differ across
magnifications, embeddings are not directly paired between resolutions;
instead, each magnification is treated as an independent replication of
the similarity analysis.

\section{Results}

To be concise, only a specific subset of results is presented in this section. Additional results across all cohorts along with hyperparameters are provided in the supplementary material (Appendix \textbf{F}).

\subsection{Representational Similarity Analysis}

\subsubsection{Patch-Level Representational Similarity}

For the 224-px encoder group (Table.~\ref{tab:similarity_224px_vertical}), the strongest agreement is observed at lower and intermediate magnifications, with a consistent drop at \(20\times\). On CPTAC, Linear CKA decreases from \(0.555\) at \(5\times\) to \(0.498\) at \(20\times\), while Kernel CKA falls from \(0.594\) to \(0.526\), RSA from \(0.530\) to \(0.485\), and dCor from \(0.765\) to \(0.722\). A similar pattern appears on TCGA, where Linear CKA drops from \(0.517\) at \(5\times\) to \(0.459\) at \(20\times\), Kernel CKA from \(0.554\) to \(0.488\), RSA from \(0.490\) to \(0.396\), and dCor from \(0.732\) to \(0.693\). In contrast, CCA-based measures are comparatively stable across magnification, with PWCCA remaining between \(0.772\)--\(0.786\) on CPTAC and \(0.767\)--\(0.838\) on TCGA. Overall, the repeated decline in geometry-based similarity at \(20\times\) suggests that encoder representations diverge more strongly as finer morphological detail becomes dominant.

\subsubsection{Slide-Level Representational Similarity}

We evaluate pairwise representational similarity among six slide-level encoders (shown for \textbf{CPTAC} in Fig.~\ref{fig:slide}). Linear CKA shows moderate agreement (mean \(=0.689\), range \(0.548\)--\(0.810\)), with geometry-based measures producing highly consistent rankings; Linear CKA correlates strongly with Kernel CKA (\(\rho=0.996\)) and distance correlation (\(\rho=0.964\)).

\begin{table}[htbp]
\centering
\caption{\textbf{Representational similarity for the 224-px encoder group.}
Mean pairwise similarity across CPTAC and TCGA at different magnifications.}
\label{tab:similarity_224px_vertical}

\scriptsize
\setlength{\tabcolsep}{2.5pt}
\renewcommand{\arraystretch}{1.08}

\begin{tabular}{lccc|ccc}
\toprule
& \multicolumn{3}{c|}{\textbf{CPTAC}} 
& \multicolumn{3}{c}{\textbf{TCGA}} \\
\cmidrule(lr){2-4}\cmidrule(lr){5-7}
\textbf{Metric} &
\textbf{5$\times$} &
\textbf{10$\times$} &
\textbf{20$\times$} &
\textbf{5$\times$} &
\textbf{10$\times$} &
\textbf{20$\times$} \\
\midrule

LinCKA  & 0.555 & 0.540 & 0.498 & 0.517 & 0.532 & 0.459 \\
KerCKA  & 0.594 & 0.571 & 0.526 & 0.554 & 0.571 & 0.488 \\
SVCCA   & 0.555 & 0.579 & 0.567 & 0.602 & 0.619 & 0.568 \\
PWCCA   & 0.772 & 0.786 & 0.779 & 0.836 & 0.838 & 0.767 \\
Procr.  & 0.700 & 0.711 & 0.684 & 0.705 & 0.707 & 0.650 \\
RSA     & 0.530 & 0.506 & 0.485 & 0.490 & 0.478 & 0.396 \\
dCor    & 0.765 & 0.751 & 0.722 & 0.732 & 0.745 & 0.693 \\

\bottomrule

\end{tabular}
\end{table}

In contrast, CCA-based measures behave differently: PWCCA yields higher absolute similarity (mean \(=0.828\)) but correlates weakly with Linear CKA (\(\rho=0.129\)), while SVCCA and PWCCA are also weakly correlated (\(\rho=0.118\)). Given the relatively small slide count compared with the embedding dimensionalities (\(d=512\)--\(1{,}280\)), we primarily interpret CKA, Procrustes, cosine RSA, and distance correlation, while retaining CCA-based metrics as complementary measures. Overall, these findings address \textbf{RQ1} by demonstrating substantial, but metric-dependent, convergence among slide-level pathology representations.

\end{itemize}

\begin{figure}
    \centering
    \includegraphics[width=\linewidth]{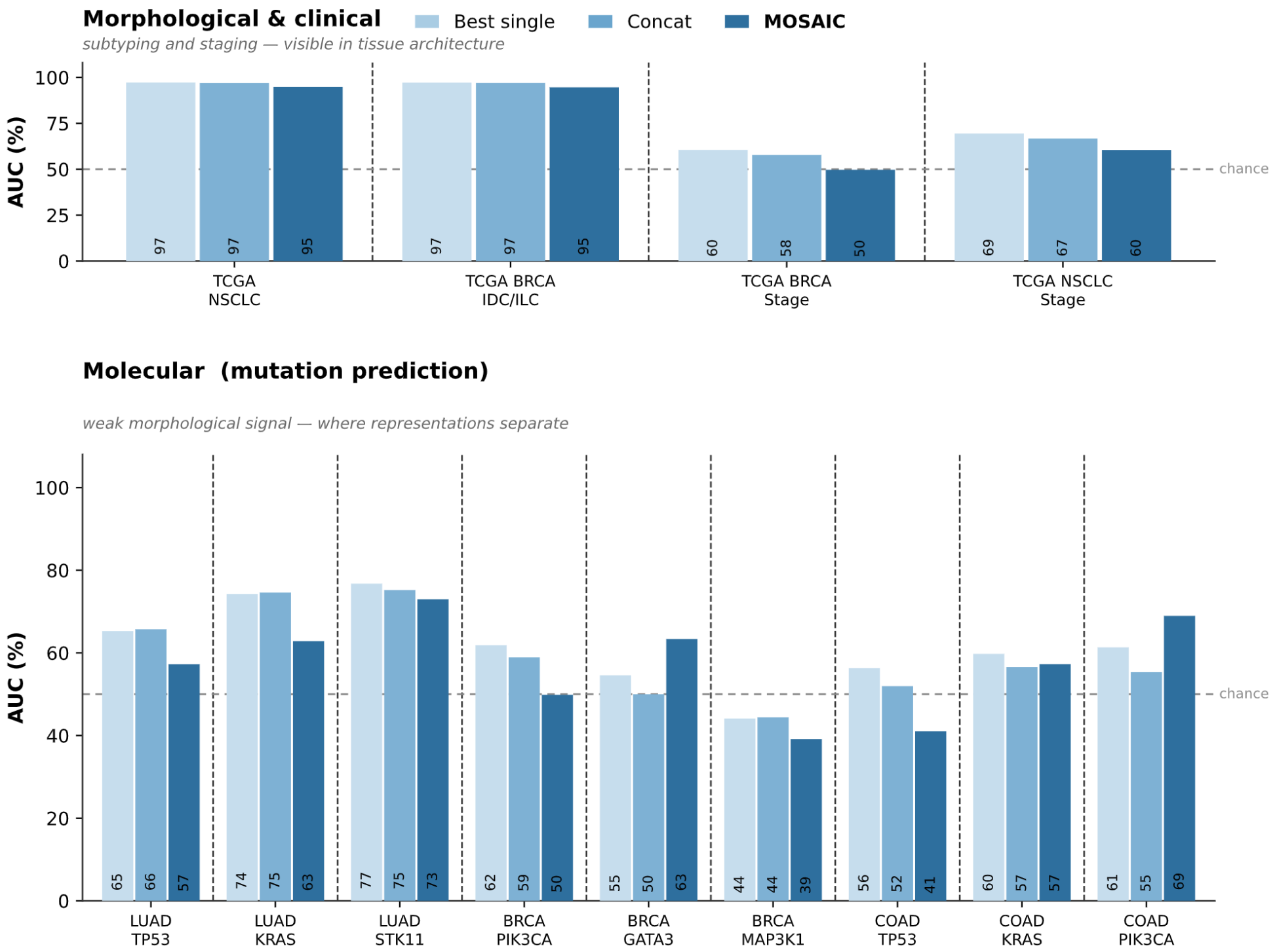}
    \caption{Downstream MIL performance across representation strategies. Best single, feature concatenation, and MOSAIC shared-space representations are compared on morphological/clinical tasks (\textbf{top}) and mutation prediction tasks (\textbf{bottom}). Morphological tasks show similar performance across representations, while molecular tasks exhibit larger differences, with MOSAIC improving selected targets. Computed with the 256-px encoder group at 10$\times$ magnification.}
    \label{fig:downstream}
\end{figure}

\subsection{Downstream Utility of the Shared Representation Space}

We evaluate whether the MOSAIC shared space preserves information useful for slide-level prediction using attention-based MIL, which answers \textbf{RQ4}. For this analysis, patch embeddings are extracted at $10\times$ magnification from the 256-px encoder group. We compare the best individual encoder, direct feature concatenation, and a 64-dimensional GCCA-based MOSAIC space learned using training slides only. The MIL formulation is held fixed, and results are averaged across ABMIL and TransMIL with patient-grouped, stratified splits.

As shown in Fig.~\ref{fig:downstream}, morphological and clinical tasks exhibit relatively small differences across representation strategies. TCGA NSCLC subtyping and BRCA IDC/ILC classification reach AUCs near $0.95$--$0.97$, while staging tasks show the same general pattern at lower absolute performance. This suggests that visually explicit phenotypes are already captured consistently by individual pathology encoders, limiting the benefit of either concatenation or shared-space alignment. In contrast, molecular prediction tasks show substantially larger representation-dependent variation. Direct concatenation does not consistently outperform the best single encoder despite its much higher dimensionality, indicating that additional features do not necessarily provide additional useful signal.

MOSAIC produces selective gains on molecular tasks, improving BRCA GATA3 from approximately $0.55$ to $0.63$ AUC and COAD PIK3CA from approximately $0.61$ to $0.69$, while underperforming on targets such as LUAD TP53 and COAD TP53. These results indicate that the value of representation alignment is strongly task-dependent. Morphological tasks appear to rely largely on information shared across encoders, whereas molecular prediction depends more on complementary and encoder-specific signals. Importantly, the compact 64-dimensional MOSAIC space can outperform both the best individual encoder and high-dimensional concatenation on selected targets, suggesting that alignment can concentrate useful shared information, although it does not preserve all task-specific signals equally well.

\subsection{Layer-wise Representational Alignment}

\begin{figure}[htbp]
    \centering
    \includegraphics[width=0.85\linewidth]{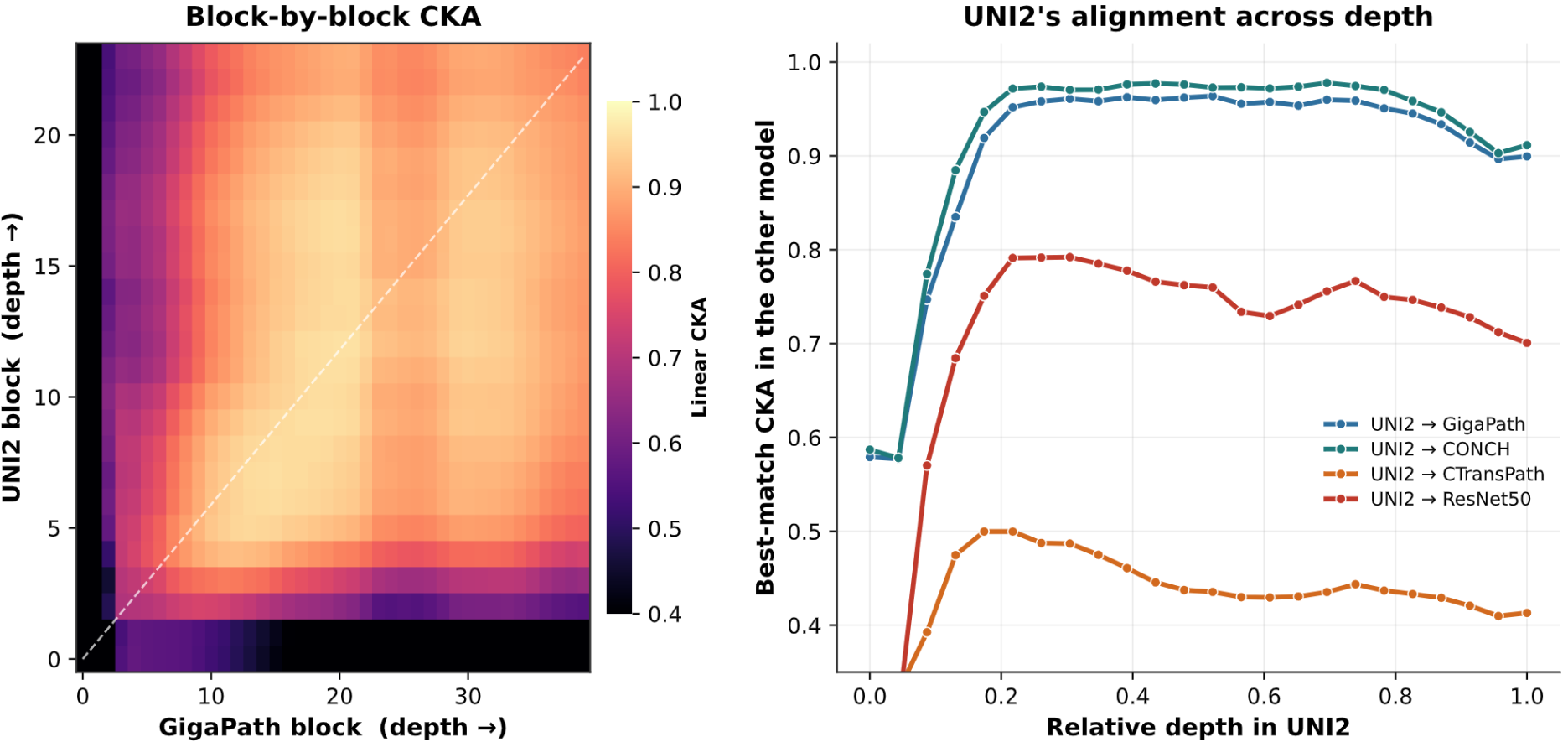}
    \caption{\textbf{Layer-wise representational alignment across pathology encoders on TCGA at 10$\times$ for the 256-px encoder group.}
Left: block-by-block linear CKA between CONCH and UNI2 representations. Right: best-matching CKA between each CONCH layer and layers from other encoders, plotted over relative CONCH depth.}
    \label{fig:layerwise_alignment}
\end{figure}

To further investigate \textbf{RQ2}, we examine (Fig.~\ref{fig:layerwise_alignment}) how representational similarity evolves across network depth on \textbf{TCGA} at 10$\times$ magnification with 256-px encoder group, rather than comparing only final-layer embeddings. Fig.~\ref{fig:layerwise_alignment} shows block-level alignment using CONCH as the reference encoder. The CONCH--UNI2 CKA matrix reveals a broad region of high similarity across intermediate and deeper layers, indicating that correspondence is not restricted to blocks at identical depths. Instead, representations learned at one depth in CONCH can closely match representations occurring at different depths in UNI2.
The depth-wise analysis further shows that cross-model alignment is strongly model-dependent. CONCH becomes increasingly aligned with UNI2 and GigaPath through the early and intermediate layers, reaching very high similarity in the deeper representation hierarchy. In contrast, alignment with CTransPath remains substantially lower, while ResNet50 exhibits an intermediate pattern with similarity decreasing toward later CONCH layers. These results suggest that different encoder families follow distinct representational trajectories across depth, even when evaluated on the same TCGA patches.

Together, these findings extend \textbf{RQ2} beyond endpoint similarity, showing that pathology encoders differ not only in the representations they ultimately produce, but also in where along network depth comparable representations emerge.

\subsection{Effect of Magnification on Cross-Model Similarity}

\begin{figure}[htbp]
    \centering
    \includegraphics[width=0.7\linewidth]{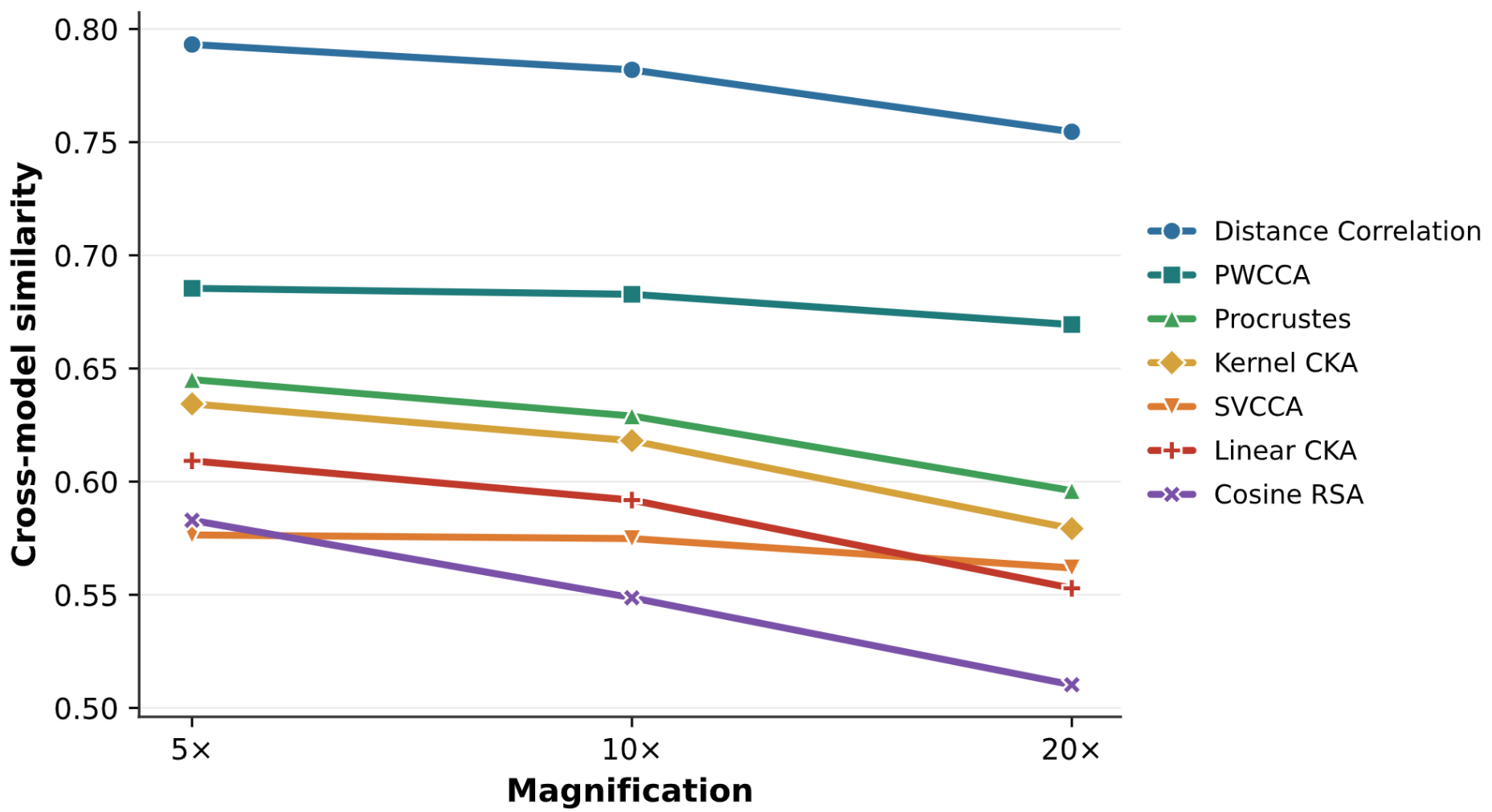}
    \caption{\textbf{Cross-model similarity across magnification levels.}
Mean pairwise similarity among the five encoders shared across $5\times$, $10\times$, and $20\times$ CPTAC experiments. All seven similarity metrics decrease with increasing magnification, indicating greater agreement at coarse tissue scale and increasing representational divergence at finer spatial resolution.}
    \label{fig:magnification}
\end{figure}

We next examine whether cross-model representational similarity depends on image magnification. The analysis is repeated independently at $5\times$, $10\times$, and $20\times$ using $256\times256$ pixel patches from the common CPTAC slide set. To ensure a consistent comparison, we restrict the analysis to the five encoders available at all three magnifications: \textbf{CTransPath}, \textbf{Prov-GigaPath}, \textbf{KEEP}, \textbf{ResNet50}, and \textbf{UNI2-H}. Each point in Fig.~\ref{fig:magnification} represents the mean off-diagonal similarity across all encoder pairs. Importantly, the patch sets at different magnifications are sampled independently; therefore, the experiment compares the resulting cross-model similarity structure rather than establishing patch-level correspondence across scales.

A consistent trend emerges across all seven similarity measures: \textbf{cross-model similarity decreases as magnification increases}. For example, Linear CKA decreases from $0.609$ at $5\times$ to $0.592$ at $10\times$ and $0.553$ at $20\times$. The same monotonic pattern is observed for the other metrics. Although the absolute similarity values differ substantially across metrics, their agreement on the direction of change indicates that the observed effect is not specific to a particular similarity formulation. These results suggest that pathology encoders converge more strongly on \textbf{coarse tissue organization and global morphology} captured at lower magnification, while becoming increasingly model-specific as finer cellular and nuclear detail is introduced. This analysis directly addresses \textbf{RQ1} by examining whether cross-model representational similarity remains consistent across different tissue magnification levels.

\section{Multi-Encoder Shared-Space Evaluation}

\subsection{Alignment Quality}

\begin{figure}[htbp]
    \centering
    \includegraphics[width=0.9\linewidth]{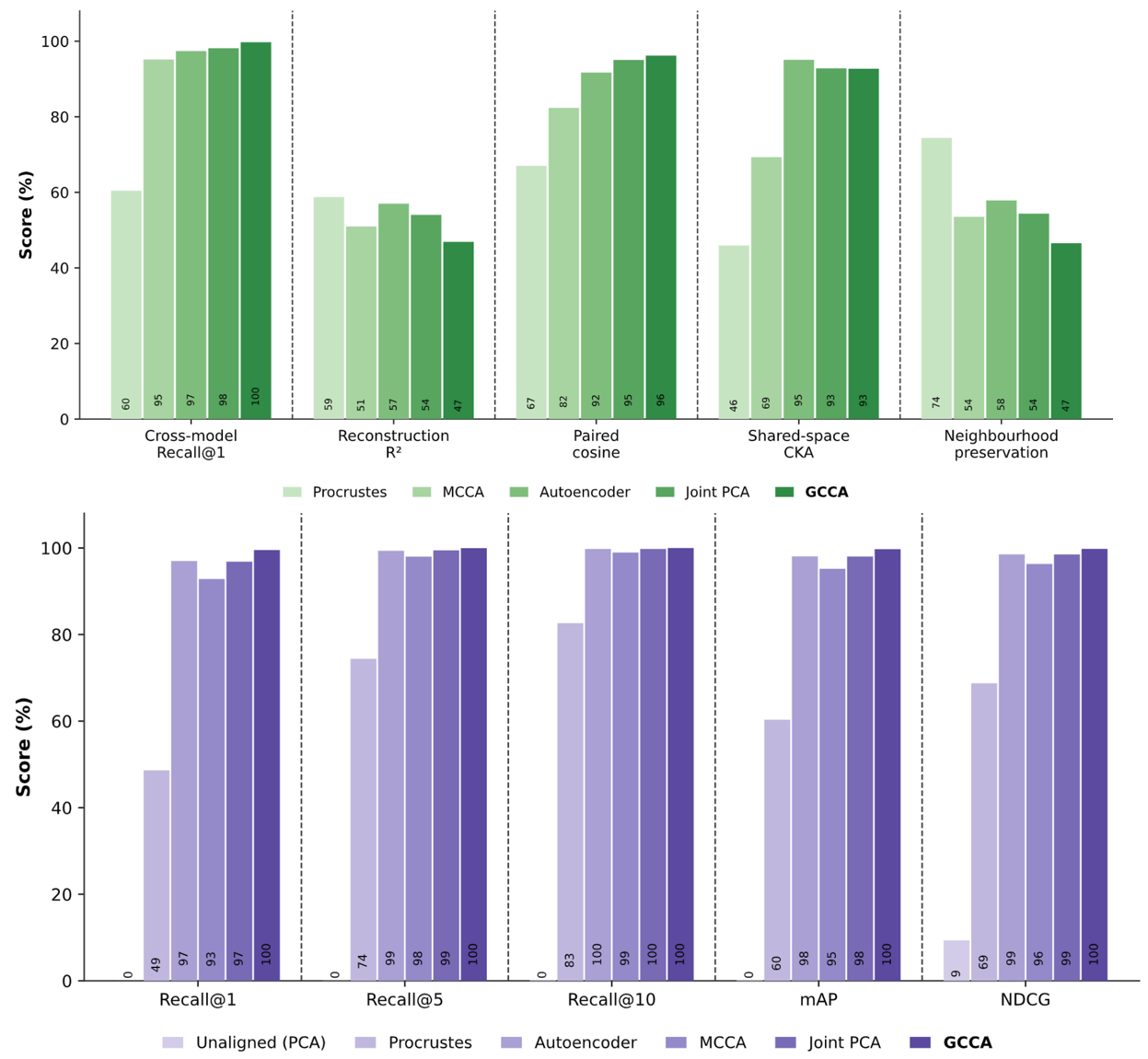}
  \caption{\textbf{Shared-space alignment and cross-encoder retrieval.}
Top: comparison of alignment methods across reconstruction, correspondence, shared-space similarity, and neighborhood preservation. Bottom: cross-encoder retrieval performance before and after alignment, showing the extent to which heterogeneous encoder representations can be mapped into a common latent space.}
    \label{fig:alignment_retrieval}
\end{figure}

To address \textbf{RQ3}, we first evaluate (Fig.~\ref{fig:alignment_retrieval}) how well each alignment method balances cross-model correspondence with preservation of the original representation geometry. A clear trade-off emerges. Procrustes provides the strongest reconstruction quality (\(R^2=0.588\)) and neighborhood preservation (\(0.744\)), but substantially weaker cross-model correspondence, with paired cosine similarity of \(0.670\) and shared-space Recall@1 of \(0.604\).

In contrast, GCCA achieves the lowest alignment error (\(0.030\)), highest paired cosine similarity (\(0.962\)), and highest shared-space Recall@1 (\(0.997\)), while using nearly the full 64-dimensional latent space. Joint PCA and the shared autoencoder also produce strong alignment, with the autoencoder achieving the highest shared-space CKA (\(0.951\)). These results indicate that preserving encoder-specific geometry and enforcing cross-model correspondence are competing objectives, with GCCA favoring alignment and Procrustes favoring structural preservation.

\subsection{Cross-Encoder Retrieval}

We further evaluate \textbf{RQ3} through cross-encoder retrieval~(Fig.~\ref{fig:alignment_retrieval}), testing whether aligned representations become directly interchangeable across models. Each of 5,000 query patches is embedded by one encoder, projected into the shared space, and retrieved from 5,000 candidate embeddings produced by a different encoder. Without alignment, Recall@1 remains approximately at the \(1/5000\) chance level, and median ranks lie near the middle of the candidate set, indicating that native encoder spaces are not directly compatible.

After alignment, retrieval improves dramatically. GCCA achieves the strongest performance, followed by the shared autoencoder, Joint PCA, and MCCA. GCCA is also highly consistent across encoder pairs, whereas Procrustes exhibits substantially greater variation and directional asymmetry. These results provide strong evidence for \textbf{RQ3}: although PFMs occupy incompatible native embedding spaces, their representations can be mapped into a highly consistent shared latent space that enables direct cross-model retrieval.

\begin{figure}
    \centering
    \includegraphics[width=0.8\linewidth]{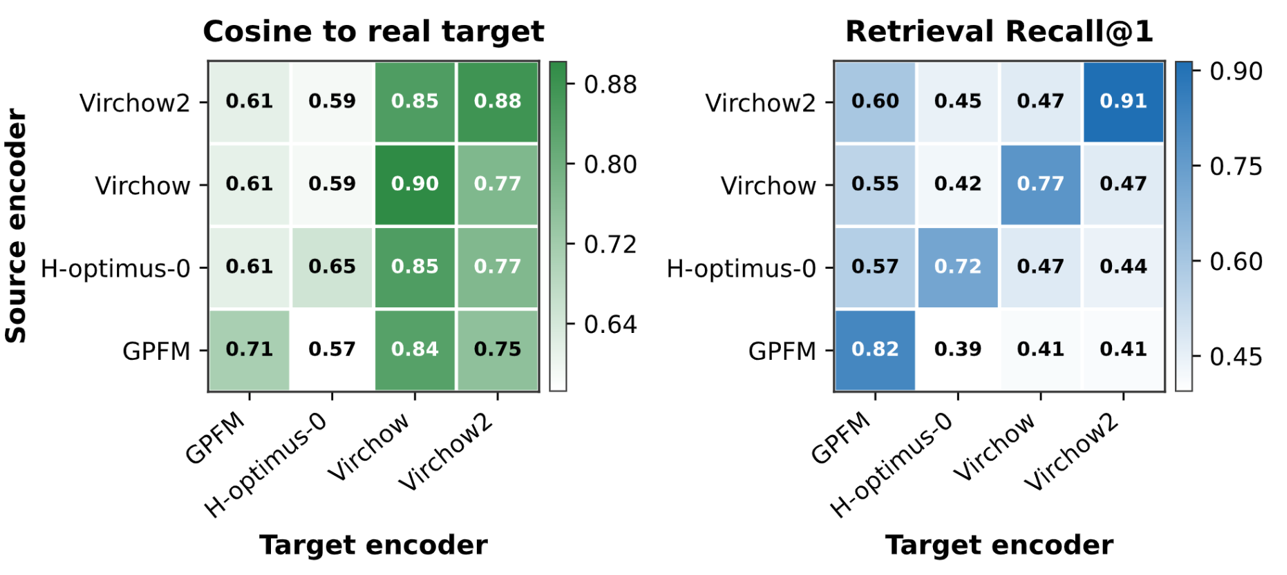}
    \caption{\textbf{Cross-encoder translation through the shared latent space.}
Each source representation is mapped through the shared space into a target encoder's representation space. The heatmaps report cosine similarity to the true target representation and cross-encoder retrieval Recall@1; diagonal entries correspond to self round-trip reconstruction and provide an upper bound imposed by the shared-space bottleneck.}
    \label{fig:transfer}
\end{figure}

\subsection{Cross-Encoder Representation Translation}
To address \textbf{RQ5}, we test (Fig.~\ref{fig:transfer}, Table.~\ref{tab:cross_encoder_translation}) whether the shared latent space can translate representations between encoders. A patch embedding from source encoder \(A\) is projected into the shared space and reconstructed in target encoder \(B\)'s space, then compared with \(B\)'s native embedding for the same patch. We also evaluate \textit{self round-trip} translation (\(A=B\)) to quantify information loss introduced by the 64-dimensional bottleneck.

\begin{table}[t]
\centering
\caption{\textbf{Cross-model translation versus self round-trip reconstruction.}
Self round-trip provides a bottleneck-aware upper bound for cross-encoder transfer.}
\label{tab:cross_encoder_translation}
\resizebox{0.72\linewidth}{!}{
\begin{tabular}{lcccc}
\toprule
\textbf{Condition} & \textbf{Cosine} & \textbf{$R^2$} & \textbf{Recall@1} & \textbf{mAP} \\
\midrule
Cross-model      & 0.702 & 0.393 & 0.470 & 0.572 \\
Self round-trip  & 0.784 & 0.563 & 0.807 & 0.863 \\
\bottomrule
\end{tabular}}
\end{table}

Cross-model Recall@1 reaches \(58\%\) of the self round-trip ceiling (\(0.470\) vs.\ \(0.807\)), showing that part of the transfer loss arises from the shared-space bottleneck itself rather than cross-model translation alone. The encoder-pair matrices also reveal strong directionality: translation into Virchow is consistently easier, whereas H-optimus-0 is a more difficult target, indicating that source and target transferability are distinct properties.

Overall, these results support \textbf{RQ5}: the shared latent space enables meaningful transfer between heterogeneous PFMs while exposing directional asymmetries that symmetric similarity measures cannot capture.

\subsection{Architecture or Pretraining Objective Analysis}

This particular analysis (Table.~\ref{tab:rq2_overall}, Table.~\ref{tab:rq2_cka}, Table.~\ref{tab:rq2_exact_arch}) is based on the formulation provided in Eq.~\ref{eq:6} and Eq.~\ref{eq:7}. Across the six groups where both effects are defined, \(\Delta_{\mathrm{obj}}\) is positive for all \(42\) group--metric combinations, whereas \(\Delta_{\mathrm{arch}}\) becomes negative in seven cases. The mean objective effect is also larger (\(+0.076\) vs.\ \(+0.051\)), indicating that models sharing a pretraining objective are more consistently representationally similar than models sharing a backbone family.

\begin{table}[t]
\centering
\caption{\textbf{Overall objective and architecture effects} across six groups and seven similarity metrics.}
\label{tab:rq2_overall}
\scriptsize
\setlength{\tabcolsep}{3pt}
\renewcommand{\arraystretch}{0.9}
\begin{tabular}{lcccc}
\toprule
\textbf{Effect} & \textbf{$n$} & \textbf{All positive} & \textbf{Range} & \textbf{Mean} \\
\midrule
$\Delta_{\mathrm{obj}}$  & 42 & Yes & $+0.026$--$+0.192$ & $\mathbf{+0.076}$ \\
$\Delta_{\mathrm{arch}}$ & 42 & No (7 negative) & $-0.047$--$+0.180$ & $+0.051$ \\
\bottomrule
\end{tabular}
\end{table}

The separation is particularly clear on TCGA. For Linear CKA, the objective effect exceeds the backbone effect at every magnification, while at \(20\times\) the architecture effect becomes negative despite a positive objective effect. On CPTAC, the two effects are closer because ResNet50 is simultaneously the only supervised model and the only pure CNN, partially confounding objective and architecture.
\begin{table}[t]
\centering
\caption{\textbf{Objective versus backbone effect for Linear CKA.}}
\label{tab:rq2_cka}
\scriptsize
\setlength{\tabcolsep}{5pt}
\renewcommand{\arraystretch}{1.2}
\begin{tabular}{lcc}
\toprule
\textbf{Group} & $\boldsymbol{\Delta_{\mathrm{obj}}}$ & $\boldsymbol{\Delta_{\mathrm{arch}}}$ \\
\midrule
CPTAC $5\times$/256 px  & +0.098 & +0.092 \\
CPTAC $10\times$/256 px & +0.138 & +0.134 \\
CPTAC $20\times$/256 px & +0.192 & +0.180 \\
TCGA $5\times$/256 px   & +0.032 & +0.002 \\
TCGA $10\times$/256 px  & +0.054 & +0.023 \\
TCGA $20\times$/256 px  & +0.080 & $\mathbf{-0.014}$ \\
\bottomrule
\end{tabular}
\end{table}
A cleaner architecture test is provided by the \(224\)-px groups, where all four encoders share the same vision-SSL objective. Virchow and Virchow2 are the only exact architecture-matched pair, yet their architecture effect is strongly negative for geometry-sensitive metrics and replicates across TCGA and CPTAC.
\begin{table}[t]
\centering
\caption{\textbf{Exact-architecture effect} in the objective-controlled $(20\times/224$ px setting).}
\label{tab:rq2_exact_arch}
\scriptsize
\setlength{\tabcolsep}{5pt}
\renewcommand{\arraystretch}{1.2}
\begin{tabular}{lcc}
\toprule
\textbf{Metric} & \textbf{CPTAC} & \textbf{TCGA} \\
\midrule
Linear CKA           & $\mathbf{-0.216}$ & $\mathbf{-0.197}$ \\
Kernel CKA           & -0.188 & -0.182 \\
Cosine RSA           & -0.185 & -0.150 \\
Distance Correlation & -0.130 & -0.133 \\
PWCCA                & -0.055 & -0.050 \\
Procrustes           & -0.022 & +0.024 \\
SVCCA                & -0.017 & -0.005 \\
\bottomrule
\end{tabular}
\end{table}
Overall, \textbf{RQ2} indicates that representational geometry tracks pretraining objective more consistently than architectural family. Shared architecture alone does not imply similar representations, while the objective effect remains positive across every evaluated group and metric.

\paragraph{Subcohort-level patch similarity.}
Across CPTAC and TCGA, the subcohort analysis reveals several consistent patterns in patch-level representational agreement. At 224\,px, the similarity structure is particularly stable across subcohorts: \textbf{GPFM--H-optimus-0} generally shows the strongest agreement under Linear/Kernel CKA, Procrustes similarity, Cosine RSA, and dCor, whereas \textbf{GPFM--Virchow2} more often attains the highest SVCCA and PWCCA values. For many subcohorts, geometric agreement decreases as magnification increases from $5\times$ to $20\times$, suggesting that encoders agree more strongly on coarse tissue organization and diverge as finer morphological detail becomes dominant, although some subcohort-specific exceptions remain.

At 256\,px, the similarity structure is substantially more heterogeneous: the strongest encoder pairing changes across magnification, subcohort, and metric, with relationships such as \textbf{KEEP--UNI2-h}, \textbf{CONCH--KEEP}, \textbf{Prov-GigaPath--UNI2-h}, and \textbf{CTransPath--ResNet50} emerging in different settings. In contrast, the 512\,px comparison between \textbf{CONCH} and \textbf{CONCH v1.5} remains consistently high across subcohorts and magnifications, indicating strong preservation of representational geometry between closely related model versions. Overall, these results show that encoder agreement is jointly influenced by model family, tissue subcohort, input resolution, and magnification, rather than being a fixed property of an encoder pair.
{
    \small
    \bibliographystyle{ieeenat_fullname}
    \bibliography{main}
}

\end{document}

%%%%%%%%%%%%%%%%%%%%%%%%%%%%%%%%%%%%%%%%%%%%%%%%%%%%%%%%%%%%%%%%%%%%%%%%%%%%%%
% Supplementary Title
%%%%%%%%%%%%%%%%%%%%%%%%%%%%%%%%%%%%%%%%%%%%%%%%%%%%%%%%%%%%%%%%%%%%%%%%%%%%%%

\maketitlesupplementary

%%%%%%%%%%%%%%%%%%%%%%%%%%%%%%%%%%%%%%%%%%%%%%%%%%%%%%%%%%%%%%%%%%%%%%%%%%%%%%
% Numbering
%%%%%%%%%%%%%%%%%%%%%%%%%%%%%%%%%%%%%%%%%%%%%%%%%%%%%%%%%%%%%%%%%%%%%%%%%%%%%%

\appendix

\renewcommand{\thesection}{\Alph{section}}
\renewcommand{\thesubsection}{\thesection.\arabic{subsection}}

\setcounter{figure}{0}
\setcounter{table}{0}
\setcounter{equation}{0}

\renewcommand{\thefigure}{S\arabic{figure}}
\renewcommand{\thetable}{S\arabic{table}}
\renewcommand{\theequation}{S\arabic{equation}}

\section{Discussion on PFMs used in MOSAIC}

\subsection{Patch-Level Encoders}
To ensure that each encoder is evaluated under the input conditions for which it was originally designed, we retain the \textbf{native input resolution} of PFM rather than forcing all models to a common patch size. Resizing inputs across substantially different native resolutions can alter fine-grained morphology, introduce interpolation artifacts, and confound representational comparisons with preprocessing effects. We therefore group encoders according to their native spatial resolution and conduct comparisons within these resolution-matched families. 

\begin{table}[!h]
\centering
\caption{\textbf{Patch encoder families used in MOSAIC-Bench.}
Encoders are grouped by their native input resolution, together with the number of models and the distribution of pretraining objectives within each group.}
\label{tab:encoder_families}
\scriptsize
\setlength{\tabcolsep}{4pt}
\renewcommand{\arraystretch}{1.15}

\begin{tabularx}{\linewidth}{
>{\bfseries}l
>{\raggedright\arraybackslash}X
c
>{\raggedright\arraybackslash}p{4.2cm}
}
\toprule
\textbf{Input Size} &
\textbf{Encoders} &
\textbf{$n$} &
\textbf{Pretraining Objectives} \\
\midrule

\rowcolor{gray!8}
256 px &
CONCH, CTransPath, Prov-GigaPath, KEEP, ResNet50, UNI2-H &
6 &
2 Vision--Language (VL), 3 Self-Supervised Learning (SSL), 1 Supervised \\

224 px &
GPFM, H-optimus-0, Virchow, Virchow2 &
4 &
4 Self-Supervised Learning (SSL) \\

\rowcolor{gray!8}
512 px &
CONCH, CONCH v1.5 &
2 &
2 Vision--Language (VL) \\

384 px &
MUSK &
1 &
1 Vision--Language (VL) \\

\bottomrule
\end{tabularx}
\end{table}

Table~\ref{tab:encoder_families} summarizes the resulting encoder groups, together with the number of models and the distribution of pretraining objectives represented in each group.

\subsection{Slide-Level Encoder}
To evaluate slide-level encoders under their intended operating conditions, we preserve the patch-feature pipeline associated with each model rather than supplying a common patch representation to all slide encoders. Each slide encoder therefore receives features extracted by the patch encoder used in its original or TRIDENT-supported implementation, together with the corresponding patch size and magnification. Table~\ref{tab:slide_patch_encoders} summarizes these slide-to-patch encoder mappings, ensuring that differences in slide-level representations are not generated by mismatched upstream feature extractors.

\begin{table}[t]
\centering
\caption{\textbf{Patch features used by the slide-level encoders.}
Each slide encoder is evaluated using the patch encoder and extraction
configuration specified by its corresponding TRIDENT implementation.}
\label{tab:slide_patch_encoders}
\scriptsize
\setlength{\tabcolsep}{4pt}
\renewcommand{\arraystretch}{1.1}

\begin{tabular}{llll}
\toprule
\textbf{Slide Encoder} &
\textbf{Patch Encoder} &
\textbf{Patch Size} &
\textbf{Magnification} \\
\midrule

CHIEF~\cite{wang2024pathology}
& CTransPath~\cite{wang2022transformer}
& 256 px & 10$\times$ \\

Madeleine~\cite{jaume2024multistain}
& CONCH~\cite{lu2024visual}
& 256 px & 10$\times$ \\

PRISM~\cite{shaikovski2024prism}
& Virchow~\cite{vorontsov2023virchow}
& 224 px & 20$\times$ \\

GigaPath~\cite{xu2024whole}
& Prov-GigaPath~\cite{xu2024whole}
& 256 px & 20$\times$ \\

TITAN~\cite{ding2025multimodal}
& CONCH v1.5~\cite{lu2024visual}
& 512 px & 20$\times$ \\

Feather~\cite{shao2025multiple}
& CONCH v1.5~\cite{lu2024visual}
& 512 px & 20$\times$ \\

\bottomrule
\end{tabular}
\end{table}

\subsection{WSI Processing with TRIDENT}

We use TRIDENT~\cite{zhang2025accelerating} as the unified preprocessing and feature-extraction framework for MOSAIC. TRIDENT provides a common pipeline for tissue segmentation, patch coordinate generation, patch-level feature extraction and slide-level embedding generation, allowing all encoders to be evaluated under a consistent WSI processing workflow.

\begin{table*}[!t]
\centering

\caption{
\textbf{Foundation models evaluated in MOSAIC.}
Summary of the patch-level and slide-level encoders used in our study,
including their pretraining data, architecture, pretraining objective,
representation dimensionality, and publicly available resources.
ResNet50 is included as an ImageNet-supervised reference model.
}

\label{tab:mosaic_foundation_models}

\scriptsize
\setlength{\tabcolsep}{2.0pt}
\renewcommand{\arraystretch}{1.25}

\begin{tabularx}{\textwidth}{
    l
    c
    >{\raggedright\arraybackslash}X
    >{\raggedright\arraybackslash}X
    >{\raggedright\arraybackslash}X
    c
    p{1.45cm}
}

\toprule

\textbf{Model} &
\textbf{Year} &
\textbf{Pretraining Data Size} &
\textbf{Architecture} &
\textbf{Pretraining Objective} &
\textbf{Dim.} &
\textbf{Resources}
\\

\midrule

% ============================================================
% PATCH-LEVEL ENCODERS
% ============================================================

\rowcolor{catgray}
\multicolumn{7}{c}{
\textbf{\textit{Patch-level Encoders}}
}
\\

\midrule

UNI2-H~\cite{chen2024towards}
& 2025
& 200M tiles / 350K WSIs
& ViT-H/14, 8 Register Tokens
& DINO self-distillation + iBOT + KoLeo
& 1536
&
\href{https://github.com/mahmoodlab/UNI}
{\includegraphics[height=1.05em]{ 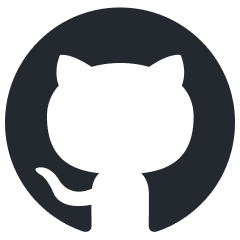}}
\href{https://huggingface.co/MahmoodLab/UNI2-h}
{\includegraphics[height=1.25em]{ 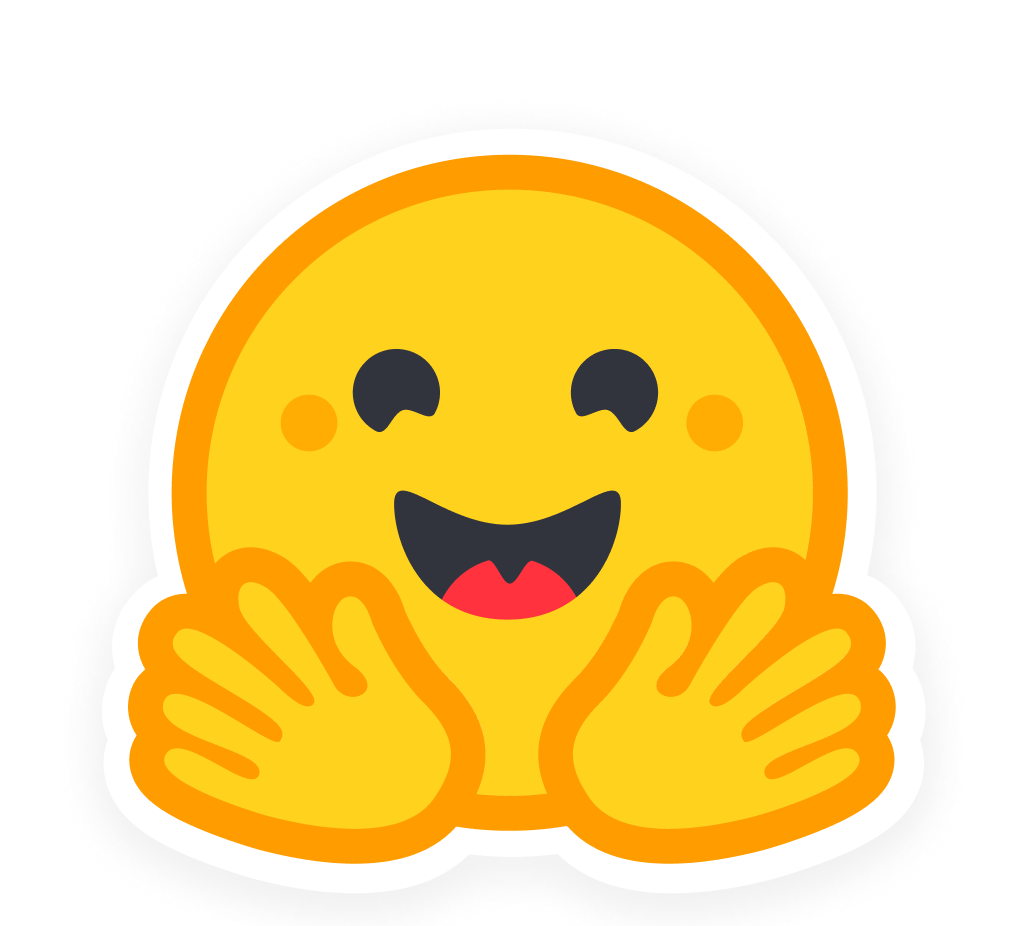}}
\href{https://doi.org/10.1038/s41591-024-02857-3}
{\includegraphics[height=1.05em]{ 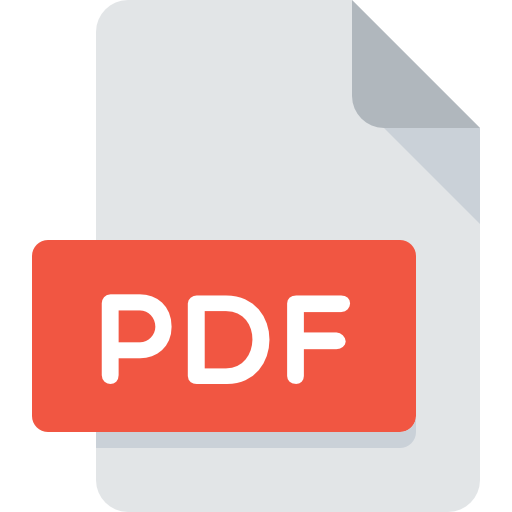}}
\\

\addlinespace[2pt]

Prov-GigaPath~\cite{xu2024whole}
& 2024
& 1.3B tiles / 171K WSIs
& ViT-G/14
& DINOv2 self-supervision
& 1536
&
\href{https://github.com/prov-gigapath/prov-gigapath}
{\includegraphics[height=1.05em]{ github-mark.png}}
\href{https://huggingface.co/prov-gigapath/prov-gigapath}
{\includegraphics[height=1.25em]{ hf-logo.png}}
\href{https://www.nature.com/articles/s41586-024-07441-w}
{\includegraphics[height=1.05em]{ pdf.png}}
\\

\addlinespace[2pt]

Virchow~\cite{vorontsov2023virchow}
& 2024
& 1.5M WSIs / 100K patients
& ViT-H/14
& DINOv2
& 2560
&
\href{https://huggingface.co/paige-ai/Virchow}
{\includegraphics[height=1.25em]{ hf-logo.png}}
\href{https://doi.org/10.1038/s41591-024-03141-0}
{\includegraphics[height=1.05em]{ pdf.png}}
\\

\addlinespace[2pt]

Virchow2~\cite{zimmermann2024virchow2}
& 2024
& 3.1M WSIs
& ViT-H/14, 4 Register Tokens
& Modified DINOv2 + KDE regularization
& 2560
&
\href{https://huggingface.co/paige-ai/Virchow2}
{\includegraphics[height=1.25em]{ hf-logo.png}}
\href{https://arxiv.org/abs/2408.00738}
{\includegraphics[height=1.05em]{ pdf.png}}
\\

\addlinespace[2pt]

H-optimus-0~\cite{filiot2025distilling}
& 2024
& 500K WSIs
& ViT-G/14, 4 Register Tokens
& DINOv2-style self-supervision
& 1536
&
\href{https://github.com/bioptimus/releases/tree/main/models/h-optimus/v0}
{\includegraphics[height=1.05em]{ github-mark.png}}
\href{https://huggingface.co/bioptimus/H-optimus-0}
{\includegraphics[height=1.25em]{ hf-logo.png}}
\\

\addlinespace[2pt]

GPFM~\cite{ma2024towards}
& 2025
& 190M tiles / 72K WSIs
& ViT-L/14
& Multi-teacher knowledge distillation
& 1024
&
\href{https://github.com/birkhoffkiki/GPFM}
{\includegraphics[height=1.05em]{ github-mark.png}}
\href{https://huggingface.co/majiabo/GPFM}
{\includegraphics[height=1.25em]{ hf-logo.png}}
\href{https://doi.org/10.1038/s41551-025-01488-4}
{\includegraphics[height=1.05em]{ pdf.png}}
\\

\addlinespace[2pt]

CTransPath~\cite{wang2022transformer}
& 2022
& 35K WSIs
& CNN Stem + Swin Transformer
& Semantically Relevant Contrastive Learning (SRCL)
& 768
&
\href{https://github.com/Xiyue-Wang/TransPath}
{\includegraphics[height=1.05em]{ github-mark.png}}
\href{https://huggingface.co/1aurent/swin_tiny_patch4_window7_224.CTransPath}
{\includegraphics[height=1.25em]{ hf-logo.png}}
\href{https://www.sciencedirect.com/science/article/pii/S1361841522002043}
{\includegraphics[height=1.05em]{ pdf.png}}
\\

\addlinespace[2pt]

CONCH~\cite{lu2024visual}
& 2024
& 1.17M Image–Text Pairs
& ViT-B/16 + Text Transformer
& Unimodal pretraining + CoCa
& 512
&
\href{https://github.com/mahmoodlab/CONCH}
{\includegraphics[height=1.05em]{ github-mark.png}}
\href{https://huggingface.co/MahmoodLab/CONCH}
{\includegraphics[height=1.25em]{ hf-logo.png}}
\href{https://doi.org/10.1038/s41591-024-02856-4}
{\includegraphics[height=1.05em]{ pdf.png}}
\\

\addlinespace[2pt]

CONCH v1.5~\cite{lu2024visual}
& 2024
& 1.26M Image–Text Pairs
& ViT-L/16
& CoCa vision--language pretraining
& 768
&
\href{https://huggingface.co/MahmoodLab/conchv1_5}
{\includegraphics[height=1.25em]{ hf-logo.png}}
\href{https://doi.org/10.1038/s41591-025-03982-3}
{\includegraphics[height=1.05em]{ pdf.png}}
\\

\addlinespace[2pt]

KEEP~\cite{zhou2026knowledge}
& 2026
& 143K Semantic Groups
& ViT-L/16 + PubMedBERT
& Knowledge-enhanced vision-language alignment
& 768
&
\href{https://github.com/MAGIC-AI4Med/KEEP}
{\includegraphics[height=1.05em]{ github-mark.png}}
\href{https://huggingface.co/Astaxanthin/KEEP}
{\includegraphics[height=1.25em]{ hf-logo.png}}
\href{https://arxiv.org/abs/2412.13126}
{\includegraphics[height=1.05em]{ pdf.png}}
\\

\addlinespace[2pt]

MUSK~\cite{xiang2025vision}
& 2025
& 50M images / 1B text tokens
& BEiT-3 Multiway Transformer
& MIM + MLM + image--text contrastive learning
& 1024
&
\href{https://github.com/lilab-stanford/MUSK}
{\includegraphics[height=1.05em]{ github-mark.png}}
\href{https://huggingface.co/xiangjx/musk}
{\includegraphics[height=1.25em]{ hf-logo.png}}
\href{https://doi.org/10.1038/s41586-024-08378-w}
{\includegraphics[height=1.05em]{ pdf.png}}
\\

\addlinespace[2pt]

ResNet50~\cite{he2016deep}
& 2016
& 1.28M ImageNet Images
& ResNet-50 CNN
& Supervised ImageNet classification
& 1024
&
\href{https://github.com/pytorch/vision}
{\includegraphics[height=1.05em]{ github-mark.png}}
\href{https://openaccess.thecvf.com/content_cvpr_2016/html/He_Deep_Residual_Learning_CVPR_2016_paper.html}
{\includegraphics[height=1.05em]{ pdf.png}}
\\

% ============================================================
% SEPARATOR
% ============================================================

\addlinespace[5pt]
\midrule
\addlinespace[2pt]

% ============================================================
% SLIDE-LEVEL ENCODERS
% ============================================================

\rowcolor{catgray}
\multicolumn{7}{c}{
\textbf{\textit{Slide-level Encoders}}
}
\\

\midrule

CHIEF~\cite{wang2024pathology}
& 2024
& 61K WSIs
& CTransPath + Attention-Based MIL
& Tile SSL + weakly supervised slide learning
& 768
&
\href{https://github.com/hms-dbmi/CHIEF}
{\includegraphics[height=1.05em]{ github-mark.png}}
\href{https://www.nature.com/articles/s41586-024-07894-z}
{\includegraphics[height=1.05em]{ pdf.png}}
\\

\addlinespace[2pt]

Madeleine~\cite{jaume2024multistain}
& 2024
& 4K Breast WSIs / 5 Stains
& CONCH + Multi-Head Attention MIL
& Global-local cross-stain alignment
& 512
&
\href{https://github.com/mahmoodlab/MADELEINE}
{\includegraphics[height=1.05em]{ github-mark.png}}
\href{https://huggingface.co/MahmoodLab/madeleine}
{\includegraphics[height=1.25em]{ hf-logo.png}}
\href{https://arxiv.org/abs/2408.02859}
{\includegraphics[height=1.05em]{ pdf.png}}
\\

\addlinespace[2pt]

PRISM~\cite{shaikovski2024prism}
& 2024
& 587K WSIs + 195K Reports
& Virchow + Perceiver + BioGPT
& Slide-text contrastive alignment + captioning
& 1280
&
\href{https://huggingface.co/paige-ai/Prism}
{\includegraphics[height=1.25em]{ hf-logo.png}}
\href{https://arxiv.org/abs/2405.10254}
{\includegraphics[height=1.05em]{ pdf.png}}
\\

\addlinespace[2pt]

GigaPath~\cite{xu2024whole}
& 2024
& 1.3B tiles / 171K WSIs
& ViT-G/14 + LongNet
& DINOv2 Tile SSL + Slide-Level Masked Autoencoding
& 768
&
\href{https://github.com/prov-gigapath/prov-gigapath}
{\includegraphics[height=1.05em]{ github-mark.png}}
\href{https://huggingface.co/prov-gigapath/prov-gigapath}
{\includegraphics[height=1.25em]{ hf-logo.png}}
\href{https://www.nature.com/articles/s41586-024-07441-w}
{\includegraphics[height=1.05em]{ pdf.png}}
\\

\addlinespace[2pt]

TITAN~\cite{ding2025multimodal}
& 2025
& 336K WSIs / 183K Reports
& CONCH v1.5 + ViT Slide Encoder
& iBOT SSL + CoCa vision--language alignment
& 768
&
\href{https://github.com/mahmoodlab/TITAN}
{\includegraphics[height=1.05em]{ github-mark.png}}
\href{https://huggingface.co/MahmoodLab/TITAN}
{\includegraphics[height=1.25em]{ hf-logo.png}}
\href{https://doi.org/10.1038/s41591-025-03982-3}
{\includegraphics[height=1.05em]{ pdf.png}}
\\

\addlinespace[2pt]

Feather~\cite{shao2025multiple}
& 2025
& 24K WSIs
& CONCH v1.5 + ABMIL
& Supervised 108-class pan-cancer classification
& 512
&
\href{https://github.com/mahmoodlab/MIL-Lab}
{\includegraphics[height=1.05em]{ github-mark.png}}
\href{https://huggingface.co/MahmoodLab/abmil.base.conch_v15.pc108-24k}
{\includegraphics[height=1.25em]{ hf-logo.png}}
\href{https://arxiv.org/abs/2506.09022}
{\includegraphics[height=1.05em]{ pdf.png}}
\\

\bottomrule

\end{tabularx}
\end{table*}

\subsubsection{Tissue Segmentation and Patch Extraction}

WSIs are first segmented to identify tissue-containing regions and exclude background. Patch coordinates are then generated only within the segmented tissue regions. For experiments spanning multiple magnifications, TRIDENT generates an independent coordinate grid for each magnification; consequently, patches extracted at different magnifications are treated as separate samples rather than spatially paired observations. Patch size is selected according to the native input resolution of the corresponding encoder, avoiding unnecessary resizing of the model input. TRIDENT stores the resulting coordinates and associated slide metadata for subsequent feature extraction.

\subsubsection{Patch-Level Feature Extraction}

Patch representations are extracted directly from the generated tissue coordinates using TRIDENT's encoder interface. Each foundation model is evaluated at its native input resolution, while magnification and patch-size configurations are kept consistent between coordinate generation and feature extraction.

\subsubsection{Slide-Level Feature Extraction}

For slide-level foundation models, we use TRIDENT's slide-encoder pipeline, which couples each slide encoder with its required upstream patch encoder. When the corresponding patch embeddings are not already available, TRIDENT extracts them automatically before aggregating them into a single slide representation. We retain these model-specific patch-to-slide configurations rather than replacing them with a common patch encoder, thereby preserving the feature distribution expected by each pretrained slide model. The patch encoder, patch size, and magnification associated with each slide encoder are summarized in Table~\ref{tab:slide_patch_encoders}. 

\subsection{Summary of Foundation Models Used in MOSAIC}

Table~\ref{tab:mosaic_foundation_models} summarizes the foundation models evaluated in MOSAIC, comprising 12 patch-level encoders and 6 slide-level encoders. Together, these models span a broad range of architectural families, pretraining paradigms, dataset scales, and representation dimensionalities, providing a heterogeneous collection for investigating representational similarity and cross-model alignment. The patch-level encoders include transformer-based architectures such as ViT-H, ViT-G, ViT-L, and ViT-B, as well as Swin Transformer, BEiT-3, and the CNN-based ResNet50 reference model. Their output dimensionalities range from 512 to 2560, allowing MOSAIC to examine whether representational correspondence persists despite substantial differences in feature-space dimensionality and model capacity. 

The patch encoders also differ considerably in their pretraining objectives and data sources. Models such as UNI2-H, Prov-GigaPath, Virchow, Virchow2, and H-optimus-0 rely primarily on self-supervised objectives derived from DINO/DINOv2-style training, while GPFM employs multi-teacher knowledge distillation. In contrast, CONCH, CONCH v1.5, KEEP, and MUSK incorporate vision--language or multimodal supervision, using image--text pairs, pathology-specific textual knowledge, or joint image--text objectives. Their pretraining scales range from tens or hundreds of thousands of WSIs to hundreds of millions or billions of image tiles, reflecting substantial variation in both the quantity and type of supervision available during representation learning. ResNet50 is included as an ImageNet-supervised reference.

The six slide-level encoders introduce an additional source of heterogeneity because they differ not only in their pretrained representations, but also in how patch-level information is aggregated into a whole-slide representation. CHIEF and Madeleine employ attention-based MIL-style aggregation, GigaPath uses a LongNet-based slide encoder, PRISM incorporates Perceiver-based aggregation together with multimodal supervision, TITAN combines CONCH v1.5 features with a ViT-based slide encoder, and Feather uses an ABMIL-based aggregation strategy. 

\section{Downstream Task Definitions}

We evaluate 13 downstream tasks spanning histologic subtyping, pathologic staging, and molecular mutation prediction. Morphological and clinical tasks assess information that is more directly reflected in tissue organization, whereas molecular tasks test whether representations capture subtler genotype-associated morphological patterns.

\subsection{TCGA-NSCLC: LUAD vs.\ LUSC Subtyping}

This task distinguishes \textbf{lung adenocarcinoma (LUAD)} from \textbf{lung squamous cell carcinoma (LUSC)}, the two major histologic subtypes of non-small cell lung cancer (NSCLC). The task evaluates whether learned representations capture characteristic differences in tumor morphology and tissue organization between the two subtypes.

\subsection{TCGA-BRCA: IDC vs.\ ILC Subtyping}

This task distinguishes \textbf{invasive ductal carcinoma (IDC)} from \textbf{invasive lobular carcinoma (ILC)}. IDC typically exhibits cohesive tumor-cell organization, whereas ILC is commonly characterized by more dispersed and single-file infiltration patterns. The task therefore provides a morphology-driven assessment of representation quality.

\subsection{TCGA-NSCLC: Pathologic Staging}

This task predicts pathologic disease stage in non-small cell lung cancer from H\&E whole-slide images. Unlike histologic subtyping, staging reflects the extent of disease progression and therefore tests whether representations capture morphological characteristics associated with tumor advancement.

\subsection{TCGA-BRCA: Pathologic Staging}

This task predicts pathologic stage in breast cancer. It evaluates whether slide representations contain information associated with disease progression in addition to histologic subtype and local tumor morphology.

\subsection{CPTAC-LUAD: \textit{TP53} Mutation Prediction}

This task predicts whether a lung adenocarcinoma harbors a mutation in \textit{TP53}. \textit{TP53} is a major tumor-suppressor gene involved in DNA-damage response, cell-cycle regulation, and apoptosis. The task evaluates whether genotype-associated morphological patterns are encoded in H\&E representations.

\subsection{CPTAC-LUAD: \textit{KRAS} Mutation Prediction}

This task predicts \textit{KRAS} mutation status in lung adenocarcinoma. \textit{KRAS} is an oncogene in the RAS--MAPK signaling pathway, and activating mutations are common molecular drivers of LUAD. Prediction from morphology tests whether the learned representation captures phenotypic consequences associated with this alteration.

\subsection{CPTAC-LUAD: \textit{STK11} Mutation Prediction}

This task predicts mutation status of \textit{STK11}, a tumor-suppressor kinase involved in cellular metabolism and growth regulation. \textit{STK11} alterations define an important molecular subgroup of lung adenocarcinoma and provide a comparatively subtle molecular prediction target from histology.

\subsection{CPTAC-BRCA: \textit{PIK3CA} Mutation Prediction}

This task predicts \textit{PIK3CA} mutation status in breast cancer. \textit{PIK3CA} encodes the catalytic subunit of phosphoinositide 3-kinase and is among the most frequently altered oncogenes in breast cancer. The task assesses whether representations capture morphology associated with PI3K-pathway alterations.

\subsection{CPTAC-BRCA: \textit{GATA3} Mutation Prediction}

This task predicts mutation status of \textit{GATA3}, a transcription factor involved in mammary epithelial differentiation. \textit{GATA3} mutations occur recurrently in breast cancer and are associated with distinct tumor biology, making them a useful target for evaluating molecular information encoded in morphology.

\subsection{CPTAC-BRCA: \textit{MAP3K1} Mutation Prediction}

This task predicts mutations in \textit{MAP3K1}, a kinase involved in MAPK signaling and cellular stress responses. Recurrent \textit{MAP3K1} alterations occur in breast cancer and provide an additional molecular target for assessing the sensitivity of learned representations to genotype-associated tissue phenotypes.

\subsection{CPTAC-COAD: \textit{TP53} Mutation Prediction}

This task predicts \textit{TP53} mutation status in colon adenocarcinoma. Because \textit{TP53} alterations are common during colorectal tumor progression, this task evaluates whether representations capture morphological patterns associated with disruption of this major tumor-suppressor pathway.

\subsection{CPTAC-COAD: \textit{KRAS} Mutation Prediction}

This task predicts \textit{KRAS} mutation status in colon adenocarcinoma. Activating \textit{KRAS} mutations are frequent drivers of colorectal cancer and influence MAPK signaling. The task tests whether these molecular differences produce reproducible signals within learned histopathology representations.

\subsection{CPTAC-COAD: \textit{PIK3CA} Mutation Prediction}

This task predicts \textit{PIK3CA} mutation status in colon adenocarcinoma. Alterations in \textit{PIK3CA} activate the PI3K signaling pathway and define a molecular subset of colorectal cancers. This task provides an additional test of whether aligned representations preserve subtle molecular information beyond visually apparent histologic characteristics.

\section{Representation Similarity Metrics}
\label{sec:supp_similarity}

Let
\(\mathbf{X}\in\mathbb{R}^{N\times d_x}\) and
\(\mathbf{Y}\in\mathbb{R}^{N\times d_y}\)
denote representations produced by two encoders for the same \(N\)
matched samples, where \(d_x\) and \(d_y\) may differ. Representations
are centered across samples before comparison. We employ seven
complementary metrics that capture different notions of similarity,
including global geometry, correlated subspaces, pairwise relationships
and nonlinear statistical dependence.

\paragraph{\textbf{Linear CKA.}}
Centered Kernel Alignment (CKA) measures agreement between the
sample-wise geometry induced by two representations and is invariant to
orthogonal transformations and isotropic scaling
~\citep{kornblith2019similarity}. For centered feature matrices, linear
CKA can be computed directly as:

\begin{equation}
S_{\operatorname{CKA}}^{\mathrm{lin}}(\mathbf{X},\mathbf{Y})
=
\frac{
\|\mathbf{X}^{\top}\mathbf{Y}\|_F^2
}{
\|\mathbf{X}^{\top}\mathbf{X}\|_F
\|\mathbf{Y}^{\top}\mathbf{Y}\|_F
}
\end{equation}

Higher values indicate stronger agreement in the linear geometry of the
two representation spaces. Here, $\|\cdot\|_F$ denotes the Frobenius norm; which measures the overall magnitude of a matrix as the square root of the sum of the squared magnitudes of all its entries.

\paragraph{\textbf{Kernel CKA.}}
To capture nonlinear correspondence between two representations
$\mathbf{X}\in\mathbb{R}^{N\times d_x}$ and
$\mathbf{Y}\in\mathbb{R}^{N\times d_y}$, we additionally compute CKA
using radial basis function (RBF) kernels. Here, $N$ denotes the number
of shared samples, and $\mathbf{x}_a$ and $\mathbf{y}_a$ denote the
feature vectors of sample $a$ in the two representation spaces,
respectively. For $\mathbf{X}$, the RBF Gram matrix
$\mathbf{K}\in\mathbb{R}^{N\times N}$ is defined entry-wise as

\begin{equation}
K_{ab}
=
\exp\left(
-\frac{\|\mathbf{x}_a-\mathbf{x}_b\|_2^2}
{2\sigma_X^2}
\right),
\qquad a,b\in\{1,\ldots,N\}
\end{equation}

where $\sigma_X$ is the RBF bandwidth for representation $\mathbf{X}$.
An analogous Gram matrix $\mathbf{L}$ is constructed for
$\mathbf{Y}$ using bandwidth $\sigma_Y$. The bandwidths
$\sigma_X$ and $\sigma_Y$ are determined independently using the
median-distance heuristic.

To remove mean effects in the kernel space, we define the centering
matrix

\begin{equation}
\mathbf{H}
=
\mathbf{I}_N
-
\frac{1}{N}\mathbf{1}\mathbf{1}^{\top}
\end{equation}

where $\mathbf{I}_N$ is the $N\times N$ identity matrix and
$\mathbf{1}\in\mathbb{R}^{N}$ is the all-ones vector. The centered
Gram matrices are then

\begin{equation}
\widetilde{\mathbf{K}}
=
\mathbf{H}\mathbf{K}\mathbf{H},
\qquad
\widetilde{\mathbf{L}}
=
\mathbf{H}\mathbf{L}\mathbf{H}
\end{equation}

Kernel CKA is defined as follows:

\begin{equation}
S_{\operatorname{CKA}}^{\mathrm{ker}}(\mathbf{X},\mathbf{Y})
=
\frac{
\left\langle
\widetilde{\mathbf{K}},
\widetilde{\mathbf{L}}
\right\rangle_F
}{
\|\widetilde{\mathbf{K}}\|_F
\|\widetilde{\mathbf{L}}\|_F
}
\end{equation}

where $\langle\cdot,\cdot\rangle_F$ denotes the Frobenius inner product
and $\|\cdot\|_F$ denotes the Frobenius norm. Unlike linear CKA, which
compares similarity under a linear kernel, RBF-kernel CKA can capture
nonlinear agreement in the pairwise relationships among samples.

\paragraph{\textbf{SVCCA.}}
Singular Vector Canonical Correlation Analysis (SVCCA) compares the
dominant correlated subspaces of two representations
~\citep{raghu2017svcca}. Each representation is first reduced using SVD,
retaining the smallest number of components that explain \(99\%\) of
the variance. Canonical correlation analysis is then applied to the
reduced representations. If
\(\rho_1,\ldots,\rho_K\) denote the resulting canonical correlations,
the SVCCA similarity is defined as follows:

\begin{equation}
S_{\mathrm{SVCCA}}
=
\frac{1}{K}
\sum_{k=1}^{K}\rho_k
\end{equation}

SVCCA therefore focuses on the dominant shared linear structure while
suppressing low-variance directions.

\begin{table*}[t]
\centering
\caption{\textbf{Representation similarity metrics used in MOSAIC.}
Each metric probes a complementary aspect of similarity between two
representation spaces.}
\label{tab:representation_similarity_questions}
\small
\setlength{\tabcolsep}{5pt}
\renewcommand{\arraystretch}{1.18}

\begin{tabular}{
p{0.18\linewidth}
p{0.08\linewidth}
p{0.35\linewidth}
p{0.32\linewidth}
}
\toprule
\textbf{Metric}
& \textbf{Symbol}
& \textbf{Question}
& \textbf{Interpretation} \\
\midrule

\textbf{Linear CKA}
& $S_{\operatorname{CKA}}^{\mathrm{lin}}$
& Do the two encoders organize samples similarly under
\textbf{linear relationships}?
& $\uparrow$ \textbf{Higher:} more similar linear geometry
\newline
$\downarrow$ \textbf{Lower:} more distinct linear geometry \\

\textbf{Kernel CKA}
& $S_{\operatorname{CKA}}^{\mathrm{ker}}$
& Do the two encoders organize samples similarly when
\textbf{nonlinear relationships} are considered?
& $\uparrow$ \textbf{Higher:} more similar nonlinear geometry
\newline
$\downarrow$ \textbf{Lower:} more distinct nonlinear geometry \\

\textbf{SVCCA}
& $S_{\mathrm{SVCCA}}$
& Do the two encoders share
\textbf{similar dominant representational subspaces}?
& $\uparrow$ \textbf{Higher:} greater shared subspace structure
\newline
$\downarrow$ \textbf{Lower:} more distinct dominant subspaces \\

\textbf{PWCCA}
& $S_{\mathrm{PWCCA}}$
& How \textbf{similar are the shared directions} after weighting
them by their contribution to the original representations?
& $\uparrow$ \textbf{Higher:} stronger alignment of important directions
\newline
$\downarrow$ \textbf{Lower:} weaker alignment of important directions \\

\textbf{Procrustes}
& $S_{\mathrm{Proc}}$
& Can one representation be
\textbf{geometrically aligned} with the other through an optimal
orthogonal transformation?
& $\uparrow$ \textbf{Higher:} greater geometric alignability
\newline
$\downarrow$ \textbf{Lower:} greater geometric discrepancy \\

\textbf{Distance Correlation}
& $S_{\mathrm{dCor}}$
& Do the two representations exhibit
\textbf{statistical dependence} in their sample geometry, including
nonlinear dependence?
& $\uparrow$ \textbf{Higher:} stronger statistical dependence
\newline
$\downarrow$ \textbf{Lower:} weaker statistical dependence \\

\textbf{Cosine RSA}
& $S_{\mathrm{RSA}}^{\mathrm{cos}}$
& Do the two encoders preserve
\textbf{similar pairwise cosine relationships} among samples?
& $\uparrow$ \textbf{Higher:} greater relational agreement
\newline
$\downarrow$ \textbf{Lower:} weaker relational agreement \\

\bottomrule
\end{tabular}
\end{table*}

\paragraph{\textbf{PWCCA.}}
Projection-Weighted CCA (PWCCA) also operates on canonical correlations,
but weights each direction according to its contribution to the
original representation~\citep{morcos2018insights}. Given normalized
weights \(\alpha_k\), it is defined as follows:

\begin{equation}
S_{\mathrm{PWCCA}}
=
\sum_{k=1}^{K}\alpha_k\rho_k,
\qquad
\sum_{k=1}^{K}\alpha_k=1
\end{equation}

This reduces the influence of canonical directions that explain little
of the original feature space. Since the weighting is directional, we
compute PWCCA in both directions and report their average to obtain a
symmetric encoder-to-encoder similarity.

\paragraph{\textbf{Procrustes Similarity.}}
Procrustes similarity measures whether two representations exhibit
similar geometry up to an orthogonal transformation and global scaling
~\citep{gower1975generalized}. We use the normalized nuclear norm of
their cross-covariance:

\begin{equation}
S_{\mathrm{Proc}}
=
\frac{
\|\mathbf{X}^{\top}\mathbf{Y}\|_*
}{
\|\mathbf{X}\|_F
\|\mathbf{Y}\|_F
}
\end{equation}

where \(\|\cdot\|_*\) denotes the nuclear norm. A high score indicates
that the two spaces can be closely related through a rigid change of
basis, making Procrustes a stricter notion of geometric correspondence
than CCA-based measures.

\paragraph{\textbf{Cosine RSA.}}
Representational Similarity Analysis (RSA) compares relationships among
samples rather than directly comparing feature coordinates
~\citep{kriegeskorte2008representational}. For each representation,
we construct a cosine representational similarity matrix. For
$\mathbf{X}$, each entry is

\begin{equation}
R^{X}_{ab}
=
\frac{
\mathbf{x}_a^{\top}\mathbf{x}_b
}{
\|\mathbf{x}_a\|_2
\|\mathbf{x}_b\|_2
}
\end{equation}

where $\mathbf{x}_a$ and $\mathbf{x}_b$ denote the feature vectors of
samples $a$ and $b$, respectively. An analogous matrix
$\mathbf{R}^{Y}$ is constructed for $\mathbf{Y}$.

We then vectorize the upper-triangular entries of the two similarity
matrices, excluding the diagonal, and define the cosine RSA similarity
as follows:

\begin{equation}
S_{\mathrm{RSA}}^{\mathrm{cos}}(\mathbf{X},\mathbf{Y})
=
\rho_{\mathrm{Spearman}}
\left(
\operatorname{vec}_{\triangle}(\mathbf{R}^{X}),
\operatorname{vec}_{\triangle}(\mathbf{R}^{Y})
\right)
\end{equation}

where $\operatorname{vec}_{\triangle}(\cdot)$ extracts the
upper-triangular entries excluding the diagonal, and
$\rho_{\mathrm{Spearman}}$ denotes Spearman's rank correlation.
Cosine RSA is therefore high when two encoders preserve a similar
ordering of pairwise relationships among samples, even when their
feature coordinates differ substantially.

\paragraph{\textbf{Distance Correlation.}}
Distance correlation measures general statistical dependence between
two representations and does not require equal feature dimensionality
or a linear relationship~\citep{szekely2007measuring}. We first compute
the pairwise Euclidean distance matrices for
$\mathbf{X}$ and $\mathbf{Y}$ and apply U-centering to obtain
$\mathbf{A}$ and $\mathbf{B}$, respectively. The unbiased squared
distance covariance is defined as follows:

\begin{equation}
\operatorname{dCov}^{2}(\mathbf{X},\mathbf{Y})
=
\frac{1}{N(N-3)}
\sum_{a=1}^{N}\sum_{b=1}^{N}
A_{ab}B_{ab}
\end{equation}

The corresponding squared distance correlation is obtained by
normalizing the distance covariance by the distance variances of the
two representations:

\begin{equation}
\operatorname{dCor}^{2}(\mathbf{X},\mathbf{Y})
=
\frac{
\operatorname{dCov}^{2}(\mathbf{X},\mathbf{Y})
}{
\sqrt{
\operatorname{dCov}^{2}(\mathbf{X},\mathbf{X})
\operatorname{dCov}^{2}(\mathbf{Y},\mathbf{Y})
}
}
\end{equation}

We define the resulting representation-similarity score as

\begin{equation}
S_{\mathrm{dCor}}(\mathbf{X},\mathbf{Y})
=
\sqrt{
\max\!\left(
\operatorname{dCor}^{2}(\mathbf{X},\mathbf{Y}),\,0
\right)
}
\end{equation}

A larger $S_{\mathrm{dCor}}$ indicates stronger statistical dependence
between the two representation spaces. Unlike measures restricted to
linear correspondence, distance correlation can capture more general
dependencies in their pairwise distance structure.

\paragraph{\textbf{Pairwise Similarity Matrices.}}
For each similarity metric $m$, we compute the pairwise representation
similarity

\begin{equation}
S_{ij}^{(m)}
=
S_m\!\left(
\mathbf{X}^{(i)},
\mathbf{X}^{(j)}
\right)
\end{equation}

where $\mathbf{X}^{(i)}$ and $\mathbf{X}^{(j)}$ denote the feature
representations produced by encoders $i$ and $j$, respectively, and
$S_m$ denotes the similarity function associated with metric $m$.
Evaluating all encoder pairs produces an encoder-by-encoder similarity
matrix which is represented as follows:

\begin{equation}
\mathbf{S}^{(m)}
=
\left[
S_{ij}^{(m)}
\right]_{i,j=1}^{M}
\end{equation}

where $M$ is the number of encoders. All pairwise comparisons use the
same matched sample indices across encoders, ensuring that differences
between similarity scores reflect representational structure rather
than differences in the sampled tissue instances.

\section{Shared-Space Alignment Methods}
\label{sec:supp_alignment}

Let \(\mathbf{X}^{(i)}\in\mathbb{R}^{N\times d_i}\) denote the
representations from encoder \(f_i\) for the same \(N\) matched samples.
Each alignment method learns a mapping
\(T_i:\mathbb{R}^{d_i}\rightarrow\mathbb{R}^{d}\), producing
\(\mathbf{Z}^{(i)}=T_i(\mathbf{X}^{(i)})\) in a common
\(d\)-dimensional space.

\paragraph{\textbf{Generalized CCA (GCCA).}}
We use the MAX-VAR formulation of Generalized Canonical Correlation
Analysis (GCCA) to learn encoder-specific linear projections
$\mathbf{W}_i$ together with a common latent representation
$\mathbf{G}$ shared across all $M$ encoders
~\citep{kettenring1971canonical,benton2019deep}. Specifically, GCCA
seeks projections that map each encoder representation
$\mathbf{X}^{(i)}$ as close as possible to the same shared space:

\begin{equation}
\min_{\mathbf{G},\{\mathbf{W}_i\}}
\sum_{i=1}^{M}
\left\|
\mathbf{X}^{(i)}\mathbf{W}_i-\mathbf{G}
\right\|_F^2,
\qquad
\mathbf{G}^{\top}\mathbf{G}=\mathbf{I}
\end{equation}

Here, $\mathbf{W}_i$ denotes the projection associated with encoder
$i$, while $\mathbf{G}\in\mathbb{R}^{N\times d}$ represents the
shared $d$-dimensional latent representation. The orthogonality
constraint $\mathbf{G}^{\top}\mathbf{G}=\mathbf{I}$ prevents
degenerate solutions and encourages the shared latent dimensions to be
non-redundant. GCCA therefore identifies a common subspace that
captures correlated structure shared across multiple encoder
representations simultaneously.

This identifies latent directions jointly supported across all encoder
spaces.

\paragraph{\textbf{Multiset CCA (MCCA).}}
We use the SUMCOR formulation of Multiset Canonical Correlation
Analysis (MCCA), which learns encoder-specific linear projections
$\mathbf{W}_i$ by maximizing the total pairwise correlation among all
projected encoder representations~\citep{kettenring1971canonical}.
Specifically, MCCA solves

\begin{equation}
\max_{\{\mathbf{W}_i\}}
\sum_{i<j}
\operatorname{tr}
\left(
\mathbf{W}_i^{\top}
\mathbf{X}^{(i)\top}
\mathbf{X}^{(j)}
\mathbf{W}_j
\right)
\end{equation}

subject to appropriate normalization constraints on the projected
representations. Here, $\mathbf{W}_i$ denotes the linear projection
associated with encoder $i$, such that

\begin{equation}
\mathbf{Z}^{(i)}
=
\mathbf{X}^{(i)}\mathbf{W}_i
\in
\mathbb{R}^{N\times d}
\end{equation}

The cross-product
$\mathbf{X}^{(i)\top}\mathbf{X}^{(j)}$ captures the cross-covariance
structure between encoders $i$ and $j$, while the trace term measures
the alignment between their projected representations. By summing this
quantity over all encoder pairs, SUMCOR MCCA encourages every projected
representation $\mathbf{Z}^{(i)}$ to be maximally correlated with the
others. Unlike GCCA, which explicitly introduces a common latent
representation $\mathbf{G}$, MCCA aligns the encoder spaces indirectly
through their pairwise correlations, yielding a coordinated
$d$-dimensional shared space across all encoders.

\paragraph{\textbf{Generalized Procrustes Alignment (GPA).}}
Because standard Procrustes alignment requires representations with
matched dimensionality, each encoder representation is first reduced
to the target dimension $d$ using PCA, yielding

\begin{equation}
\widetilde{\mathbf{X}}^{(i)}
\in
\mathbb{R}^{N\times d}.
\end{equation}

Generalized Procrustes Alignment (GPA) then learns an encoder-specific
orthogonal transformation $\mathbf{Q}_i$ and global scaling factor
$\alpha_i$ that align each reduced representation to a common consensus
configuration $\mathbf{G}$~\citep{gower1975generalized}. Specifically,
GPA solves the following:

\begin{equation}
\min_{\mathbf{G},\{\alpha_i,\mathbf{Q}_i\}}
\sum_{i=1}^{M}
\left\|
\alpha_i
\widetilde{\mathbf{X}}^{(i)}
\mathbf{Q}_i
-
\mathbf{G}
\right\|_F^2,
\qquad
\mathbf{Q}_i^{\top}\mathbf{Q}_i
=
\mathbf{I}
\end{equation}

Here, $\mathbf{Q}_i\in\mathbb{R}^{d\times d}$ denotes the
encoder-specific orthogonal transformation, corresponding to a
rotation or reflection of the reduced representation, while
$\alpha_i$ adjusts for differences in global scale. The consensus
matrix $\mathbf{G}\in\mathbb{R}^{N\times d}$ represents the common
geometric configuration shared across the aligned encoder spaces.
The aligned representation for encoder $i$ is therefore

\begin{equation}
\mathbf{Z}^{(i)}
=
\alpha_i
\widetilde{\mathbf{X}}^{(i)}
\mathbf{Q}_i
\end{equation}

By minimizing the total discrepancy between all transformed
representations and the shared consensus, GPA preserves the internal
geometry of each representation up to rotation, reflection, and global
scaling while placing all encoders into a common $d$-dimensional
coordinate system.

\paragraph{\textbf{Joint PCA.}}
We first concatenate the matched encoder representations along the
feature dimension to form a joint representation

\begin{equation}
\mathbf{X}_{\mathrm{joint}}
=
\left[
\mathbf{X}^{(1)}
\;\middle|\;
\mathbf{X}^{(2)}
\;\middle|\;
\cdots
\;\middle|\;
\mathbf{X}^{(M)}
\right]
\in
\mathbb{R}^{N\times \sum_{i=1}^{M} d_i}
\end{equation}

Principal Component Analysis (PCA) is then applied to
$\mathbf{X}_{\mathrm{joint}}$, and the leading $d$ principal components
are retained to obtain a common latent target
$\mathbf{G}\in\mathbb{R}^{N\times d}$. This joint target captures the
dominant directions of variation across all encoder representations
simultaneously.

To obtain an encoder-specific mapping into this common space, we
independently fit a ridge regression projection $\mathbf{W}_i$ for each
encoder by solving the following:

\begin{equation}
\mathbf{W}_i^{*}
=
\arg\min_{\mathbf{W}_i}
\left\|
\mathbf{X}^{(i)}\mathbf{W}_i-\mathbf{G}
\right\|_F^2
+
\lambda\|\mathbf{W}_i\|_F^2
\end{equation}

where $\lambda$ controls the strength of the ridge regularization.
The aligned representation of encoder $i$ is therefore

\begin{equation}
\mathbf{Z}^{(i)}
=
\mathbf{X}^{(i)}\mathbf{W}_i^{*}
\in
\mathbb{R}^{N\times d}
\end{equation}

Unlike GCCA or MCCA, Joint PCA does not explicitly optimize
cross-encoder correlation. Instead, it defines a shared coordinate
system from the dominant variance of the concatenated multi-encoder
representation and subsequently learns how each individual encoder
maps into that space. Because each $\mathbf{W}_i^{*}$ acts on a single
encoder representation, the learned mappings can also be applied to
unseen samples from any individual encoder without requiring the other
encoder views at inference time.

\paragraph{\textbf{Shared Autoencoder.}}
To capture nonlinear correspondences across encoder representation
spaces, we use encoder-specific neural networks
$E_i$ that map each representation
$\mathbf{X}^{(i)}$ into a common $d$-dimensional bottleneck:

\begin{equation}
\mathbf{Z}^{(i)}
=
E_i\!\left(\mathbf{X}^{(i)}\right)
\in
\mathbb{R}^{N\times d}
\end{equation}

Each encoder view is paired with a corresponding decoder
$D_i$, which reconstructs the original representation from the shared
latent code:

\begin{equation}
\widehat{\mathbf{X}}^{(i)}
=
D_i\!\left(\mathbf{Z}^{(i)}\right)
\end{equation}

For matched samples across all $M$ encoders, we define the consensus
latent representation as

\begin{equation}
\bar{\mathbf{Z}}
=
\frac{1}{M}
\sum_{i=1}^{M}
\mathbf{Z}^{(i)}
\end{equation}

Training jointly balances reconstruction fidelity and cross-encoder
alignment through the objective

\begin{equation}
\mathcal{L}
=
\mathcal{L}_{\mathrm{recon}}
+
\lambda_{\mathrm{align}}
\mathcal{L}_{\mathrm{align}}
\end{equation}

where the reconstruction term is

\begin{equation}
\mathcal{L}_{\mathrm{recon}}
=
\sum_{i=1}^{M}
\left\|
\mathbf{X}^{(i)}
-
\widehat{\mathbf{X}}^{(i)}
\right\|_F^2
\end{equation}

and the alignment term encourages encoder-specific latent
representations to remain close to the shared consensus:

\begin{equation}
\mathcal{L}_{\mathrm{align}}
=
\sum_{i=1}^{M}
\left\|
\mathbf{Z}^{(i)}
-
\bar{\mathbf{Z}}
\right\|_F^2.
\end{equation}

Here, $\lambda_{\mathrm{align}}$ controls the trade-off between
preserving information from each original encoder and enforcing
cross-encoder consistency in the shared latent space. In contrast to
the linear mappings used by GCCA, MCCA, GPA, and Joint PCA, the shared
autoencoder learns nonlinear encoder-specific transformations, allowing
more flexible alignment when corresponding structures across
representation spaces are not well described by linear mappings.

\paragraph{\textbf{Training Protocol.}}
All normalization, dimensionality reduction, and alignment parameters
are fitted using training samples only and are frozen for validation and
test data. Patch-level alignment is restricted to representations from
matched spatial coordinates, while slide-level alignment uses matched
WSIs.

\section{Additional Shared-Space Evaluation Details}\label{app:shared_space_eval}

This section provides additional evaluation details for assessing whether
the learned shared space supports cross-model correspondence,
preserves encoder-specific geometry, retains downstream utility, and
enables translation between independently learned feature spaces.

% ============================================================
\subsection{Cross-Model Agreement and Retrieval}
\label{app:agreement_retrieval}

For each alignment method, the aligner is fitted using a training subset
of row-paired patch representations and evaluated exclusively on held-out
patches. Given aligned representations
\(\mathbf{Z}^{(i)},\mathbf{Z}^{(j)}\in\mathbb{R}^{N\times d}\),
cross-model agreement is measured using the mean cosine similarity of
corresponding patches,

\begin{equation}
S_{\mathrm{align}}(i,j)
=
\frac{1}{N}
\sum_{n=1}^{N}
\cos\!\left(
\mathbf{z}^{(i)}_n,
\mathbf{z}^{(j)}_n
\right)
\end{equation}

We additionally evaluate whether this correspondence is sufficiently
strong to support cross-model retrieval. For every ordered encoder pair,
representations from encoder \(i\) are used as queries against a database
constructed from encoder \(j\). We consider two retrieval settings:
\emph{identity retrieval}, where the relevant item is the representation
of the same physical patch, and \emph{label retrieval}, where patches
sharing the same slide-level class label are considered relevant.
Performance is measured using Recall@\(K\),
\(K\in\{1,5,10\}\), mean reciprocal rank (MRR), mean average precision
(mAP), and normalized discounted cumulative gain (NDCG).

\paragraph{Recall@\(K\).}
Recall@\(K\) measures whether a relevant target is retrieved within the top-\(K\) ranked results for a given query. We report Recall@\(K\) for \(K\in\{1,5,10\}\), where larger values indicate that corresponding or label-matched samples are recovered more frequently near the top of the retrieval list. This metric is particularly useful for evaluating whether the shared representation places relevant cross-model samples within a small candidate set.

\paragraph{Mean Reciprocal Rank (MRR).}
MRR measures how highly the first relevant target is ranked for each query. If the first relevant item appears at rank \(r_q\) for query \(q\), its reciprocal rank is \(1/r_q\), and MRR averages this quantity across all queries. Higher MRR therefore indicates that the first correct cross-model match tends to appear closer to the top of the ranked list.

\paragraph{Mean Average Precision (mAP).}
mAP evaluates retrieval quality when multiple relevant samples may exist for a query, as in label-based retrieval. For each query, average precision summarizes the precision obtained at ranks where relevant samples are retrieved, and mAP averages this value across all queries. Higher mAP indicates that relevant samples are not only retrieved, but are consistently concentrated toward the top of the ranked list.

\paragraph{Normalized Discounted Cumulative Gain (NDCG).}
NDCG evaluates the overall ordering of relevant samples in the retrieved ranking while assigning greater importance to relevant items appearing at higher ranks. The discounted cumulative gain is normalized by the score of an ideal ranking, producing values in a comparable range across queries. Higher NDCG therefore indicates that the retrieved list more closely matches the ideal relevance ordering, making it useful for assessing the quality of the full cross-model ranking rather than only the first relevant match.

\paragraph{Shared-Space CKA.}
Shared-space CKA measures the global representational correspondence between two encoders after projection into the common latent space. Higher values indicate greater similarity in the geometry of the aligned representations, complementing paired cosine similarity, which measures correspondence between individually matched samples.

% ============================================================
\subsection{Geometry Preservation}
\label{app:geometry_preservation}

High cross-model agreement alone does not guarantee a useful shared
representation, since alignment could increase correspondence while
discarding important structure from the original feature spaces. We
therefore evaluate whether the geometry of each encoder is preserved
after projection.

\paragraph{Neighborhood Preservation.}
Neighborhood preservation measures the fraction of local neighbors that remain unchanged after alignment. For each sample, the overlap between its native-space and shared-space \(k\)-nearest-neighbor sets is computed and averaged across samples. A value of 1 indicates perfect preservation of local structure, whereas lower values indicate increasing distortion of the encoder's local geometry.

\paragraph{Normalized Mean Squared Error (NMSE).}
NMSE measures reconstruction error relative to the energy of the original representation, making reconstruction quality comparable across encoders with different feature scales. Lower values indicate greater fidelity, with an NMSE of 0 corresponding to exact reconstruction.

\paragraph{\(\mathbf{R^2}\).}
The coefficient of determination, \(R^2\), quantifies how much of the variance in the original representation is retained after reconstruction. Values closer to 1 indicate stronger preservation of feature variability, while values near or below 0 indicate that substantial information has been lost through the shared representation.

\paragraph{Reconstruction Cosine Similarity.}
Reconstruction cosine similarity measures directional agreement between the original and reconstructed feature vectors independently of their magnitude. Values approaching 1 indicate that the reconstruction preserves the orientation of the original representation, complementing NMSE and \(R^2\), which are sensitive to reconstruction error and explained variance.

% ============================================================
\subsection{Downstream Utility}
\label{app:downstream_utility}

We evaluate whether the aligned representation retains information
required for slide-level prediction. Under an identical patient-level MIL
evaluation protocol, we compare three representation conditions:
\emph{single}, where each encoder is evaluated independently;
\emph{concatenated}, where representations from all compatible encoders
are concatenated at the patch level; and \emph{shared}, where patches are
represented using the learned common latent space.

ABMIL and TransMIL are evaluated as the downstream MIL heads so that
conclusions are not tied to a single aggregation architecture. The same
patient-disjoint train/test partition is used across representation
conditions. Importantly, for the shared-space condition, the alignment
function is fitted exclusively using patches from training slides before
being applied to either training or test bags, preventing information
from the held-out patients from entering the alignment procedure.

Area under the ROC curve (AUC) is used as the primary evaluation metric.
Accuracy, macro-F1, and balanced accuracy are additionally retained as
secondary measures of downstream performance. This experiment tests
whether the shared representation merely aligns encoder coordinates or
also preserves task-relevant information necessary for slide-level
prediction.

% ============================================================
\subsection{Cross-Model Transfer}
\label{app:cross_model_transfer}

Finally, we evaluate whether the shared space can serve as an intermediate
representation for direct translation between encoder feature spaces.
For a representation \(\mathbf{X}^{(i)}\) produced by source encoder
\(i\), we first project it into the shared space using \(E_i\) and then
decode it using the target-specific mapping \(D_j\),

\begin{equation}
\mathbf{X}^{(i)}
\xrightarrow{E_i}
\mathbf{Z}
\xrightarrow{D_j}
\widetilde{\mathbf{X}}^{(j)},
\qquad i\neq j.
\end{equation}

Here, \(\widetilde{\mathbf{X}}^{(j)}\) is the representation of the
source patch translated into the native feature space of encoder \(j\).
Transfer is evaluated for all valid ordered source--target pairs whose
features share the same patch-coordinate grid. The translated embedding
is compared with the native target representation
\(\mathbf{X}^{(j)}\) of the same patch using cosine similarity and
feature-space reconstruction fidelity.

 We also evaluate \emph{self round-trip} reconstruction,

\begin{equation}
\mathbf{X}^{(i)}
\xrightarrow{E_i}
\mathbf{Z}
\xrightarrow{D_i}
\widetilde{\mathbf{X}}^{(i)}
\end{equation}

which isolates information lost through the shared bottleneck itself.
The gap between self round-trip and cross-model transfer therefore
quantifies the additional difficulty introduced by translating between
different encoder representation spaces.

\section{Additional Results}
The supplementary analyses provide a broader characterization of representation geometry, shared-space interoperability, and downstream utility across pathology foundation models.

\paragraph{Representation similarity.}
Section~\ref{f1} shows that slide-level encoder agreement is substantial but varies across cohorts and similarity metrics, indicating that no single encoder relationship is universally dominant. Section~\ref{f2} extends this analysis to patch encoders and shows that representational similarity is also influenced by image resolution and magnification, while Section~\ref{f3} demonstrates additional variation across cancer subcohorts. Together, these results show that encoder agreement is structured by both model characteristics and the underlying tissue distribution.

\paragraph{Shared-space interoperability.}
Section~\ref{f4} evaluates whether aligned representations support direct cross-model retrieval and shows a large improvement over unaligned controls, with GCCA providing the most consistently strong performance across configurations. Section~\ref{f5} further tests interoperability through cross-encoder representation translation, demonstrating that features from one encoder can be mapped through the shared space and reconstructed in another encoder's native feature space. 

\paragraph{Functional utility and representation depth.}
Section~\ref{f6} evaluates whether the shared representation retains information required for downstream prediction. The shared features remain competitive on several tasks and occasionally outperform single-encoder or concatenated representations, but the gains are task- and MIL-architecture-dependent, highlighting a trade-off between cross-model alignment and preservation of task-specific information. Finally, Section~\ref{f7} analyzes how similarity evolves through network depth and shows that several encoder pairs become more aligned in later layers under CLS/first-token pooling, while mean-pooled representations remain broadly similar throughout the network.

\paragraph{Hyperparameters.}
The core experimental settings are summarized in Tables~\ref{tab:full_run_configuration}--\ref{tab:alignment_parameters}. Table~\ref{tab:full_run_configuration} reports the global sampling and latent-space configuration used across MOSAIC, while Table~\ref{tab:similarity_parameters} provides the implementation details for the seven representation-similarity metrics. Table~\ref{tab:alignment_parameters} then summarizes the method-specific settings for Joint PCA, GCCA, MCCA, GPA, and the shared autoencoder used to construct the common latent space.

\clearpage
\onecolumn

\begin{table}[t]
\centering
\caption{\textbf{Core configuration used for the full MOSAIC experiments.}}
\label{tab:full_run_configuration}
\small
\setlength{\tabcolsep}{6pt}
\renewcommand{\arraystretch}{1.08}
\begin{tabular}{lr}
\toprule
\textbf{Parameter} & \textbf{Value} \\
\midrule
Random seed & 0 \\
Patches sampled per grid & 50,000 \\
Maximum slides sampled & 500 \\
Similarity-metric subsample & 8,000 \\
Shared latent dimension & 64 \\
Retrieval evaluation sample & 5,000 \\
Patches retained per MIL slide & 4,000 \\
Patches used to fit downstream aligner & 50,000 \\
\bottomrule
\end{tabular}
\end{table}

\begin{table*}[t]
\centering
\caption{\textbf{Implementation settings for representation-similarity metrics.}
All metrics are evaluated on the same 8,000-patch subsample with random seed 0.}
\label{tab:similarity_parameters}
\small
\setlength{\tabcolsep}{6pt}
\renewcommand{\arraystretch}{1.08}
\begin{tabular}{p{0.18\textwidth} p{0.75\textwidth}}
\toprule
\textbf{Metric} & \textbf{Configuration} \\
\midrule
Linear CKA &
Centered representations; biased estimator. \\

Kernel CKA &
RBF bandwidth set to the median pairwise distance; centered; biased estimator. \\

SVCCA &
99\% variance retained; rank tolerance $10^{-10}$. \\

PWCCA &
Rank tolerance $10^{-10}$; symmetric two-direction average. \\

Procrustes similarity &
Centered; scale invariant; dimensions zero-padded when required. \\

Cosine RSA &
Cosine similarity matrices compared using Spearman rank correlation. \\

Distance correlation &
Euclidean distances with unbiased U-centering. \\
\bottomrule
\end{tabular}
\end{table*}

\begin{table*}[t]
\centering
\caption{\textbf{Configuration of the shared-space alignment methods.}
Unless otherwise specified, the shared latent dimensionality is 64.}
\label{tab:alignment_parameters}
\small
\setlength{\tabcolsep}{6pt}
\renewcommand{\arraystretch}{1.08}
\begin{tabular}{p{0.22\textwidth} p{0.71\textwidth}}
\toprule
\textbf{Method} & \textbf{Configuration} \\
\midrule
Joint PCA &
Ridge coefficient $10^{-4}$; no pre-PCA or whitening. \\

GCCA &
Ridge coefficient $10^{-4}$; rank tolerance $10^{-8}$; no pre-PCA. \\

MCCA &
Ridge coefficient $10^{-3}$; per-view PCA dimension 128. \\

GPA &
Per-view PCA to 64 dimensions; maximum 300 iterations;
tolerance $10^{-7}$; isotropic scaling enabled. \\

Shared autoencoder &
Hidden dimensions $(512,256)$; GELU activation; batch size 256;
learning rate $10^{-3}$; weight decay $10^{-5}$;
alignment weight 1.0; latent LayerNorm. \\
\bottomrule
\end{tabular}
\end{table*}

\begin{table*}[t]
\centering
\caption{\textbf{Configuration of cross-model retrieval and
cross-encoder representation translation.}}
\label{tab:retrieval_translation_settings}
\small
\setlength{\tabcolsep}{6pt}
\renewcommand{\arraystretch}{1.08}
\begin{tabular}{p{0.25\textwidth} p{0.32\textwidth} p{0.32\textwidth}}
\toprule
\textbf{Setting} &
\textbf{Cross-model retrieval} &
\textbf{Representation translation} \\
\midrule
Loaded patches & 50,000 & 50,000 \\
Train/test split & 70\% / 30\% & 70\% / 30\% \\
Evaluation sample & 5,000 & 5,000 \\
Latent dimension & 64 & 64 \\
Seed & 0 & 0 \\
\bottomrule
\end{tabular}
\end{table*}

\clearpage

\subsection{Similarity Metrics for Slide Encoders}\label{f1}

\begin{table*}[!h]
\centering
\begin{threeparttable}

\caption{\textbf{Pairwise representational similarity among slide-level encoders.}
Similarity is computed between slide-level representations produced by
each encoder using its native preprocessing and patch grid.}
\label{tab:slide_encoder_similarity}

\scriptsize
\setlength{\tabcolsep}{6.0pt}
\renewcommand{\arraystretch}{1.08}

\sisetup{
    table-format=1.3,
    detect-weight=true,
    detect-inline-weight=math
}

\begin{tabular}{
@{}
ll
*{7}{S}
@{}
}
\toprule

& &
\multicolumn{2}{c}{\textbf{CKA}} &
\multicolumn{2}{c}{\textbf{CCA}} &
\multicolumn{3}{c}{\textbf{Relational / Shape}} \\
\cmidrule(lr){3-4}
\cmidrule(lr){5-6}
\cmidrule(l){7-9}

\textbf{Encoder}
& \textbf{Compared with}
& {\textbf{Linear}}
& {\textbf{Kernel}}
& {\textbf{SVCCA}}
& {\textbf{PWCCA}}
& {\textbf{Proc.}}
& {\textbf{Cos.\ RSA}}
& {\textbf{dCor}} \\

\midrule

% =========================================================
\multicolumn{9}{@{}l}{%
    \textbf{(a) CPTAC}
    \hfill
    \textit{$n=2{,}162$ paired slides}
} \\
\addlinespace[2pt]
\midrule

\multirow{5}{*}{\textbf{CHIEF}}
& Feather
& 0.548 & 0.603 & 0.507 & 0.792 & 0.652 & 0.551 & 0.784 \\

& GigaPath
& \bfseries 0.810
& \bfseries 0.829
& \bfseries 0.787
& 0.813
& \bfseries 0.799
& \bfseries 0.803
& \bfseries 0.910 \\

& Madeleine
& 0.736 & 0.765 & 0.607 & 0.849 & 0.769 & 0.729 & 0.876 \\

& PRISM
& 0.629 & 0.665 & 0.592 & 0.893 & 0.662 & 0.637 & 0.816 \\

& TITAN
& 0.685 & 0.713 & 0.526 & 0.762 & 0.689 & 0.680 & 0.846 \\

\addlinespace[3pt]

\multirow{4}{*}{\textbf{Feather}}
& GigaPath
& 0.548 & 0.593 & 0.669 & 0.780 & 0.630 & 0.571 & 0.778 \\

& Madeleine
& 0.647 & 0.700 & 0.575 & 0.710 & 0.729 & 0.658 & 0.847 \\

& PRISM
& 0.549 & 0.645 & 0.601
& \bfseries 0.905
& 0.660 & 0.597 & 0.814 \\

& TITAN
& 0.658 & 0.704 & 0.544 & 0.800 & 0.729 & 0.667 & 0.852 \\

\addlinespace[3pt]

\multirow{3}{*}{\textbf{GigaPath}}
& Madeleine
& 0.795 & 0.805 & 0.742 & 0.849 & 0.770 & 0.786 & 0.897 \\

& PRISM
& 0.688 & 0.705 & 0.537 & 0.885 & 0.662 & 0.710 & 0.838 \\

& TITAN
& 0.762 & 0.777 & 0.675 & 0.766 & 0.709 & 0.772 & 0.883 \\

\addlinespace[3pt]

\multirow{2}{*}{\textbf{Madeleine}}
& PRISM
& 0.741 & 0.776 & 0.581 & 0.903 & 0.718 & 0.746 & 0.883 \\

& TITAN
& 0.773 & 0.795 & 0.534 & 0.823 & 0.766 & 0.767 & 0.894 \\

\addlinespace[3pt]

\textbf{PRISM}
& TITAN
& 0.762 & 0.781 & 0.569 & 0.892 & 0.723 & 0.744 & 0.884 \\

% =========================================================
\midrule
\addlinespace[2pt]

\multicolumn{9}{@{}l}{%
    \textbf{(b) TCGA}
    \hfill
    \textit{$n=2{,}169$ paired slides}
} \\
\addlinespace[2pt]
\midrule

\multirow{5}{*}{\textbf{CHIEF}}
& Feather
& 0.450 & 0.494 & 0.501 & 0.778 & 0.581 & 0.513 & 0.701 \\

& GigaPath
& 0.702 & 0.708
& \bfseries 0.746
& 0.768 & 0.731 & 0.676 & 0.839 \\

& Madeleine
& 0.522 & 0.556 & 0.597 & 0.809 & 0.683 & 0.544 & 0.744 \\

& PRISM
& 0.440 & 0.469 & 0.596 & 0.880 & 0.581 & 0.476 & 0.681 \\

& TITAN
& 0.489 & 0.518 & 0.491 & 0.740 & 0.629 & 0.518 & 0.720 \\

\addlinespace[3pt]

\multirow{4}{*}{\textbf{Feather}}
& GigaPath
& 0.522 & 0.546 & 0.655 & 0.751 & 0.540 & 0.592 & 0.738 \\

& Madeleine
& 0.751 & 0.772 & 0.597 & 0.691 & 0.714 & 0.792 & 0.879 \\

& PRISM
& 0.661 & 0.697 & 0.618
& \bfseries 0.893
& 0.642 & 0.744 & 0.835 \\

& TITAN
& \bfseries 0.816
& \bfseries 0.846
& 0.543
& 0.801
& \bfseries 0.757
& \bfseries 0.858
& \bfseries 0.922 \\

\addlinespace[3pt]

\multirow{3}{*}{\textbf{GigaPath}}
& Madeleine
& 0.688 & 0.709 & 0.671 & 0.807 & 0.697 & 0.708 & 0.841 \\

& PRISM
& 0.506 & 0.518 & 0.516 & 0.870 & 0.564 & 0.563 & 0.714 \\

& TITAN
& 0.583 & 0.593 & 0.667 & 0.736 & 0.609 & 0.623 & 0.769 \\

\addlinespace[3pt]

\multirow{2}{*}{\textbf{Madeleine}}
& PRISM
& 0.640 & 0.661 & 0.557 & 0.886 & 0.636 & 0.706 & 0.810 \\

& TITAN
& 0.770 & 0.786 & 0.553 & 0.800 & 0.728 & 0.807 & 0.886 \\

\addlinespace[3pt]

\textbf{PRISM}
& TITAN
& 0.773 & 0.784 & 0.593 & 0.885 & 0.701 & 0.805 & 0.886 \\

\bottomrule
\end{tabular}

\end{threeparttable}
\end{table*}

Table~\ref{tab:slide_encoder_similarity} compares pairwise slide-level representational similarity across six encoders using seven complementary similarity metrics on CPTAC and TCGA.

\paragraph{CPTAC.}
On CPTAC, \textbf{CHIEF--GigaPath} exhibits the strongest overall representational agreement, achieving the highest Linear CKA (0.810), Kernel CKA (0.829), SVCCA (0.787), Procrustes similarity (0.799), Cosine RSA (0.803), and dCor (0.910). The only exception is PWCCA, where \textbf{Feather--PRISM} achieves the highest similarity (0.905). Other pairs such as GigaPath--Madeleine and Madeleine--TITAN also show consistently high agreement across several metrics, indicating a broader cluster of closely aligned representations.

\paragraph{TCGA.}
On TCGA, the strongest alignment shifts to \textbf{Feather--TITAN}, which achieves the highest Linear CKA (0.816), Kernel CKA (0.846), Procrustes similarity (0.757), Cosine RSA (0.858), and dCor (0.922). In contrast, \textbf{CHIEF--GigaPath} remains strongest under SVCCA (0.746), while \textbf{Feather--PRISM} again achieves the highest PWCCA score (0.893). The change in dominant encoder pairs between CPTAC and TCGA highlights the cohort-dependent nature of slide-level representational similarity.

\clearpage

\begin{table*}[t]
\centering

\caption{\textbf{Subcohort-level representational similarity among slide-level encoders on CPTAC.}
Pairwise similarity is reported for CPTAC-BRCA, CPTAC-COAD, and CPTAC-LUAD using seven complementary metrics.
Only slides shared across all six encoders within each subcohort are considered.
Bold indicates the highest informative similarity within each subcohort and metric.}
\label{tab:cptac_slide_encoder_subcohort_similarity}

\scriptsize
\setlength{\tabcolsep}{5.5pt}
\renewcommand{\arraystretch}{1.05}

\sisetup{
    table-format=1.3,
    detect-weight=true,
    detect-inline-weight=math
}

\begin{tabular}{
@{}
ll
*{7}{S}
@{}
}
\toprule

& &
\multicolumn{2}{c}{\textbf{CKA}} &
\multicolumn{2}{c}{\textbf{CCA}} &
\multicolumn{3}{c}{\textbf{Relational / Shape}} \\
\cmidrule(lr){3-4}
\cmidrule(lr){5-6}
\cmidrule(l){7-9}

\textbf{Encoder}
& \textbf{Compared with}
& {\textbf{Linear}}
& {\textbf{Kernel}}
& {\textbf{SVCCA}}
& {\textbf{PWCCA}}
& {\textbf{Proc.}}
& {\textbf{Cos.\ RSA}}
& {\textbf{dCor}} \\

\midrule

% =========================================================
% CPTAC-BRCA
% =========================================================
\multicolumn{9}{@{}l}{%
    \textbf{(a) CPTAC-BRCA}
    \hfill
    \textit{$n=654$ slides}
} \\
\addlinespace[2pt]
\midrule

\multirow{5}{*}{\textbf{CHIEF}}
& Feather
& 0.418 & 0.495 & 0.590 & 1.000 & 0.616 & 0.417 & 0.691 \\

& GigaPath
& 0.727 & 0.734
& \bfseries 0.833
& 1.000
& 0.776 & 0.684 & 0.854 \\

& Madeleine
& 0.669 & 0.677 & 0.661 & 1.000 & 0.763 & 0.603 & 0.822 \\

& PRISM
& 0.534 & 0.557 & 0.739 & 1.000 & 0.655 & 0.502 & 0.739 \\

& TITAN
& 0.530 & 0.559 & 0.651 & 1.000 & 0.661 & 0.492 & 0.743 \\

\addlinespace[3pt]

\multirow{4}{*}{\textbf{Feather}}
& GigaPath
& 0.408 & 0.475 & 0.749 & 1.000 & 0.585 & 0.410 & 0.675 \\

& Madeleine
& 0.413 & 0.487 & 0.631 & 0.917 & 0.641 & 0.400 & 0.687 \\

& PRISM
& 0.406 & 0.461 & 0.744 & 1.000 & 0.587 & 0.377 & 0.668 \\

& TITAN
& 0.516 & 0.581 & 0.676 & 1.000 & 0.669 & 0.528 & 0.759 \\

\addlinespace[3pt]

\multirow{3}{*}{\textbf{GigaPath}}
& Madeleine
& \bfseries 0.802
& \bfseries 0.802
& 0.820
& 1.000
& \bfseries 0.790
& \bfseries 0.773
& \bfseries 0.895 \\

& PRISM
& 0.531 & 0.558 & 0.643 & 1.000 & 0.655 & 0.513 & 0.736 \\

& TITAN
& 0.580 & 0.606 & 0.723 & 1.000 & 0.682 & 0.545 & 0.771 \\

\addlinespace[3pt]

\multirow{2}{*}{\textbf{Madeleine}}
& PRISM
& 0.571 & 0.597 & 0.728 & 1.000 & 0.663 & 0.559 & 0.768 \\

& TITAN
& 0.590 & 0.618 & 0.662 & 1.000 & 0.703 & 0.560 & 0.784 \\

\addlinespace[3pt]

\textbf{PRISM}
& TITAN
& 0.555 & 0.581 & 0.686 & 1.000 & 0.665 & 0.515 & 0.756 \\

% =========================================================
% CPTAC-COAD
% =========================================================
\midrule
\addlinespace[2pt]

\multicolumn{9}{@{}l}{%
    \textbf{(b) CPTAC-COAD}
    \hfill
    \textit{$n=369$ slides}
} \\
\addlinespace[2pt]
\midrule

\multirow{5}{*}{\textbf{CHIEF}}
& Feather
& 0.408 & 0.473 & 0.661 & 1.000 & 0.657 & 0.422 & 0.673 \\

& GigaPath
& 0.693 & 0.702
& \bfseries 0.889
& 1.000
& 0.796 & 0.656 & 0.830 \\

& Madeleine
& 0.638 & 0.652 & 0.746 & 1.000 & 0.772 & 0.615 & 0.800 \\

& PRISM
& 0.449 & 0.452 & 0.715 & 1.000 & 0.646 & 0.370 & 0.658 \\

& TITAN
& 0.468 & 0.495 & 0.716 & 1.000 & 0.655 & 0.424 & 0.688 \\

\addlinespace[3pt]

\multirow{4}{*}{\textbf{Feather}}
& GigaPath
& 0.429 & 0.493 & 0.847 & 1.000 & 0.665 & 0.426 & 0.684 \\

& Madeleine
& 0.438 & 0.515 & 0.739 & 1.000 & 0.681 & 0.445 & 0.703 \\

& PRISM
& 0.356 & 0.390 & 0.722 & 1.000 & 0.607 & 0.307 & 0.604 \\

& TITAN
& 0.445 & 0.506 & 0.752 & 1.000 & 0.679 & 0.415 & 0.692 \\

\addlinespace[3pt]

\multirow{3}{*}{\textbf{GigaPath}}
& Madeleine
& \bfseries 0.742
& \bfseries 0.756
& 0.862
& 1.000
& \bfseries 0.811
& \bfseries 0.711
& \bfseries 0.864 \\

& PRISM
& 0.462 & 0.473 & 0.767 & 1.000 & 0.681 & 0.390 & 0.669 \\

& TITAN
& 0.518 & 0.543 & 0.789 & 1.000 & 0.704 & 0.469 & 0.717 \\

\addlinespace[3pt]

\multirow{2}{*}{\textbf{Madeleine}}
& PRISM
& 0.490 & 0.501 & 0.687 & 1.000 & 0.679 & 0.405 & 0.692 \\

& TITAN
& 0.590 & 0.602 & 0.703 & 1.000 & 0.717 & 0.529 & 0.764 \\

\addlinespace[3pt]

\textbf{PRISM}
& TITAN
& 0.548 & 0.548 & 0.647 & 1.000 & 0.688 & 0.470 & 0.728 \\

% =========================================================
% CPTAC-LUAD
% =========================================================
\midrule
\addlinespace[2pt]

\multicolumn{9}{@{}l}{%
    \textbf{(c) CPTAC-LUAD}
    \hfill
    \textit{$n=1{,}139$ slides}
} \\
\addlinespace[2pt]
\midrule

\multirow{5}{*}{\textbf{CHIEF}}
& Feather
& 0.491 & 0.564 & 0.597 & 0.918 & 0.644 & 0.562 & 0.752 \\

& GigaPath
& \bfseries 0.787
& \bfseries 0.816
& \bfseries 0.840
& 0.925
& \bfseries 0.809
& \bfseries 0.789
& \bfseries 0.903 \\

& Madeleine
& 0.656 & 0.705 & 0.669 & 0.943 & 0.757 & 0.695 & 0.841 \\

& PRISM
& 0.542 & 0.583 & 0.597 & 1.000 & 0.632 & 0.578 & 0.762 \\

& TITAN
& 0.606 & 0.658 & 0.618 & 0.906 & 0.683 & 0.636 & 0.812 \\

\addlinespace[3pt]

\multirow{4}{*}{\textbf{Feather}}
& GigaPath
& 0.462 & 0.515 & 0.778 & 0.913 & 0.611 & 0.492 & 0.718 \\

& Madeleine
& 0.693 & 0.761 & 0.683 & 0.826 & 0.758 & 0.765 & 0.875 \\

& PRISM
& 0.681 & 0.752 & 0.632 & 1.000 & 0.685 & 0.774 & 0.869 \\

& TITAN
& 0.616 & 0.691 & 0.666 & 0.923 & 0.707 & 0.690 & 0.833 \\

\addlinespace[3pt]

\multirow{3}{*}{\textbf{GigaPath}}
& Madeleine
& 0.674 & 0.686 & 0.805 & 0.944 & 0.747 & 0.674 & 0.830 \\

& PRISM
& 0.514 & 0.528 & 0.668 & 1.000 & 0.618 & 0.505 & 0.724 \\

& TITAN
& 0.674 & 0.697 & 0.738 & 0.908 & 0.712 & 0.666 & 0.835 \\

\addlinespace[3pt]

\multirow{2}{*}{\textbf{Madeleine}}
& PRISM
& 0.757 & 0.780 & 0.584 & 1.000 & 0.706 & 0.787 & 0.881 \\

& TITAN
& 0.723 & 0.759 & 0.621 & 0.933 & 0.755 & 0.752 & 0.871 \\

\addlinespace[3pt]

\textbf{PRISM}
& TITAN
& 0.713 & 0.736 & 0.554 & 1.000 & 0.701 & 0.738 & 0.856 \\

\bottomrule
\end{tabular}

\end{table*}

Table~\ref{tab:cptac_slide_encoder_subcohort_similarity} reports pairwise slide-level representational similarity across the \textbf{CPTAC-BRCA}, \textbf{CPTAC-COAD} and \textbf{CPTAC-LUAD} subcohorts using seven complementary metrics. Across the three CPTAC subcohorts, the strongest representational alignment consistently involves GigaPath, although the most similar pairing changes by cancer type. \textbf{GigaPath--Madeleine} dominates BRCA and COAD across most metrics. PWCCA is reported for completeness but interpreted cautiously, as the smaller subcohort sample sizes relative to representation dimensionality cause substantial numerical saturation at 1.000.

\clearpage
\begin{table*}[t]
\centering

\caption{\textbf{Subcohort-level representational similarity among slide-level encoders on TCGA.}
Pairwise similarity is reported for TCGA-BRCA, TCGA-LUAD, and TCGA-LUSC using seven complementary metrics.
Only slides shared across all six encoders within each subcohort are considered.
Bold indicates the highest informative similarity within each subcohort and metric.
PWCCA is reported for completeness but excluded from ranking because several values saturate at 1.000 at the subcohort level.}
\label{tab:tcga_slide_encoder_subcohort_similarity}

\scriptsize
\setlength{\tabcolsep}{5.5pt}
\renewcommand{\arraystretch}{1.05}

\sisetup{
    table-format=1.3,
    detect-weight=true,
    detect-inline-weight=math
}

\begin{tabular}{
@{}
ll
*{7}{S}
@{}
}
\toprule

& &
\multicolumn{2}{c}{\textbf{CKA}} &
\multicolumn{2}{c}{\textbf{CCA}} &
\multicolumn{3}{c}{\textbf{Relational / Shape}} \\
\cmidrule(lr){3-4}
\cmidrule(lr){5-6}
\cmidrule(l){7-9}

\textbf{Encoder}
& \textbf{Compared with}
& {\textbf{Linear}}
& {\textbf{Kernel}}
& {\textbf{SVCCA}}
& {\textbf{PWCCA}}
& {\textbf{Proc.}}
& {\textbf{Cos.\ RSA}}
& {\textbf{dCor}} \\

\midrule

% =========================================================
% TCGA-BRCA
% =========================================================

\multicolumn{9}{@{}l}{%
    \textbf{(a) TCGA-BRCA}
    \hfill
    \textit{$n=1{,}126$ slides}
} \\
\addlinespace[2pt]
\midrule

\multirow{5}{*}{\textbf{CHIEF}}
& Feather
& 0.281 & 0.329 & 0.551 & 0.909 & 0.561 & 0.292 & 0.564 \\

& GigaPath
& \bfseries 0.632
& 0.630
& \bfseries 0.787
& 0.892
& \bfseries 0.716
& 0.539
& 0.789 \\

& Madeleine
& 0.479 & 0.512 & 0.650 & 0.925 & 0.696 & 0.415 & 0.708 \\

& PRISM
& 0.337 & 0.355 & 0.663 & 1.000 & 0.574 & 0.300 & 0.585 \\

& TITAN
& 0.473 & 0.500 & 0.594 & 0.889 & 0.646 & 0.407 & 0.699 \\

\addlinespace[3pt]

\multirow{4}{*}{\textbf{Feather}}
& GigaPath
& 0.248 & 0.278 & 0.726 & 0.895 & 0.504 & 0.231 & 0.513 \\

& Madeleine
& 0.450 & 0.485 & 0.653 & 0.803 & 0.662 & 0.454 & 0.691 \\

& PRISM
& 0.347 & 0.389 & 0.681 & 1.000 & 0.558 & 0.343 & 0.614 \\

& TITAN
& 0.520 & 0.562 & 0.654 & 0.921 & 0.672 & 0.516 & 0.746 \\

\addlinespace[3pt]

\multirow{3}{*}{\textbf{GigaPath}}
& Madeleine
& 0.630
& \bfseries 0.651
& 0.738
& 0.923
& 0.709
& \bfseries 0.547
& \bfseries 0.799 \\

& PRISM
& 0.289 & 0.299 & 0.592 & 1.000 & 0.552 & 0.255 & 0.530 \\

& TITAN
& 0.468 & 0.472 & 0.698 & 0.883 & 0.630 & 0.375 & 0.675 \\

\addlinespace[3pt]

\multirow{2}{*}{\textbf{Madeleine}}
& PRISM
& 0.358 & 0.386 & 0.628 & 1.000 & 0.585 & 0.341 & 0.610 \\

& TITAN
& 0.554 & 0.578 & 0.563 & 0.920 & 0.697 & 0.496 & 0.753 \\

\addlinespace[3pt]

\textbf{PRISM}
& TITAN
& 0.529 & 0.562 & 0.627 & 1.000 & 0.648 & 0.497 & 0.742 \\

% =========================================================
% TCGA-LUAD
% =========================================================

\midrule
\addlinespace[2pt]

\multicolumn{9}{@{}l}{%
    \textbf{(b) TCGA-LUAD}
    \hfill
    \textit{$n=531$ slides}
} \\
\addlinespace[2pt]
\midrule

\multirow{5}{*}{\textbf{CHIEF}}
& Feather
& 0.273 & 0.331 & 0.657 & 1.000 & 0.585 & 0.306 & 0.555 \\

& GigaPath
& \bfseries 0.725
& \bfseries 0.741
& \bfseries 0.860
& 1.000
& \bfseries 0.780
& \bfseries 0.674
& \bfseries 0.854 \\

& Madeleine
& 0.454 & 0.486 & 0.711 & 1.000 & 0.698 & 0.429 & 0.683 \\

& PRISM
& 0.356 & 0.387 & 0.680 & 1.000 & 0.614 & 0.346 & 0.602 \\

& TITAN
& 0.391 & 0.429 & 0.628 & 1.000 & 0.640 & 0.381 & 0.639 \\

\addlinespace[3pt]

\multirow{4}{*}{\textbf{Feather}}
& GigaPath
& 0.262 & 0.306 & 0.838 & 1.000 & 0.550 & 0.273 & 0.526 \\

& Madeleine
& 0.459 & 0.494 & 0.751 & 0.986 & 0.671 & 0.453 & 0.692 \\

& PRISM
& 0.496 & 0.558 & 0.733 & 1.000 & 0.648 & 0.515 & 0.742 \\

& TITAN
& 0.560 & 0.617 & 0.705 & 1.000 & 0.707 & 0.582 & 0.778 \\

\addlinespace[3pt]

\multirow{3}{*}{\textbf{GigaPath}}
& Madeleine
& 0.569 & 0.593 & 0.825 & 1.000 & 0.724 & 0.537 & 0.756 \\

& PRISM
& 0.346 & 0.364 & 0.769 & 1.000 & 0.607 & 0.303 & 0.572 \\

& TITAN
& 0.377 & 0.402 & 0.817 & 1.000 & 0.637 & 0.336 & 0.609 \\

\addlinespace[3pt]

\multirow{2}{*}{\textbf{Madeleine}}
& PRISM
& 0.467 & 0.489 & 0.648 & 1.000 & 0.653 & 0.434 & 0.683 \\

& TITAN
& 0.524 & 0.546 & 0.674 & 1.000 & 0.706 & 0.485 & 0.725 \\

\addlinespace[3pt]

\textbf{PRISM}
& TITAN
& 0.606 & 0.629 & 0.671 & 1.000 & 0.696 & 0.597 & 0.784 \\

% =========================================================
% TCGA-LUSC
% =========================================================

\midrule
\addlinespace[2pt]

\multicolumn{9}{@{}l}{%
    \textbf{(c) TCGA-LUSC}
    \hfill
    \textit{$n=512$ slides}
} \\
\addlinespace[2pt]
\midrule

\multirow{5}{*}{\textbf{CHIEF}}
& Feather
& 0.203 & 0.263 & 0.591 & 1.000 & 0.537 & 0.217 & 0.481 \\

& GigaPath
& \bfseries 0.696
& \bfseries 0.713
& \bfseries 0.864
& 1.000
& \bfseries 0.763
& \bfseries 0.635
& \bfseries 0.836 \\

& Madeleine
& 0.353 & 0.406 & 0.716 & 1.000 & 0.662 & 0.347 & 0.616 \\

& PRISM
& 0.245 & 0.282 & 0.699 & 1.000 & 0.559 & 0.235 & 0.501 \\

& TITAN
& 0.287 & 0.337 & 0.656 & 1.000 & 0.598 & 0.287 & 0.554 \\

\addlinespace[3pt]

\multirow{4}{*}{\textbf{Feather}}
& GigaPath
& 0.270 & 0.319 & 0.811 & 1.000 & 0.554 & 0.257 & 0.531 \\

& Madeleine
& 0.405 & 0.444 & 0.708 & 1.000 & 0.654 & 0.384 & 0.650 \\

& PRISM
& 0.298 & 0.342 & 0.712 & 1.000 & 0.570 & 0.277 & 0.557 \\

& TITAN
& 0.406 & 0.455 & 0.687 & 1.000 & 0.662 & 0.406 & 0.657 \\

\addlinespace[3pt]

\multirow{3}{*}{\textbf{GigaPath}}
& Madeleine
& 0.558 & 0.588 & 0.820 & 1.000 & 0.722 & 0.528 & 0.750 \\

& PRISM
& 0.328 & 0.355 & 0.743 & 1.000 & 0.600 & 0.290 & 0.562 \\

& TITAN
& 0.352 & 0.394 & 0.806 & 1.000 & 0.639 & 0.331 & 0.598 \\

\addlinespace[3pt]

\multirow{2}{*}{\textbf{Madeleine}}
& PRISM
& 0.401 & 0.421 & 0.660 & 1.000 & 0.627 & 0.362 & 0.627 \\

& TITAN
& 0.424 & 0.457 & 0.649 & 1.000 & 0.681 & 0.405 & 0.657 \\

\addlinespace[3pt]

\textbf{PRISM}
& TITAN
& 0.444 & 0.467 & 0.678 & 1.000 & 0.644 & 0.404 & 0.664 \\

\bottomrule
\end{tabular}

\end{table*}

Table~\ref{tab:tcga_slide_encoder_subcohort_similarity} summarizes pairwise slide-level representational similarity across the TCGA-BRCA, TCGA-LUAD, and TCGA-LUSC subcohorts. Across TCGA subcohorts, GigaPath consistently appears in the strongest representational pairings, with \textbf{CHIEF--GigaPath} dominating both LUAD and LUSC across all informative metrics, while BRCA shows a split between CHIEF--GigaPath and GigaPath--Madeleine. PWCCA is reported for completeness but interpreted cautiously because the relatively small subcohort sample sizes lead to substantial saturation at 1.000.

\subsection{Patch-Level Similarity Metrics for Patch Encoders}\label{f2}

\begin{table*}[!h]
\centering
\caption{\textbf{Patch-level representational similarity at 224 px on CPTAC.}
Pairwise similarity across encoders at 5$\times$, 10$\times$, and 20$\times$ magnification.
Bold denotes the highest value for each metric within a magnification.}
\label{tab:patch_similarity_224}

\scriptsize
\setlength{\tabcolsep}{5pt}
\renewcommand{\arraystretch}{1.08}
\sisetup{table-format=1.3, detect-weight=true, detect-inline-weight=math}

\begin{tabular}{@{}cll*{7}{S}@{}}
\toprule
&&&
\multicolumn{2}{c}{\textbf{CKA}} &
\multicolumn{2}{c}{\textbf{CCA}} &
\multicolumn{3}{c}{\textbf{Relational / Shape}} \\
\cmidrule(lr){4-5}
\cmidrule(lr){6-7}
\cmidrule(l){8-10}

\textbf{Mag.} &
\textbf{Encoder A} &
\textbf{Encoder B} &
{\textbf{Linear}} &
{\textbf{Kernel}} &
{\textbf{SVCCA}} &
{\textbf{PWCCA}} &
{\textbf{Proc.}} &
{\textbf{Cos.\ RSA}} &
{\textbf{dCor}} \\
\midrule

\multirow{3}{*}{5$\times$}
& GPFM & H-optimus-0
& \bfseries 0.781 & \bfseries 0.783 & 0.583 & 0.744
& \bfseries 0.756 & \bfseries 0.734 & \bfseries 0.886 \\
& GPFM & Virchow2
& 0.479 & 0.542 & \bfseries 0.605 & \bfseries 0.825
& 0.700 & 0.465 & 0.736 \\
& H-optimus-0 & Virchow2
& 0.406 & 0.456 & 0.477 & 0.748 & 0.644 & 0.392 & 0.674 \\

\midrule

\multirow{3}{*}{10$\times$}
& GPFM & H-optimus-0
& \bfseries 0.702 & \bfseries 0.702 & 0.606 & 0.752
& \bfseries 0.745 & \bfseries 0.640 & \bfseries 0.835 \\
& GPFM & Virchow2
& 0.463 & 0.517 & \bfseries 0.625 & \bfseries 0.839
& 0.703 & 0.431 & 0.719 \\
& H-optimus-0 & Virchow2
& 0.454 & 0.494 & 0.506 & 0.766 & 0.684 & 0.448 & 0.699 \\

\midrule

\multirow{6}{*}{20$\times$}
& GPFM & H-optimus-0
& \bfseries 0.674 & \bfseries 0.666 & 0.593 & 0.737
& \bfseries 0.720 & 0.597 & \bfseries 0.809 \\
& GPFM & Virchow
& 0.610 & 0.633 & 0.605 & 0.829 & 0.689 & \bfseries 0.611 & 0.801 \\
& GPFM & Virchow2
& 0.432 & 0.481 & \bfseries 0.617 & \bfseries 0.837
& 0.687 & 0.391 & 0.694 \\
& H-optimus-0 & Virchow
& 0.498 & 0.516 & 0.501 & 0.768 & 0.666 & 0.540 & 0.717 \\
& H-optimus-0 & Virchow2
& 0.454 & 0.489 & 0.531 & 0.772 & 0.676 & 0.439 & 0.697 \\
& Virchow & Virchow2
& 0.318 & 0.369 & 0.552 & 0.733 & 0.666 & 0.331 & 0.614 \\

\bottomrule
\end{tabular}
\end{table*}

\begin{table*}[!h]
\centering
\caption{\textbf{Patch-level representational similarity at 256 px and 5$\times$ on CPTAC.}
Bold denotes the highest value for each metric.}
\label{tab:patch_similarity_256_5x}

\scriptsize
\setlength{\tabcolsep}{5.5pt}
\renewcommand{\arraystretch}{1.08}
\sisetup{table-format=1.3, detect-weight=true, detect-inline-weight=math}

\begin{tabular}{@{}ll*{7}{S}@{}}
\toprule
&&
\multicolumn{2}{c}{\textbf{CKA}} &
\multicolumn{2}{c}{\textbf{CCA}} &
\multicolumn{3}{c}{\textbf{Relational / Shape}} \\
\cmidrule(lr){3-4}
\cmidrule(lr){5-6}
\cmidrule(l){7-9}

\textbf{Encoder} &
\textbf{Compared with} &
{\textbf{Linear}} &
{\textbf{Kernel}} &
{\textbf{SVCCA}} &
{\textbf{PWCCA}} &
{\textbf{Proc.}} &
{\textbf{Cos.\ RSA}} &
{\textbf{dCor}} \\
\midrule

\multirow{4}{*}{\textbf{CTransPath}}
& Prov-GigaPath & 0.660 & 0.705 & 0.655 & \bfseries 0.721 & 0.695 & 0.663 & 0.844 \\
& KEEP & 0.608 & 0.627 & 0.451 & 0.566 & 0.643 & 0.556 & 0.792 \\
& ResNet50 & 0.650 & 0.713 & 0.486 & 0.615 & \bfseries 0.710 & 0.705 & 0.848 \\
&UNI2-H & 0.686 & 0.709 & \bfseries 0.664 & 0.706 & 0.656 & 0.636 & 0.843 \\

\addlinespace[3pt]

\multirow{3}{*}{\textbf{Prov-GigaPath}}
& KEEP & 0.548 & 0.574 & 0.570 & 0.666 & 0.604 & 0.515 & 0.760 \\
& ResNet50  & 0.623 & 0.625 & 0.484 & 0.602 & 0.603 & 0.561 & 0.790 \\
&UNI2-H & 0.689 & 0.714 & 0.500 & 0.607 & 0.692 & 0.666 & 0.847 \\

\addlinespace[3pt]

\multirow{2}{*}{\textbf{KEEP}}
& ResNet50  & 0.401 & 0.412 & 0.373 & 0.522 & 0.524 & 0.381 & 0.639 \\
&UNI2-H
& \bfseries 0.783 & \bfseries 0.802 & 0.619 & 0.686
& 0.646 & \bfseries 0.744 & \bfseries 0.898 \\

\addlinespace[3pt]

\textbf{ResNet50}
&UNI2-H & 0.445 & 0.466 & 0.494 & 0.581 & 0.527 & 0.410 & 0.682 \\

\bottomrule
\end{tabular}
\end{table*}

\begin{table*}[!h]
\centering
\caption{\textbf{Patch-level representational similarity at 256 px and 10$\times$ on CPTAC.}
Bold denotes the highest value for each metric.}
\label{tab:patch_similarity_256_10x}

\scriptsize
\setlength{\tabcolsep}{5.5pt}
\renewcommand{\arraystretch}{1.08}
\sisetup{table-format=1.3, detect-weight=true, detect-inline-weight=math}

\begin{tabular}{@{}ll*{7}{S}@{}}
\toprule
&&
\multicolumn{2}{c}{\textbf{CKA}} &
\multicolumn{2}{c}{\textbf{CCA}} &
\multicolumn{3}{c}{\textbf{Relational / Shape}} \\
\cmidrule(lr){3-4}
\cmidrule(lr){5-6}
\cmidrule(l){7-9}

\textbf{Encoder} &
\textbf{Compared with} &
{\textbf{Linear}} &
{\textbf{Kernel}} &
{\textbf{SVCCA}} &
{\textbf{PWCCA}} &
{\textbf{Proc.}} &
{\textbf{Cos.\ RSA}} &
{\textbf{dCor}} \\
\midrule

\multirow{5}{*}{\textbf{CONCH}}
& CTransPath & 0.671 & 0.703 & 0.616 & 0.646 & 0.660 & 0.655 & 0.841 \\
& Prov-GigaPath & 0.613 & 0.635 & 0.723 & 0.722 & 0.549 & 0.556 & 0.799 \\
& KEEP & 0.743 & 0.764 & 0.605 & 0.629 & 0.675 & \bfseries 0.728 & 0.876 \\
& ResNet50& 0.564 & 0.583 & 0.534 & 0.595 & 0.595 & 0.551 & 0.762 \\
&UNI2-H & 0.681 & 0.714 & \bfseries 0.746 & \bfseries 0.731 & 0.575 & 0.657 & 0.849 \\

\addlinespace[3pt]

\multirow{4}{*}{\textbf{CTransPath}}
& Prov-GigaPath & 0.680 & 0.704 & 0.633 & 0.699 & 0.651 & 0.613 & 0.840 \\
& KEEP & 0.611 & 0.635 & 0.436 & 0.559 & 0.638 & 0.562 & 0.798 \\
& ResNet50 & 0.594 & 0.654 & 0.477 & 0.612 & 0.695 & 0.649 & 0.811 \\
&UNI2-H & 0.663 & 0.693 & 0.649 & 0.698 & 0.639 & 0.613 & 0.834 \\

\addlinespace[3pt]

\multirow{3}{*}{\textbf{Prov-GigaPath}}
& KEEP & 0.601 & 0.615 & 0.585 & 0.667 & 0.594 & 0.532 & 0.784 \\
& ResNet50 & 0.454 & 0.489 & 0.489 & 0.595 & 0.538 & 0.410 & 0.700 \\
&UNI2-H
& \bfseries 0.767 & \bfseries 0.781 & 0.500 & 0.612
& \bfseries 0.700 & 0.703 & \bfseries 0.883 \\

\addlinespace[3pt]

\multirow{2}{*}{\textbf{KEEP}}
& ResNet50 & 0.386 & 0.403 & 0.374 & 0.524 & 0.524 & 0.380 & 0.633 \\
&UNI2-H & 0.748 & 0.766 & 0.622 & 0.687 & 0.635 & 0.709 & 0.876 \\

\addlinespace[3pt]

\textbf{ResNet50}
&UNI2-H & 0.398 & 0.429 & 0.498 & 0.584 & 0.519 & 0.344 & 0.656 \\

\bottomrule
\end{tabular}
\end{table*}

\begin{table*}[!h]
\centering
\caption{\textbf{Patch-level representational similarity at 256 px and 20$\times$ on CPTAC.}
Bold denotes the highest value for each metric; ties are jointly highlighted.}
\label{tab:patch_similarity_256_20x}

\scriptsize
\setlength{\tabcolsep}{5.5pt}
\renewcommand{\arraystretch}{1.08}
\sisetup{table-format=1.3, detect-weight=true, detect-inline-weight=math}

\begin{tabular}{@{}ll*{7}{S}@{}}
\toprule
&&
\multicolumn{2}{c}{\textbf{CKA}} &
\multicolumn{2}{c}{\textbf{CCA}} &
\multicolumn{3}{c}{\textbf{Relational / Shape}} \\
\cmidrule(lr){3-4}
\cmidrule(lr){5-6}
\cmidrule(l){7-9}

\textbf{Encoder} &
\textbf{Compared with} &
{\textbf{Linear}} &
{\textbf{Kernel}} &
{\textbf{SVCCA}} &
{\textbf{PWCCA}} &
{\textbf{Proc.}} &
{\textbf{Cos.\ RSA}} &
{\textbf{dCor}} \\
\midrule

\multirow{4}{*}{\textbf{CTransPath}}
& Prov-GigaPath & 0.694 & 0.717 & 0.600 & \bfseries 0.674 & 0.618 & 0.645 & 0.847 \\
& KEEP & 0.565 & 0.588 & 0.407 & 0.536 & 0.609 & 0.511 & 0.767 \\
& ResNet50 & 0.578 & 0.643 & 0.460 & 0.599 & \bfseries 0.679 & 0.631 & 0.803 \\
&UNI2-H & 0.579 & 0.605 & \bfseries 0.611 & 0.671 & 0.583 & 0.516 & 0.777 \\

\addlinespace[3pt]

\multirow{3}{*}{\textbf{Prov-GigaPath}}
& KEEP & 0.606 & 0.621 & 0.575 & 0.655 & 0.572 & 0.540 & 0.788 \\
& ResNet50 & 0.425 & 0.470 & 0.481 & 0.587 & 0.506 & 0.408 & 0.685 \\
&UNI2-H
& \bfseries 0.774 & \bfseries 0.778 & 0.491 & 0.594
& 0.665 & \bfseries 0.717 & \bfseries 0.881 \\

\addlinespace[3pt]

\multirow{2}{*}{\textbf{KEEP}}
& ResNet50 & 0.342 & 0.361 & 0.371 & 0.515 & 0.496 & 0.336 & 0.598 \\
&UNI2-H & 0.641 & 0.652 & \bfseries 0.611 & 0.672 & 0.585 & 0.579 & 0.809 \\

\addlinespace[3pt]

\textbf{ResNet50}
&UNI2-H & 0.273 & 0.313 & 0.488 & 0.574 & 0.467 & 0.240 & 0.558 \\

\bottomrule
\end{tabular}
\end{table*}

\begin{table}[!h]
\centering
\caption{\textbf{Representational similarity between CONCH and CONCH v1.5 at 512 px on CPTAC.}
Results are reported across three magnifications; bold indicates the highest value across magnifications.}
\label{tab:patch_similarity_512}

\scriptsize
\setlength{\tabcolsep}{5pt}
\renewcommand{\arraystretch}{1.10}
\sisetup{table-format=1.3, detect-weight=true, detect-inline-weight=math}

\begin{tabular}{@{}c*{7}{S}@{}}
\toprule

&
\multicolumn{2}{c}{\textbf{CKA}} &
\multicolumn{2}{c}{\textbf{CCA}} &
\multicolumn{3}{c}{\textbf{Relational / Shape}} \\
\cmidrule(lr){2-3}
\cmidrule(lr){4-5}
\cmidrule(l){6-8}

\textbf{Mag.} &
{\textbf{Linear}} &
{\textbf{Kernel}} &
{\textbf{SVCCA}} &
{\textbf{PWCCA}} &
{\textbf{Proc.}} &
{\textbf{Cos.\ RSA}} &
{\textbf{dCor}} \\
\midrule

5$\times$
& 0.844 & 0.855 & 0.566 & 0.651 & 0.773 & 0.831 & 0.925 \\

10$\times$
& \bfseries 0.848
& \bfseries 0.859
& \bfseries 0.569
& \bfseries 0.658
& \bfseries 0.780
& \bfseries 0.833
& \bfseries 0.927 \\

20$\times$
& 0.834 & 0.845 & 0.560 & 0.654 & 0.765 & 0.814 & 0.920 \\

\bottomrule
\end{tabular}
\end{table}

\paragraph{Patch-level representational similarity on CPTAC.}
Tables~\ref{tab:patch_similarity_224}--\ref{tab:patch_similarity_512} examine how strongly different patch encoders agree across input resolutions and magnifications. Since the available encoder sets differ across resolutions, comparisons are most informative within each resolution and magnification rather than directly across tables.

\paragraph{224 px.}
At 224 px, \textbf{GPFM--H-optimus-0} exhibits the strongest overall correspondence, particularly under CKA, Procrustes similarity, and distance correlation. However, its agreement generally decreases from $5\times$ to $20\times$ (e.g., Linear CKA: $0.781 \rightarrow 0.674$; dCor: $0.886 \rightarrow 0.809$), suggesting greater divergence as increasingly fine morphological detail is represented. In contrast, \textbf{GPFM--Virchow2} shows the strongest SVCCA/PWCCA correspondence, illustrating that subspace-based similarity can identify relationships that differ from those captured by global geometric metrics.

\paragraph{256 px.}
At 256 px, the dominant encoder pair depends more strongly on magnification and metric. At $5\times$, \textbf{KEEP--UNI2-h} shows the strongest agreement for Linear/Kernel CKA, Cosine RSA, and dCor, whereas at $10\times$ and $20\times$, \textbf{Prov-GigaPath--UNI2-h} becomes the most consistently aligned pair, reaching Linear CKA values of $0.767$ and $0.774$, respectively. The identity of the strongest pair under SVCCA, PWCCA, and Procrustes similarity differs in several cases, again indicating that the metrics capture complementary aspects of representational geometry. Conversely, pairs involving ImageNet-pretrained ResNet50 generally exhibit weaker agreement, particularly with KEEP andUNI2-H, suggesting a larger representational gap between generic visual features and pathology-specific representations.

\paragraph{512 px.}
The controlled comparison between \textbf{CONCH} and \textbf{CONCH v1.5} at 512 px reveals consistently high similarity across all three magnifications. Their representations are remarkably stable from $5\times$ to $20\times$, with Linear CKA remaining between $0.834$ and $0.848$ and dCor between $0.920$ and $0.927$; similarity peaks slightly at $10\times$ across all seven metrics. This strong cross-magnification consistency indicates that the two closely related encoder versions preserve largely similar representational geometry despite model-version differences.

\paragraph{Overall summary.}
Across resolutions and magnifications, the analysis shows that representational similarity is structured rather than uniform: some encoder pairs remain consistently close, while others diverge substantially depending on scale and metric. Pathology-specific models generally exhibit stronger agreement with one another than with ImageNet-pretrained ResNet50, while the dominant pair can shift with magnification, as seen at 256 px. At 224 px, GPFM--H-optimus-0 shows the strongest overall geometric correspondence, whereas Prov-GigaPath--UNI2-h becomes particularly prominent at higher magnifications for 256 px. The CONCH--CONCH v1.5 comparison at 512 px is notably stable across all magnifications, indicating strong preservation of representational geometry between closely related model versions.

\begin{table*}[t]
\centering
\caption{\textbf{Patch-level representational similarity on TCGA at 224 px.}
Pairwise similarity is reported across $5\times$, $10\times$, and $20\times$ magnifications.
Bold denotes the highest value for each metric within each magnification.}
\label{tab:tcga_patch_similarity_224}

\scriptsize
\setlength{\tabcolsep}{5pt}
\renewcommand{\arraystretch}{1.08}
\sisetup{table-format=1.3, detect-weight=true, detect-inline-weight=math}

\begin{tabular}{@{}cll*{7}{S}@{}}
\toprule
&&&
\multicolumn{2}{c}{\textbf{CKA}} &
\multicolumn{2}{c}{\textbf{CCA}} &
\multicolumn{3}{c}{\textbf{Relational / Shape}} \\
\cmidrule(lr){4-5}
\cmidrule(lr){6-7}
\cmidrule(l){8-10}

\textbf{Mag.} &
\textbf{Encoder A} &
\textbf{Encoder B} &
{\textbf{Linear}} &
{\textbf{Kernel}} &
{\textbf{SVCCA}} &
{\textbf{PWCCA}} &
{\textbf{Proc.}} &
{\textbf{Cos.\ RSA}} &
{\textbf{dCor}} \\
\midrule

\multirow{3}{*}{$5\times$}
& GPFM & H-optimus-0
& \bfseries 0.811 & \bfseries 0.813 & 0.577 & 0.742
& \bfseries 0.751 & \bfseries 0.763 & \bfseries 0.902 \\
& GPFM & Virchow2
& 0.378 & 0.435 & \bfseries 0.599 & \bfseries 0.818
& 0.675 & 0.363 & 0.656 \\
& H-optimus-0 & Virchow2
& 0.343 & 0.390 & 0.491 & 0.752
& 0.634 & 0.325 & 0.622 \\

\midrule

\multirow{3}{*}{$10\times$}
& GPFM & H-optimus-0
& \bfseries 0.806 & \bfseries 0.817 & 0.588 & 0.734
& \bfseries 0.719 & \bfseries 0.725 & \bfseries 0.903 \\
& GPFM & Virchow2
& 0.383 & \bfseries 0.437 & \bfseries 0.603 & \bfseries 0.824
& 0.675 & 0.348 & 0.658 \\
& H-optimus-0 & Virchow2
& 0.385 & 0.434 & 0.510 & 0.763
& 0.659 & 0.354 & 0.655 \\

\midrule

\multirow{6}{*}{$20\times$}
& GPFM & H-optimus-0
& \bfseries 0.710 & \bfseries 0.728 & 0.590 & 0.720
& 0.668 & \bfseries 0.604 & \bfseries 0.852 \\
& GPFM & Virchow
& 0.496 & 0.514 & 0.600 & 0.811
& 0.646 & 0.416 & 0.719 \\
& GPFM & Virchow2
& 0.354 & 0.402 & \bfseries 0.604 & \bfseries 0.821
& 0.646 & 0.312 & 0.632 \\
& H-optimus-0 & Virchow
& 0.524 & 0.535 & 0.519 & 0.760
& 0.635 & 0.453 & 0.733 \\
& H-optimus-0 & Virchow2
& 0.374 & 0.415 & 0.533 & 0.765
& 0.634 & 0.321 & 0.641 \\
& Virchow & Virchow2
& 0.295 & 0.336 & 0.564 & 0.726
& \bfseries 0.669 & 0.272 & 0.582 \\

\bottomrule
\end{tabular}
\end{table*}

\begin{table*}[t]
\centering
\caption{\textbf{Patch-level representational similarity on TCGA at 256 px and $5\times$.}
Bold denotes the highest value for each metric.}
\label{tab:tcga_patch_similarity_256_5x}

\scriptsize
\setlength{\tabcolsep}{5.5pt}
\renewcommand{\arraystretch}{1.08}
\sisetup{table-format=1.3, detect-weight=true, detect-inline-weight=math}

\begin{tabular}{@{}ll*{7}{S}@{}}
\toprule
&&
\multicolumn{2}{c}{\textbf{CKA}} &
\multicolumn{2}{c}{\textbf{CCA}} &
\multicolumn{3}{c}{\textbf{Relational / Shape}} \\
\cmidrule(lr){3-4}
\cmidrule(lr){5-6}
\cmidrule(l){7-9}

\textbf{Encoder} &
\textbf{Compared with} &
{\textbf{Linear}} & {\textbf{Kernel}} &
{\textbf{SVCCA}} & {\textbf{PWCCA}} &
{\textbf{Proc.}} & {\textbf{Cos.\ RSA}} & {\textbf{dCor}} \\
\midrule

\multirow{4}{*}{\textbf{CTransPath}}
& Prov-GigaPath & 0.685 & 0.740 & 0.680 & \bfseries 0.733 & 0.706 & 0.715 & 0.863 \\
& KEEP & 0.665 & 0.686 & 0.469 & 0.573 & 0.658 & 0.612 & 0.831 \\
& ResNet50  & 0.773 & 0.793 & 0.501 & 0.621 & \bfseries 0.741 & \bfseries 0.752 & 0.894 \\
&UNI2-H & 0.726 & 0.754 & \bfseries 0.682 & 0.712 & 0.650 & 0.694 & 0.871 \\

\addlinespace[3pt]

\multirow{3}{*}{\textbf{Prov-GigaPath}}
& KEEP & 0.568 & 0.605 & 0.578 & 0.672 & 0.605 & 0.542 & 0.779 \\
& ResNet50  & 0.708 & 0.737 & 0.495 & 0.607 & 0.622 & 0.680 & 0.860 \\
&UNI2-H & 0.689 & 0.723 & 0.510 & 0.613 & 0.687 & 0.645 & 0.851 \\

\addlinespace[3pt]

\multirow{2}{*}{\textbf{KEEP}}
& ResNet50  & 0.575 & 0.587 & 0.372 & 0.525 & 0.562 & 0.532 & 0.768 \\
&UNI2-H
& \bfseries 0.781 & \bfseries 0.806 & 0.627 & 0.690
& 0.641 & 0.732 & \bfseries 0.901 \\

\addlinespace[3pt]

\textbf{ResNet50 }
&UNI2-H & 0.607 & 0.629 & 0.500 & 0.580 & 0.542 & 0.577 & 0.796 \\

\bottomrule
\end{tabular}
\end{table*}

\begin{table*}[t]
\centering
\caption{\textbf{Patch-level representational similarity on TCGA at 256 px and $10\times$.}
Bold denotes the highest value for each metric.}
\label{tab:tcga_patch_similarity_256_10x}

\scriptsize
\setlength{\tabcolsep}{5.5pt}
\renewcommand{\arraystretch}{1.08}
\sisetup{table-format=1.3, detect-weight=true, detect-inline-weight=math}

\begin{tabular}{@{}ll*{7}{S}@{}}
\toprule
&&
\multicolumn{2}{c}{\textbf{CKA}} &
\multicolumn{2}{c}{\textbf{CCA}} &
\multicolumn{3}{c}{\textbf{Relational / Shape}} \\
\cmidrule(lr){3-4}
\cmidrule(lr){5-6}
\cmidrule(l){7-9}

\textbf{Encoder} &
\textbf{Compared with} &
{\textbf{Linear}} & {\textbf{Kernel}} &
{\textbf{SVCCA}} & {\textbf{PWCCA}} &
{\textbf{Proc.}} & {\textbf{Cos.\ RSA}} & {\textbf{dCor}} \\
\midrule

\multirow{5}{*}{\textbf{CONCH}}
& CTransPath & 0.686 & 0.720 & 0.621 & 0.656 & 0.663 & 0.670 & 0.855 \\
& Prov-GigaPath & 0.598 & 0.622 & 0.732 & 0.725 & 0.529 & 0.552 & 0.793 \\
& KEEP & \bfseries 0.784 & \bfseries 0.803 & 0.614 & 0.636 & 0.686 & \bfseries 0.762 & \bfseries 0.898 \\
& ResNet50  & 0.667 & 0.691 & 0.529 & 0.593 & 0.610 & 0.648 & 0.834 \\
&UNI2-H & 0.672 & 0.706 & \bfseries 0.755 & \bfseries 0.731 & 0.528 & 0.651 & 0.847 \\

\addlinespace[3pt]

\multirow{4}{*}{\textbf{CTransPath}}
& Prov-GigaPath & 0.668 & 0.691 & 0.650 & 0.703 & 0.640 & 0.614 & 0.833 \\
& KEEP & 0.650 & 0.674 & 0.453 & 0.566 & 0.648 & 0.600 & 0.823 \\
& ResNet50  & 0.719 & 0.748 & 0.485 & 0.616 & \bfseries 0.720 & 0.717 & 0.867 \\
&UNI2-H & 0.680 & 0.708 & 0.664 & 0.696 & 0.612 & 0.638 & 0.844 \\

\addlinespace[3pt]

\multirow{3}{*}{\textbf{Prov-GigaPath}}
& KEEP & 0.588 & 0.602 & 0.590 & 0.669 & 0.580 & 0.515 & 0.776 \\
& ResNet50  & 0.626 & 0.641 & 0.497 & 0.594 & 0.549 & 0.580 & 0.802 \\
&UNI2-H & 0.644 & 0.667 & 0.506 & 0.605 & 0.663 & 0.564 & 0.815 \\

\addlinespace[3pt]

\multirow{2}{*}{\textbf{KEEP}}
& ResNet50  & 0.572 & 0.580 & 0.370 & 0.524 & 0.558 & 0.532 & 0.762 \\
&UNI2-H & 0.721 & 0.749 & 0.632 & 0.687 & 0.603 & 0.651 & 0.868 \\

\addlinespace[3pt]

\textbf{ResNet50 }
&UNI2-H & 0.538 & 0.563 & 0.506 & 0.577 & 0.504 & 0.508 & 0.752 \\

\bottomrule
\end{tabular}
\end{table*}

\begin{table*}[t]
\centering
\caption{\textbf{Patch-level representational similarity on TCGA at 256 px and $20\times$.}
Bold denotes the highest value for each metric.}
\label{tab:tcga_patch_similarity_256_20x}

\scriptsize
\setlength{\tabcolsep}{5.5pt}
\renewcommand{\arraystretch}{1.08}
\sisetup{table-format=1.3, detect-weight=true, detect-inline-weight=math}

\begin{tabular}{@{}ll*{7}{S}@{}}
\toprule
&&
\multicolumn{2}{c}{\textbf{CKA}} &
\multicolumn{2}{c}{\textbf{CCA}} &
\multicolumn{3}{c}{\textbf{Relational / Shape}} \\
\cmidrule(lr){3-4}
\cmidrule(lr){5-6}
\cmidrule(l){7-9}

\textbf{Encoder} &
\textbf{Compared with} &
{\textbf{Linear}} & {\textbf{Kernel}} &
{\textbf{SVCCA}} & {\textbf{PWCCA}} &
{\textbf{Proc.}} & {\textbf{Cos.\ RSA}} & {\textbf{dCor}} \\
\midrule

\multirow{4}{*}{\textbf{CTransPath}}
& Prov-GigaPath
& \bfseries 0.751 & 0.767 & 0.611 & \bfseries 0.679
& 0.622 & 0.731 & 0.878 \\
& KEEP
& 0.699 & 0.723 & 0.426 & 0.547 & 0.638 & 0.700 & 0.853 \\
& ResNet50 
& 0.735 & \bfseries 0.785 & 0.478 & 0.607
& \bfseries 0.726 & \bfseries 0.801 & \bfseries 0.887 \\
&UNI2-H
& 0.649 & 0.665 & \bfseries 0.629 & 0.672
& 0.570 & 0.605 & 0.818 \\

\addlinespace[3pt]

\multirow{3}{*}{\textbf{Prov-GigaPath}}
& KEEP
& 0.663 & 0.680 & 0.582 & 0.661 & 0.577 & 0.636 & 0.826 \\
& ResNet50 
& 0.616 & 0.649 & 0.484 & 0.589 & 0.535 & 0.635 & 0.807 \\
&UNI2-H
& 0.713 & 0.724 & 0.501 & 0.595 & 0.652 & 0.638 & 0.849 \\

\addlinespace[3pt]

\multirow{2}{*}{\textbf{KEEP}}
& ResNet50 
& 0.636 & 0.659 & 0.378 & 0.521 & 0.564 & 0.648 & 0.811 \\
&UNI2-H
& 0.609 & 0.628 & 0.621 & 0.674 & 0.558 & 0.578 & 0.795 \\

\addlinespace[3pt]

\textbf{ResNet50 }
&UNI2-H
& 0.495 & 0.526 & 0.502 & 0.577 & 0.487 & 0.505 & 0.727 \\

\bottomrule
\end{tabular}
\end{table*}

\begin{table}[t]
\centering
\caption{\textbf{Representational similarity between CONCH and CONCH v1.5 on TCGA at 512 px.}
Similarity is reported across three magnifications; bold denotes the highest value for each metric.}
\label{tab:tcga_patch_similarity_512}

\scriptsize
\setlength{\tabcolsep}{5pt}
\renewcommand{\arraystretch}{1.10}
\sisetup{table-format=1.3, detect-weight=true, detect-inline-weight=math}

\begin{tabular}{@{}c*{7}{S}@{}}
\toprule

&
\multicolumn{2}{c}{\textbf{CKA}} &
\multicolumn{2}{c}{\textbf{CCA}} &
\multicolumn{3}{c}{\textbf{Relational / Shape}} \\
\cmidrule(lr){2-3}
\cmidrule(lr){4-5}
\cmidrule(l){6-8}

\textbf{Mag.} &
{\textbf{Linear}} &
{\textbf{Kernel}} &
{\textbf{SVCCA}} &
{\textbf{PWCCA}} &
{\textbf{Proc.}} &
{\textbf{Cos.\ RSA}} &
{\textbf{dCor}} \\
\midrule

$5\times$
& \bfseries 0.888
& \bfseries 0.898
& 0.560
& 0.657
& 0.791
& \bfseries 0.897
& \bfseries 0.948 \\

$10\times$
& 0.870
& 0.879
& \bfseries 0.577
& \bfseries 0.666
& \bfseries 0.794
& 0.863
& 0.938 \\

$20\times$
& 0.861
& 0.872
& 0.574
& 0.665
& 0.784
& 0.852
& 0.935 \\

\bottomrule
\end{tabular}
\end{table}

\paragraph{Patch-level representational similarity.}
Tables~\ref{tab:tcga_patch_similarity_224}--\ref{tab:tcga_patch_similarity_512} summarize patch-level representational similarity on TCGA across multiple resolutions and magnifications.

\paragraph{224 px.}
At 224 px, \textbf{GPFM--H-optimus-0} shows the strongest overall geometric agreement across magnifications, with particularly high CKA, Cosine RSA, and dCor values. Its similarity nevertheless decreases at $20\times$ compared with $5\times$ and $10\times$, suggesting increasing representational divergence at finer visual scales. In contrast, \textbf{GPFM--Virchow2} consistently achieves the strongest SVCCA and PWCCA agreement.

\paragraph{256 px.}
At 256 px, the strongest pairing varies with magnification and metric. At $5\times$, \textbf{KEEP--UNI2-h} achieves the highest Linear/Kernel CKA and dCor, while at $10\times$, \textbf{CONCH--KEEP} dominates Linear/Kernel CKA, Cosine RSA, and dCor. At $20\times$, \textbf{CTransPath--Prov-GigaPath} and \textbf{CTransPath--ResNet50} exhibit the strongest correspondence across complementary metrics, highlighting scale-dependent shifts in encoder agreement.

\paragraph{512 px.}
The \textbf{CONCH--CONCH v1.5} comparison remains highly consistent across all magnifications, with Linear CKA ranging from 0.861--0.888 and dCor from 0.935--0.948. The strongest overall agreement occurs at $5\times$ for most metrics, indicating that the two closely related model versions preserve highly similar representational geometry across scales.

\clearpage
\subsection{Subcohort-level Patch Representational Similarity}\label{f3}
\begin{table*}[!h]
\centering
\caption{\textbf{Subcohort-level patch representational similarity on CPTAC at 224 px and $5\times$.}
Pairwise similarity is reported separately for BRCA, COAD, and LUAD.
Bold denotes the highest value within each subcohort and metric.}
\label{tab:cptac_patch_subcohort_224_5x}

\scriptsize
\setlength{\tabcolsep}{5.5pt}
\renewcommand{\arraystretch}{1.08}
\sisetup{table-format=1.3, detect-weight=true, detect-inline-weight=math}

\begin{tabular}{@{}lll*{7}{S}@{}}
\toprule
&&&
\multicolumn{2}{c}{\textbf{CKA}} &
\multicolumn{2}{c}{\textbf{CCA}} &
\multicolumn{3}{c}{\textbf{Relational / Shape}} \\
\cmidrule(lr){4-5}
\cmidrule(lr){6-7}
\cmidrule(l){8-10}

\textbf{Subcohort} &
\textbf{Encoder A} &
\textbf{Encoder B} &
{\textbf{Linear}} &
{\textbf{Kernel}} &
{\textbf{SVCCA}} &
{\textbf{PWCCA}} &
{\textbf{Proc.}} &
{\textbf{Cos.\ RSA}} &
{\textbf{dCor}} \\
\midrule

\multirow{3}{*}{\textbf{BRCA}}
& GPFM & H-optimus-0
& \bfseries 0.804 & \bfseries 0.803 & 0.614 & 0.784
& \bfseries 0.780 & \bfseries 0.768 & \bfseries 0.896 \\
& GPFM & Virchow2
& 0.468 & 0.528 & \bfseries 0.629 & \bfseries 0.879
& 0.714 & 0.469 & 0.724 \\
& H-optimus-0 & Virchow2
& 0.441 & 0.492 & 0.512 & 0.832
& 0.672 & 0.457 & 0.699 \\

\midrule

\multirow{3}{*}{\textbf{COAD}}
& GPFM & H-optimus-0
& \bfseries 0.851 & \bfseries 0.857 & 0.622 & 0.790
& \bfseries 0.792 & \bfseries 0.835 & \bfseries 0.926 \\
& GPFM & Virchow2
& 0.427 & 0.509 & \bfseries 0.632 & \bfseries 0.882
& 0.702 & 0.447 & 0.710 \\
& H-optimus-0 & Virchow2
& 0.400 & 0.463 & 0.508 & 0.836
& 0.658 & 0.414 & 0.678 \\

\midrule

\multirow{3}{*}{\textbf{LUAD}}
& GPFM & H-optimus-0
& \bfseries 0.773 & \bfseries 0.776 & 0.633 & 0.792
& \bfseries 0.769 & \bfseries 0.721 & \bfseries 0.881 \\
& GPFM & Virchow2
& 0.490 & 0.556 & \bfseries 0.642 & \bfseries 0.890
& 0.716 & 0.487 & 0.744 \\
& H-optimus-0 & Virchow2
& 0.436 & 0.487 & 0.526 & 0.838
& 0.664 & 0.424 & 0.695 \\

\bottomrule
\end{tabular}
\end{table*}

\begin{table*}[!h]
\centering
\caption{\textbf{Subcohort-level patch representational similarity on CPTAC at 224 px and $10\times$.}
Pairwise similarity is reported separately for BRCA, COAD, and LUAD.
Bold denotes the highest value within each subcohort and metric.}
\label{tab:cptac_patch_subcohort_224_10x}

\scriptsize
\setlength{\tabcolsep}{5.5pt}
\renewcommand{\arraystretch}{1.08}
\sisetup{table-format=1.3, detect-weight=true, detect-inline-weight=math}

\begin{tabular}{@{}lll*{7}{S}@{}}
\toprule
&&&
\multicolumn{2}{c}{\textbf{CKA}} &
\multicolumn{2}{c}{\textbf{CCA}} &
\multicolumn{3}{c}{\textbf{Relational / Shape}} \\
\cmidrule(lr){4-5}
\cmidrule(lr){6-7}
\cmidrule(l){8-10}

\textbf{Subcohort} &
\textbf{Encoder A} &
\textbf{Encoder B} &
{\textbf{Linear}} &
{\textbf{Kernel}} &
{\textbf{SVCCA}} &
{\textbf{PWCCA}} &
{\textbf{Proc.}} &
{\textbf{Cos.\ RSA}} &
{\textbf{dCor}} \\
\midrule

\multirow{3}{*}{\textbf{BRCA}}
& GPFM & H-optimus-0
& \bfseries 0.618 & \bfseries 0.619 & 0.629 & 0.788
& \bfseries 0.765 & \bfseries 0.523 & \bfseries 0.789 \\
& GPFM & Virchow2
& 0.449 & 0.504 & \bfseries 0.646 & \bfseries 0.886
& 0.729 & 0.436 & 0.710 \\
& H-optimus-0 & Virchow2
& 0.423 & 0.465 & 0.529 & 0.841
& 0.692 & 0.418 & 0.680 \\

\midrule

\multirow{3}{*}{\textbf{COAD}}
& GPFM & H-optimus-0
& \bfseries 0.736 & \bfseries 0.734 & 0.632 & 0.791
& \bfseries 0.776 & \bfseries 0.663 & \bfseries 0.857 \\
& GPFM & Virchow2
& 0.469 & 0.536 & \bfseries 0.643 & \bfseries 0.887
& 0.720 & 0.473 & 0.732 \\
& H-optimus-0 & Virchow2
& 0.380 & 0.421 & 0.519 & 0.839
& 0.676 & 0.364 & 0.646 \\

\midrule

\multirow{3}{*}{\textbf{LUAD}}
& GPFM & H-optimus-0
& \bfseries 0.790 & \bfseries 0.797 & \bfseries 0.664 & 0.800
& \bfseries 0.770 & \bfseries 0.754 & \bfseries 0.891 \\
& GPFM & Virchow2
& 0.443 & 0.493 & 0.662 & \bfseries 0.899
& 0.714 & 0.427 & 0.702 \\
& H-optimus-0 & Virchow2
& 0.468 & 0.520 & 0.588 & 0.850
& 0.701 & 0.457 & 0.720 \\

\bottomrule
\end{tabular}
\end{table*}

\begin{table*}[!h]
\centering
\caption{\textbf{Subcohort-level patch representational similarity on CPTAC at 224 px and $20\times$.}
Pairwise similarity is reported separately for BRCA, COAD, and LUAD.
Bold denotes the highest value within each subcohort and metric.}
\label{tab:cptac_patch_subcohort_224_20x}

\scriptsize
\setlength{\tabcolsep}{5.5pt}
\renewcommand{\arraystretch}{1.08}
\sisetup{table-format=1.3, detect-weight=true, detect-inline-weight=math}

\begin{tabular}{@{}lll*{7}{S}@{}}
\toprule
&&&
\multicolumn{2}{c}{\textbf{CKA}} &
\multicolumn{2}{c}{\textbf{CCA}} &
\multicolumn{3}{c}{\textbf{Relational / Shape}} \\
\cmidrule(lr){4-5}
\cmidrule(lr){6-7}
\cmidrule(l){8-10}

\textbf{Subcohort} &
\textbf{Encoder A} &
\textbf{Encoder B} &
{\textbf{Linear}} &
{\textbf{Kernel}} &
{\textbf{SVCCA}} &
{\textbf{PWCCA}} &
{\textbf{Proc.}} &
{\textbf{Cos.\ RSA}} &
{\textbf{dCor}} \\
\midrule

\multirow{6}{*}{\textbf{BRCA}}
& GPFM & H-optimus-0
& 0.601 & 0.594 & 0.626 & 0.776 & 0.736 & 0.494 & 0.771 \\
& GPFM & Virchow
& \bfseries 0.666 & \bfseries 0.688 & 0.631 & 0.883
& \bfseries 0.739 & \bfseries 0.682 & \bfseries 0.833 \\
& GPFM & Virchow2
& 0.471 & 0.517 & \bfseries 0.644 & \bfseries 0.889
& 0.722 & 0.430 & 0.719 \\
& H-optimus-0 & Virchow
& 0.499 & 0.496 & 0.546 & 0.844 & 0.707 & 0.440 & 0.704 \\
& H-optimus-0 & Virchow2
& 0.466 & 0.503 & 0.565 & 0.847 & 0.699 & 0.458 & 0.707 \\
& Virchow & Virchow2
& 0.335 & 0.379 & 0.569 & 0.826 & 0.675 & 0.337 & 0.618 \\

\midrule

\multirow{6}{*}{\textbf{COAD}}
& GPFM & H-optimus-0
& \bfseries 0.609 & \bfseries 0.603 & 0.637 & 0.783
& \bfseries 0.736 & \bfseries 0.514 & \bfseries 0.777 \\
& GPFM & Virchow
& 0.489 & 0.505 & 0.625 & 0.884 & 0.693 & 0.415 & 0.713 \\
& GPFM & Virchow2
& 0.470 & 0.526 & \bfseries 0.654 & \bfseries 0.893
& 0.720 & 0.465 & 0.725 \\
& H-optimus-0 & Virchow
& 0.402 & 0.397 & 0.543 & 0.840 & 0.667 & 0.309 & 0.630 \\
& H-optimus-0 & Virchow2
& 0.374 & 0.406 & 0.562 & 0.848 & 0.687 & 0.349 & 0.634 \\
& Virchow & Virchow2
& 0.326 & 0.367 & 0.539 & 0.820 & 0.663 & 0.313 & 0.608 \\

\midrule

\multirow{6}{*}{\textbf{LUAD}}
& GPFM & H-optimus-0
& \bfseries 0.844 & \bfseries 0.852 & 0.643 & 0.786
& \bfseries 0.761 & \bfseries 0.791 & \bfseries 0.922 \\
& GPFM & Virchow
& 0.534 & 0.616 & 0.640 & 0.891 & 0.721 & 0.545 & 0.775 \\
& GPFM & Virchow2
& 0.422 & 0.469 & \bfseries 0.646 & \bfseries 0.894
& 0.709 & 0.392 & 0.685 \\
& H-optimus-0 & Virchow
& 0.555 & 0.646 & 0.569 & 0.852 & 0.692 & 0.607 & 0.794 \\
& H-optimus-0 & Virchow2
& 0.442 & 0.488 & 0.602 & 0.851 & 0.689 & 0.409 & 0.699 \\
& Virchow & Virchow2
& 0.269 & 0.386 & 0.615 & 0.846 & 0.692 & 0.359 & 0.606 \\

\bottomrule
\end{tabular}
\end{table*}

\begin{table*}[t]
\centering
\caption{\textbf{Subcohort-level patch representational similarity on CPTAC at 256 px and $5\times$.}
Pairwise similarity is reported separately for BRCA, COAD, and LUAD.
Bold denotes the highest value within each subcohort and metric.}
\label{tab:cptac_patch_subcohort_256_5x}

\scriptsize
\setlength{\tabcolsep}{5.0pt}
\renewcommand{\arraystretch}{1.08}
\sisetup{table-format=1.3, detect-weight=true, detect-inline-weight=math}

\begin{tabular}{@{}lll*{7}{S}@{}}
\toprule
&&&
\multicolumn{2}{c}{\textbf{CKA}} &
\multicolumn{2}{c}{\textbf{CCA}} &
\multicolumn{3}{c}{\textbf{Relational / Shape}} \\
\cmidrule(lr){4-5}
\cmidrule(lr){6-7}
\cmidrule(l){8-10}

\textbf{Subcohort} &
\textbf{Encoder A} &
\textbf{Encoder B} &
{\textbf{Linear}} &
{\textbf{Kernel}} &
{\textbf{SVCCA}} &
{\textbf{PWCCA}} &
{\textbf{Proc.}} &
{\textbf{Cos.\ RSA}} &
{\textbf{dCor}} \\
\midrule

\multirow{10}{*}{\textbf{BRCA}}
& CTransPath & Prov-GigaPath
& 0.735 & 0.769 & 0.673 & \bfseries 0.756 & 0.708 & 0.757 & 0.878 \\
& CTransPath & KEEP
& 0.706 & 0.723 & 0.474 & 0.593 & 0.662 & 0.695 & 0.850 \\
& CTransPath & ResNet50 
& 0.753 & \bfseries 0.799 & 0.506 & 0.659
& \bfseries 0.756 & \bfseries 0.804 & \bfseries 0.894 \\
& CTransPath &UNI2-H
& 0.768 & 0.787 & \bfseries 0.698 & 0.754 & 0.692 & 0.779 & 0.888 \\
& Prov-GigaPath & KEEP
& 0.674 & 0.674 & 0.593 & 0.710 & 0.623 & 0.638 & 0.819 \\
& Prov-GigaPath & ResNet50 
& 0.750 & 0.751 & 0.545 & 0.673 & 0.649 & 0.754 & 0.866 \\
& Prov-GigaPath &UNI2-H
& \bfseries 0.788 & 0.788 & 0.541 & 0.664 & 0.718 & 0.769 & 0.887 \\
& KEEP & ResNet50 
& 0.625 & 0.626 & 0.398 & 0.581 & 0.583 & 0.607 & 0.789 \\
& KEEP &UNI2-H
& 0.751 & 0.759 & 0.646 & 0.732 & 0.656 & 0.727 & 0.870 \\
& ResNet50  &UNI2-H
& 0.691 & 0.688 & 0.562 & 0.660 & 0.606 & 0.696 & 0.829 \\

\midrule

\multirow{10}{*}{\textbf{COAD}}
& CTransPath & Prov-GigaPath
& 0.760 & \bfseries 0.794 & 0.676 & \bfseries 0.764
& 0.728 & 0.775 & \bfseries 0.893 \\
& CTransPath & KEEP
& 0.693 & 0.703 & 0.470 & 0.595 & 0.670 & 0.663 & 0.837 \\
& CTransPath & ResNet50 
& 0.747 & 0.792 & 0.503 & 0.652
& \bfseries 0.744 & \bfseries 0.786 & 0.891 \\
& CTransPath &UNI2-H
& 0.764 & 0.777 & \bfseries 0.692 & 0.751 & 0.688 & 0.758 & 0.882 \\
& Prov-GigaPath & KEEP
& 0.667 & 0.670 & 0.598 & 0.720 & 0.640 & 0.632 & 0.818 \\
& Prov-GigaPath & ResNet50 
& 0.757 & 0.760 & 0.530 & 0.673 & 0.656 & 0.735 & 0.871 \\
& Prov-GigaPath &UNI2-H
& \bfseries 0.786 & 0.792 & 0.541 & 0.665 & 0.723 & 0.757 & 0.890 \\
& KEEP & ResNet50 
& 0.571 & 0.574 & 0.402 & 0.578 & 0.576 & 0.549 & 0.755 \\
& KEEP &UNI2-H
& 0.777 & 0.783 & 0.647 & 0.733 & 0.667 & 0.756 & 0.884 \\
& ResNet50  &UNI2-H
& 0.645 & 0.653 & 0.550 & 0.655 & 0.585 & 0.625 & 0.808 \\

\midrule

\multirow{10}{*}{\textbf{LUAD}}
& CTransPath & Prov-GigaPath
& 0.647 & 0.705 & 0.697 & \bfseries 0.776 & 0.711 & 0.691 & 0.842 \\
& CTransPath & KEEP
& 0.601 & 0.621 & 0.478 & 0.608 & 0.656 & 0.569 & 0.789 \\
& CTransPath & ResNet50 
& 0.681 & 0.730 & 0.519 & 0.657 & \bfseries 0.726 & 0.724 & 0.855 \\
& CTransPath &UNI2-H
& 0.684 & 0.707 & \bfseries 0.704 & 0.757 & 0.666 & 0.634 & 0.841 \\
& Prov-GigaPath & KEEP
& 0.491 & 0.533 & 0.634 & 0.743 & 0.623 & 0.493 & 0.730 \\
& Prov-GigaPath & ResNet50 
& 0.668 & 0.673 & 0.533 & 0.672 & 0.627 & 0.633 & 0.821 \\
& Prov-GigaPath &UNI2-H
& 0.644 & 0.684 & 0.567 & 0.681 & 0.716 & 0.600 & 0.826 \\
& KEEP & ResNet50 
& 0.439 & 0.451 & 0.421 & 0.582 & 0.555 & 0.415 & 0.669 \\
& KEEP &UNI2-H
& \bfseries 0.783 & \bfseries 0.808 & 0.679 & 0.753
& 0.668 & \bfseries 0.790 & \bfseries 0.901 \\
& ResNet50  &UNI2-H
& 0.489 & 0.506 & 0.552 & 0.655 & 0.555 & 0.433 & 0.710 \\

\bottomrule
\end{tabular}
\end{table*}

\begin{table*}[t]
\centering
\caption{\textbf{Subcohort-level patch representational similarity on CPTAC at 256 px and $10\times$.}
Pairwise similarity is reported separately for BRCA, COAD, and LUAD.
Bold denotes the highest value within each subcohort and metric.}
\label{tab:cptac_patch_subcohort_256_10x}

\scriptsize
\setlength{\tabcolsep}{4.8pt}
\renewcommand{\arraystretch}{1.06}
\sisetup{table-format=1.3, detect-weight=true, detect-inline-weight=math}

\begin{tabular}{@{}lll*{7}{S}@{}}
\toprule
&&&
\multicolumn{2}{c}{\textbf{CKA}} &
\multicolumn{2}{c}{\textbf{CCA}} &
\multicolumn{3}{c}{\textbf{Relational / Shape}} \\
\cmidrule(lr){4-5}
\cmidrule(lr){6-7}
\cmidrule(l){8-10}

\textbf{Subcohort} &
\textbf{Encoder A} &
\textbf{Encoder B} &
{\textbf{Linear}} & {\textbf{Kernel}} &
{\textbf{SVCCA}} & {\textbf{PWCCA}} &
{\textbf{Proc.}} & {\textbf{Cos.\ RSA}} & {\textbf{dCor}} \\
\midrule

\multirow{15}{*}{\textbf{BRCA}}
& CONCH & CTransPath & 0.700 & 0.725 & 0.609 & 0.667 & 0.669 & 0.710 & 0.852 \\
& CONCH & Prov-GigaPath & 0.677 & 0.694 & 0.730 & 0.763 & 0.595 & 0.682 & 0.833 \\
& CONCH & KEEP & 0.726 & 0.745 & 0.600 & 0.641 & 0.643 & 0.722 & 0.864 \\
& CONCH & ResNet50  & 0.717 & 0.726 & 0.564 & 0.651 & 0.653 & 0.720 & 0.850 \\
& CONCH &UNI2-H & 0.702 & 0.722 & \bfseries 0.752 & \bfseries 0.768 & 0.603 & 0.727 & 0.851 \\
& CTransPath & Prov-GigaPath & 0.741 & 0.761 & 0.657 & 0.744 & 0.690 & 0.744 & 0.872 \\
& CTransPath & KEEP & 0.664 & 0.683 & 0.459 & 0.586 & 0.649 & 0.650 & 0.827 \\
& CTransPath & ResNet50  & 0.731 & 0.777 & 0.504 & 0.658 & \bfseries 0.747 & 0.781 & 0.881 \\
& CTransPath &UNI2-H & 0.711 & 0.736 & 0.679 & 0.748 & 0.678 & 0.736 & 0.859 \\
& Prov-GigaPath & KEEP & 0.697 & 0.702 & 0.606 & 0.715 & 0.635 & 0.656 & 0.836 \\
& Prov-GigaPath & ResNet50  & 0.664 & 0.687 & 0.550 & 0.675 & 0.625 & 0.692 & 0.828 \\
& Prov-GigaPath &UNI2-H & \bfseries 0.815 & \bfseries 0.824 & 0.535 & 0.668
& 0.735 & \bfseries 0.800 & \bfseries 0.907 \\
& KEEP & ResNet50  & 0.582 & 0.597 & 0.408 & 0.586 & 0.583 & 0.581 & 0.771 \\
& KEEP &UNI2-H & 0.721 & 0.734 & 0.643 & 0.732 & 0.654 & 0.702 & 0.856 \\
& ResNet50  &UNI2-H & 0.657 & 0.668 & 0.560 & 0.664 & 0.614 & 0.688 & 0.819 \\

\midrule

\multirow{15}{*}{\textbf{COAD}}
& CONCH & CTransPath & 0.706 & 0.728 & 0.618 & 0.670 & 0.679 & 0.697 & 0.854 \\
& CONCH & Prov-GigaPath & 0.705 & 0.717 & 0.732 & 0.764 & 0.613 & 0.679 & 0.846 \\
& CONCH & KEEP & 0.761 & 0.772 & 0.603 & 0.643 & 0.661 & 0.742 & 0.879 \\
& CONCH & ResNet50  & 0.669 & 0.682 & 0.558 & 0.644 & 0.631 & 0.638 & 0.823 \\
& CONCH &UNI2-H & 0.731 & 0.743 & \bfseries 0.746 & \bfseries 0.765 & 0.610 & 0.723 & 0.864 \\
& CTransPath & Prov-GigaPath & 0.745 & 0.765 & 0.655 & 0.746 & 0.699 & 0.722 & 0.875 \\
& CTransPath & KEEP & 0.669 & 0.683 & 0.455 & 0.589 & 0.660 & 0.628 & 0.826 \\
& CTransPath & ResNet50  & 0.699 & 0.745 & 0.493 & 0.648 & \bfseries 0.729 & 0.732 & 0.863 \\
& CTransPath &UNI2-H & 0.699 & 0.725 & 0.669 & 0.744 & 0.677 & 0.704 & 0.852 \\
& Prov-GigaPath & KEEP & 0.663 & 0.664 & 0.608 & 0.720 & 0.636 & 0.606 & 0.813 \\
& Prov-GigaPath & ResNet50  & 0.678 & 0.702 & 0.537 & 0.669 & 0.618 & 0.655 & 0.835 \\
& Prov-GigaPath &UNI2-H & \bfseries 0.778 & \bfseries 0.787 & 0.530 & 0.665
& 0.728 & \bfseries 0.749 & \bfseries 0.886 \\
& KEEP & ResNet50  & 0.558 & 0.562 & 0.406 & 0.580 & 0.570 & 0.516 & 0.747 \\
& KEEP &UNI2-H & 0.720 & 0.727 & 0.636 & 0.729 & 0.649 & 0.698 & 0.853 \\
& ResNet50  &UNI2-H & 0.617 & 0.632 & 0.543 & 0.655 & 0.584 & 0.579 & 0.795 \\

\midrule

\multirow{15}{*}{\textbf{LUAD}}
& CONCH & CTransPath & 0.649 & 0.683 & 0.632 & 0.681 & 0.663 & 0.647 & 0.829 \\
& CONCH & Prov-GigaPath & 0.558 & 0.579 & 0.764 & 0.779 & 0.547 & 0.507 & 0.763 \\
& CONCH & KEEP & \bfseries 0.787 & \bfseries 0.806 & 0.635 & 0.676
& 0.696 & \bfseries 0.792 & \bfseries 0.898 \\
& CONCH & ResNet50  & 0.562 & 0.574 & 0.564 & 0.643 & 0.590 & 0.536 & 0.755 \\
& CONCH &UNI2-H & 0.682 & 0.720 & \bfseries 0.785 & \bfseries 0.787 & 0.581 & 0.675 & 0.852 \\
& CTransPath & Prov-GigaPath & 0.677 & 0.700 & 0.679 & 0.752 & 0.663 & 0.642 & 0.837 \\
& CTransPath & KEEP & 0.613 & 0.637 & 0.464 & 0.604 & 0.655 & 0.587 & 0.800 \\
& CTransPath & ResNet50  & 0.605 & 0.663 & 0.512 & 0.655 & 0.706 & 0.662 & 0.814 \\
& CTransPath &UNI2-H & 0.683 & 0.710 & 0.692 & 0.752 & 0.654 & 0.642 & 0.843 \\
& Prov-GigaPath & KEEP & 0.532 & 0.550 & 0.651 & 0.738 & 0.605 & 0.471 & 0.743 \\
& Prov-GigaPath & ResNet50  & 0.511 & 0.538 & 0.549 & 0.665 & 0.554 & 0.477 & 0.733 \\
& Prov-GigaPath &UNI2-H & 0.699 & 0.723 & 0.558 & 0.681
& \bfseries 0.724 & 0.610 & 0.849 \\
& KEEP & ResNet50  & 0.432 & 0.444 & 0.420 & 0.581 & 0.545 & 0.415 & 0.662 \\
& KEEP &UNI2-H & 0.730 & 0.756 & 0.686 & 0.756 & 0.652 & 0.731 & 0.872 \\
& ResNet50  &UNI2-H & 0.451 & 0.475 & 0.559 & 0.658 & 0.538 & 0.393 & 0.688 \\

\bottomrule
\end{tabular}
\end{table*}

\begin{table*}[t]
\centering
\caption{\textbf{Subcohort-level patch representational similarity on CPTAC at 256 px and $20\times$.}
Pairwise similarity is reported separately for BRCA, COAD, and LUAD.
Bold denotes the highest value within each subcohort and metric.}
\label{tab:cptac_patch_subcohort_256_20x}

\scriptsize
\setlength{\tabcolsep}{5.0pt}
\renewcommand{\arraystretch}{1.08}
\sisetup{table-format=1.3, detect-weight=true, detect-inline-weight=math}

\begin{tabular}{@{}lll*{7}{S}@{}}
\toprule
&&&
\multicolumn{2}{c}{\textbf{CKA}} &
\multicolumn{2}{c}{\textbf{CCA}} &
\multicolumn{3}{c}{\textbf{Relational / Shape}} \\
\cmidrule(lr){4-5}
\cmidrule(lr){6-7}
\cmidrule(l){8-10}

\textbf{Subcohort} &
\textbf{Encoder A} &
\textbf{Encoder B} &
{\textbf{Linear}} & {\textbf{Kernel}} &
{\textbf{SVCCA}} & {\textbf{PWCCA}} &
{\textbf{Proc.}} & {\textbf{Cos.\ RSA}} & {\textbf{dCor}} \\
\midrule

\multirow{10}{*}{\textbf{BRCA}}
& CTransPath & Prov-GigaPath
& \bfseries 0.775 & 0.787 & 0.639 & 0.731 & 0.673 & 0.775 & 0.887 \\
& CTransPath & KEEP
& 0.668 & 0.685 & 0.441 & 0.570 & 0.632 & 0.636 & 0.828 \\
& CTransPath & ResNet50 
& 0.744 & \bfseries 0.790 & 0.498 & 0.650
& \bfseries 0.741 & \bfseries 0.787 & \bfseries 0.888 \\
& CTransPath &UNI2-H
& 0.645 & 0.655 & \bfseries 0.654 & \bfseries 0.732 & 0.642 & 0.608 & 0.810 \\
& Prov-GigaPath & KEEP
& 0.693 & 0.701 & 0.603 & 0.709 & 0.620 & 0.642 & 0.835 \\
& Prov-GigaPath & ResNet50 
& 0.651 & 0.686 & 0.544 & 0.670 & 0.608 & 0.689 & 0.826 \\
& Prov-GigaPath &UNI2-H
& 0.735 & 0.734 & 0.528 & 0.657 & 0.705 & 0.664 & 0.856 \\
& KEEP & ResNet50 
& 0.571 & 0.590 & 0.412 & 0.579 & 0.568 & 0.560 & 0.767 \\
& KEEP &UNI2-H
& 0.643 & 0.652 & 0.637 & 0.724 & 0.628 & 0.584 & 0.807 \\
& ResNet50  &UNI2-H
& 0.537 & 0.554 & 0.547 & 0.655 & 0.584 & 0.530 & 0.743 \\

\midrule

\multirow{10}{*}{\textbf{COAD}}
& CTransPath & Prov-GigaPath
& 0.754 & 0.770 & 0.633 & \bfseries 0.731 & 0.674 & 0.730 & 0.877 \\
& CTransPath & KEEP
& 0.649 & 0.660 & 0.432 & 0.576 & 0.642 & 0.596 & 0.812 \\
& CTransPath & ResNet50 
& 0.679 & 0.726 & 0.484 & 0.641
& \bfseries 0.715 & 0.707 & 0.850 \\
& CTransPath &UNI2-H
& 0.685 & 0.708 & \bfseries 0.644 & 0.730 & 0.654 & 0.675 & 0.843 \\
& Prov-GigaPath & KEEP
& 0.669 & 0.671 & 0.612 & 0.716 & 0.623 & 0.612 & 0.818 \\
& Prov-GigaPath & ResNet50 
& 0.627 & 0.657 & 0.536 & 0.666 & 0.592 & 0.613 & 0.807 \\
& Prov-GigaPath &UNI2-H
& \bfseries 0.792 & \bfseries 0.803 & 0.522 & 0.656
& 0.713 & \bfseries 0.765 & \bfseries 0.896 \\
& KEEP & ResNet50 
& 0.497 & 0.505 & 0.413 & 0.578 & 0.554 & 0.446 & 0.709 \\
& KEEP &UNI2-H
& 0.701 & 0.711 & 0.632 & 0.721 & 0.630 & 0.674 & 0.845 \\
& ResNet50  &UNI2-H
& 0.535 & 0.574 & 0.540 & 0.653 & 0.566 & 0.518 & 0.756 \\

\midrule

\multirow{10}{*}{\textbf{LUAD}}
& CTransPath & Prov-GigaPath
& \bfseries 0.716 & \bfseries 0.738 & 0.648 & 0.731
& 0.633 & \bfseries 0.682 & \bfseries 0.859 \\
& CTransPath & KEEP
& 0.584 & 0.610 & 0.437 & 0.580 & 0.630 & 0.548 & 0.781 \\
& CTransPath & ResNet50 
& 0.569 & 0.631 & 0.493 & 0.643 & \bfseries 0.689 & 0.626 & 0.795 \\
& CTransPath &UNI2-H
& 0.604 & 0.629 & 0.661 & 0.728 & 0.602 & 0.558 & 0.793 \\
& Prov-GigaPath & KEEP
& 0.574 & 0.592 & 0.640 & 0.726 & 0.588 & 0.516 & 0.770 \\
& Prov-GigaPath & ResNet50 
& 0.488 & 0.526 & 0.546 & 0.658 & 0.524 & 0.475 & 0.725 \\
& Prov-GigaPath &UNI2-H
& 0.696 & 0.718 & 0.550 & 0.664 & 0.688 & 0.621 & 0.844 \\
& KEEP & ResNet50 
& 0.379 & 0.398 & 0.417 & 0.574 & 0.521 & 0.365 & 0.626 \\
& KEEP &UNI2-H
& 0.584 & 0.614 & \bfseries 0.674 & \bfseries 0.741 & 0.596 & 0.571 & 0.787 \\
& ResNet50  &UNI2-H
& 0.329 & 0.360 & 0.557 & 0.650 & 0.487 & 0.317 & 0.597 \\

\bottomrule
\end{tabular}
\end{table*}

\begin{table*}[t]
\centering
\caption{\textbf{Subcohort-level representational similarity between CONCH and CONCH v1.5 on CPTAC at 512 px.}
Similarity is reported for BRCA, COAD, and LUAD across $5\times$, $10\times$, and $20\times$ magnifications.
Bold denotes the highest value within each subcohort and metric across magnifications.}
\label{tab:cptac_patch_subcohort_512}

\scriptsize
\setlength{\tabcolsep}{6pt}
\renewcommand{\arraystretch}{1.10}
\sisetup{
    table-format=1.3,
    detect-weight=true,
    detect-inline-weight=math
}

\begin{tabular}{@{}lc*{7}{S}@{}}
\toprule

&&
\multicolumn{2}{c}{\textbf{CKA}} &
\multicolumn{2}{c}{\textbf{CCA}} &
\multicolumn{3}{c}{\textbf{Relational / Shape}} \\
\cmidrule(lr){3-4}
\cmidrule(lr){5-6}
\cmidrule(l){7-9}

\textbf{Subcohort} &
\textbf{Mag.} &
{\textbf{Linear}} &
{\textbf{Kernel}} &
{\textbf{SVCCA}} &
{\textbf{PWCCA}} &
{\textbf{Proc.}} &
{\textbf{Cos.\ RSA}} &
{\textbf{dCor}} \\
\midrule

\multirow{3}{*}{\textbf{BRCA}}
& $5\times$
& 0.826 & 0.832 & 0.541 & 0.657 & 0.730 & 0.816 & 0.911 \\
& $10\times$
& \bfseries 0.834
& \bfseries 0.843
& \bfseries 0.544
& \bfseries 0.664
& \bfseries 0.738
& \bfseries 0.832
& \bfseries 0.918 \\
& $20\times$
& 0.821 & 0.829 & 0.543 & 0.661 & 0.724 & 0.812 & 0.910 \\

\midrule

\multirow{3}{*}{\textbf{COAD}}
& $5\times$
& \bfseries 0.849
& \bfseries 0.853
& 0.545
& 0.662
& \bfseries 0.758
& \bfseries 0.840
& \bfseries 0.923 \\
& $10\times$
& 0.835 & 0.843 & 0.544 & 0.664 & 0.753 & 0.828 & 0.918 \\
& $20\times$
& 0.812
& 0.824
& \bfseries 0.553
& \bfseries 0.667
& 0.744
& 0.801
& 0.907 \\

\midrule

\multirow{3}{*}{\textbf{LUAD}}
& $5\times$
& \bfseries 0.875
& \bfseries 0.884
& 0.560
& 0.688
& 0.786
& 0.869
& \bfseries 0.940 \\
& $10\times$
& 0.874
& 0.882
& \bfseries 0.573
& \bfseries 0.696
& \bfseries 0.789
& \bfseries 0.873
& 0.939 \\
& $20\times$
& 0.856 & 0.867 & 0.559 & 0.692 & 0.774 & 0.854 & 0.931 \\

\bottomrule
\end{tabular}
\end{table*}

\begin{table*}[t]
\centering
\caption{\textbf{Subcohort-level patch representational similarity on TCGA at 224 px and $5\times$.}
Pairwise similarity is reported separately for BRCA, LUAD, and LUSC.
Bold denotes the highest value within each subcohort and metric.}
\label{tab:tcga_patch_subcohort_224_5x}

\scriptsize
\setlength{\tabcolsep}{5.5pt}
\renewcommand{\arraystretch}{1.08}
\sisetup{table-format=1.3, detect-weight=true, detect-inline-weight=math}

\begin{tabular}{@{}lll*{7}{S}@{}}
\toprule
&&&
\multicolumn{2}{c}{\textbf{CKA}} &
\multicolumn{2}{c}{\textbf{CCA}} &
\multicolumn{3}{c}{\textbf{Relational / Shape}} \\
\cmidrule(lr){4-5}
\cmidrule(lr){6-7}
\cmidrule(l){8-10}

\textbf{Subcohort} &
\textbf{Encoder A} &
\textbf{Encoder B} &
{\textbf{Linear}} &
{\textbf{Kernel}} &
{\textbf{SVCCA}} &
{\textbf{PWCCA}} &
{\textbf{Proc.}} &
{\textbf{Cos.\ RSA}} &
{\textbf{dCor}} \\
\midrule

\multirow{3}{*}{\textbf{BRCA}}
& GPFM & H-optimus-0
& \bfseries 0.860 & \bfseries 0.862 & 0.630 & 0.793
& \bfseries 0.778 & \bfseries 0.843 & \bfseries 0.928 \\
& GPFM & Virchow2
& 0.403 & 0.461 & \bfseries 0.643 & \bfseries 0.883
& 0.697 & 0.413 & 0.675 \\
& H-optimus-0 & Virchow2
& 0.387 & 0.438 & 0.536 & 0.841
& 0.665 & 0.396 & 0.659 \\

\midrule

\multirow{3}{*}{\textbf{LUAD}}
& GPFM & H-optimus-0
& \bfseries 0.759 & \bfseries 0.760 & 0.620 & 0.790
& \bfseries 0.763 & \bfseries 0.690 & \bfseries 0.871 \\
& GPFM & Virchow2
& 0.448 & 0.511 & \bfseries 0.631 & \bfseries 0.883
& 0.701 & 0.413 & 0.712 \\
& H-optimus-0 & Virchow2
& 0.405 & 0.457 & 0.520 & 0.836
& 0.659 & 0.372 & 0.671 \\

\midrule

\multirow{3}{*}{\textbf{LUSC}}
& GPFM & H-optimus-0
& \bfseries 0.762 & \bfseries 0.760 & 0.615 & 0.788
& \bfseries 0.756 & \bfseries 0.709 & \bfseries 0.871 \\
& GPFM & Virchow2
& 0.405 & 0.469 & \bfseries 0.630 & \bfseries 0.879
& 0.684 & 0.393 & 0.680 \\
& H-optimus-0 & Virchow2
& 0.372 & 0.426 & 0.528 & 0.834
& 0.644 & 0.365 & 0.648 \\

\bottomrule
\end{tabular}
\end{table*}

\begin{table*}[t]
\centering
\caption{\textbf{Subcohort-level patch representational similarity on TCGA at 224 px and $10\times$.}
Pairwise similarity is reported separately for BRCA, LUAD, and LUSC.
Bold denotes the highest value within each subcohort and metric.}
\label{tab:tcga_patch_subcohort_224_10x}

\scriptsize
\setlength{\tabcolsep}{5.5pt}
\renewcommand{\arraystretch}{1.08}
\sisetup{table-format=1.3, detect-weight=true, detect-inline-weight=math}

\begin{tabular}{@{}lll*{7}{S}@{}}
\toprule
&&&
\multicolumn{2}{c}{\textbf{CKA}} &
\multicolumn{2}{c}{\textbf{CCA}} &
\multicolumn{3}{c}{\textbf{Relational / Shape}} \\
\cmidrule(lr){4-5}
\cmidrule(lr){6-7}
\cmidrule(l){8-10}

\textbf{Subcohort} &
\textbf{Encoder A} &
\textbf{Encoder B} &
{\textbf{Linear}} &
{\textbf{Kernel}} &
{\textbf{SVCCA}} &
{\textbf{PWCCA}} &
{\textbf{Proc.}} &
{\textbf{Cos.\ RSA}} &
{\textbf{dCor}} \\
\midrule

\multirow{3}{*}{\textbf{BRCA}}
& GPFM & H-optimus-0
& \bfseries 0.842 & \bfseries 0.852 & 0.640 & 0.786
& \bfseries 0.749 & \bfseries 0.808 & \bfseries 0.922 \\
& GPFM & Virchow2
& 0.426 & 0.479 & \bfseries 0.644 & \bfseries 0.886
& 0.698 & 0.415 & 0.688 \\
& H-optimus-0 & Virchow2
& 0.427 & 0.474 & 0.574 & 0.846
& 0.687 & 0.422 & 0.684 \\

\midrule

\multirow{3}{*}{\textbf{LUAD}}
& GPFM & H-optimus-0
& \bfseries 0.770 & \bfseries 0.780 & 0.638 & 0.785
& \bfseries 0.751 & \bfseries 0.699 & \bfseries 0.882 \\
& GPFM & Virchow2
& 0.433 & 0.487 & \bfseries 0.641 & \bfseries 0.888
& 0.706 & 0.404 & 0.696 \\
& H-optimus-0 & Virchow2
& 0.449 & 0.499 & 0.563 & 0.841
& 0.684 & 0.413 & 0.704 \\

\midrule

\multirow{3}{*}{\textbf{LUSC}}
& GPFM & H-optimus-0
& \bfseries 0.743 & \bfseries 0.752 & 0.636 & 0.785
& \bfseries 0.742 & \bfseries 0.687 & \bfseries 0.864 \\
& GPFM & Virchow2
& 0.381 & 0.436 & \bfseries 0.644 & \bfseries 0.886
& 0.689 & 0.360 & 0.657 \\
& H-optimus-0 & Virchow2
& 0.421 & 0.471 & 0.557 & 0.839
& 0.680 & 0.399 & 0.682 \\

\bottomrule
\end{tabular}
\end{table*}

\begin{table*}[t]
\centering
\caption{\textbf{Subcohort-level patch representational similarity on TCGA at 224 px and $20\times$.}
Pairwise similarity is reported separately for BRCA, LUAD, and LUSC.
Bold denotes the highest value within each subcohort and metric.}
\label{tab:tcga_patch_subcohort_224_20x}

\scriptsize
\setlength{\tabcolsep}{5.5pt}
\renewcommand{\arraystretch}{1.08}
\sisetup{table-format=1.3, detect-weight=true, detect-inline-weight=math}

\begin{tabular}{@{}lll*{7}{S}@{}}
\toprule
&&&
\multicolumn{2}{c}{\textbf{CKA}} &
\multicolumn{2}{c}{\textbf{CCA}} &
\multicolumn{3}{c}{\textbf{Relational / Shape}} \\
\cmidrule(lr){4-5}
\cmidrule(lr){6-7}
\cmidrule(l){8-10}

\textbf{Subcohort} &
\textbf{Encoder A} &
\textbf{Encoder B} &
{\textbf{Linear}} &
{\textbf{Kernel}} &
{\textbf{SVCCA}} &
{\textbf{PWCCA}} &
{\textbf{Proc.}} &
{\textbf{Cos.\ RSA}} &
{\textbf{dCor}} \\
\midrule

\multirow{6}{*}{\textbf{BRCA}}
& GPFM & H-optimus-0
& \bfseries 0.746 & \bfseries 0.768 & 0.643 & 0.775
& \bfseries 0.703 & \bfseries 0.714 & \bfseries 0.875 \\
& GPFM & Virchow
& 0.536 & 0.567 & \bfseries 0.650 & 0.880
& 0.682 & 0.524 & 0.755 \\
& GPFM & Virchow2
& 0.416 & 0.467 & 0.648 & \bfseries 0.884
& 0.673 & 0.417 & 0.681 \\
& H-optimus-0 & Virchow
& 0.550 & 0.574 & 0.556 & 0.842
& 0.676 & 0.538 & 0.759 \\
& H-optimus-0 & Virchow2
& 0.436 & 0.480 & 0.589 & 0.843
& 0.666 & 0.420 & 0.689 \\
& Virchow & Virchow2
& 0.355 & 0.409 & 0.619 & 0.829
& 0.702 & 0.372 & 0.641 \\

\midrule

\multirow{6}{*}{\textbf{LUAD}}
& GPFM & H-optimus-0
& \bfseries 0.726 & \bfseries 0.733 & 0.647 & 0.778
& \bfseries 0.712 & \bfseries 0.611 & \bfseries 0.854 \\
& GPFM & Virchow
& 0.460 & 0.471 & 0.639 & 0.881
& 0.663 & 0.400 & 0.690 \\
& GPFM & Virchow2
& 0.374 & 0.418 & \bfseries 0.648 & \bfseries 0.888
& 0.674 & 0.365 & 0.644 \\
& H-optimus-0 & Virchow
& 0.559 & 0.566 & 0.558 & 0.842
& 0.654 & 0.479 & 0.757 \\
& H-optimus-0 & Virchow2
& 0.429 & 0.472 & 0.594 & 0.845
& 0.668 & 0.378 & 0.685 \\
& Virchow & Virchow2
& 0.284 & 0.325 & 0.599 & 0.831
& 0.673 & 0.267 & 0.572 \\

\midrule

\multirow{6}{*}{\textbf{LUSC}}
& GPFM & H-optimus-0
& \bfseries 0.669 & \bfseries 0.677 & 0.641 & 0.776
& \bfseries 0.704 & \bfseries 0.574 & \bfseries 0.816 \\
& GPFM & Virchow
& 0.411 & 0.426 & 0.642 & 0.881
& 0.656 & 0.347 & 0.655 \\
& GPFM & Virchow2
& 0.329 & 0.372 & \bfseries 0.647 & \bfseries 0.887
& 0.663 & 0.337 & 0.606 \\
& H-optimus-0 & Virchow
& 0.507 & 0.513 & 0.552 & 0.841
& 0.652 & 0.425 & 0.717 \\
& H-optimus-0 & Virchow2
& 0.398 & 0.439 & 0.589 & 0.843
& 0.662 & 0.360 & 0.658 \\
& Virchow & Virchow2
& 0.290 & 0.330 & 0.597 & 0.827
& 0.675 & 0.277 & 0.576 \\

\bottomrule
\end{tabular}
\end{table*}

\begin{table*}[t]
\centering
\caption{\textbf{Subcohort-level patch representational similarity on TCGA at 256 px and $5\times$.}
Pairwise similarity is reported separately for BRCA, LUAD, and LUSC.
Bold denotes the highest value within each subcohort and metric.}
\label{tab:tcga_patch_subcohort_256_5x}

\scriptsize
\setlength{\tabcolsep}{5.0pt}
\renewcommand{\arraystretch}{1.08}
\sisetup{table-format=1.3, detect-weight=true, detect-inline-weight=math}

\begin{tabular}{@{}lll*{7}{S}@{}}
\toprule
&&&
\multicolumn{2}{c}{\textbf{CKA}} &
\multicolumn{2}{c}{\textbf{CCA}} &
\multicolumn{3}{c}{\textbf{Relational / Shape}} \\
\cmidrule(lr){4-5}
\cmidrule(lr){6-7}
\cmidrule(l){8-10}

\textbf{Subcohort} &
\textbf{Encoder A} &
\textbf{Encoder B} &
{\textbf{Linear}} & {\textbf{Kernel}} &
{\textbf{SVCCA}} & {\textbf{PWCCA}} &
{\textbf{Proc.}} & {\textbf{Cos.\ RSA}} & {\textbf{dCor}} \\
\midrule

\multirow{10}{*}{\textbf{BRCA}}
& CTransPath & Prov-GigaPath
& 0.751 & 0.796 & 0.720 & \bfseries 0.784 & 0.728 & 0.786 & 0.894 \\
& CTransPath & KEEP
& 0.736 & 0.754 & 0.507 & 0.617 & 0.684 & 0.723 & 0.871 \\
& CTransPath & ResNet50 
& \bfseries 0.851 & \bfseries 0.862 & 0.532 & 0.673
& \bfseries 0.772 & \bfseries 0.852 & \bfseries 0.930 \\
& CTransPath &UNI2-H
& 0.802 & 0.821 & \bfseries 0.733 & 0.771 & 0.683 & 0.790 & 0.908 \\
& Prov-GigaPath & KEEP
& 0.655 & 0.683 & 0.631 & 0.739 & 0.639 & 0.660 & 0.827 \\
& Prov-GigaPath & ResNet50 
& 0.737 & 0.767 & 0.566 & 0.686 & 0.651 & 0.736 & 0.878 \\
& Prov-GigaPath &UNI2-H
& 0.777 & 0.796 & 0.573 & 0.685 & 0.729 & 0.754 & 0.892 \\
& KEEP & ResNet50 
& 0.688 & 0.701 & 0.410 & 0.595 & 0.608 & 0.688 & 0.839 \\
& KEEP &UNI2-H
& 0.823 & 0.839 & 0.683 & 0.756 & 0.670 & 0.825 & 0.918 \\
& ResNet50  &UNI2-H
& 0.740 & 0.748 & 0.586 & 0.669 & 0.595 & 0.732 & 0.867 \\

\midrule

\multirow{10}{*}{\textbf{LUAD}}
& CTransPath & Prov-GigaPath
& 0.653 & 0.708 & \bfseries 0.712 & \bfseries 0.781
& 0.717 & 0.699 & 0.844 \\
& CTransPath & KEEP
& 0.546 & 0.566 & 0.489 & 0.604 & 0.636 & 0.503 & 0.754 \\
& CTransPath & ResNet50 
& 0.674 & 0.716 & 0.536 & 0.658
& \bfseries 0.722 & \bfseries 0.717 & 0.849 \\
& CTransPath &UNI2-H
& 0.635 & 0.662 & \bfseries 0.712 & 0.757 & 0.648 & 0.584 & 0.815 \\
& Prov-GigaPath & KEEP
& 0.514 & 0.540 & 0.626 & 0.735 & 0.614 & 0.481 & 0.733 \\
& Prov-GigaPath & ResNet50 
& 0.718 & 0.715 & 0.539 & 0.677 & 0.633 & 0.650 & 0.847 \\
& Prov-GigaPath &UNI2-H
& 0.642 & 0.684 & 0.562 & 0.675 & 0.709 & 0.589 & 0.826 \\
& KEEP & ResNet50 
& 0.462 & 0.466 & 0.426 & 0.581 & 0.541 & 0.413 & 0.679 \\
& KEEP &UNI2-H
& \bfseries 0.749 & \bfseries 0.771 & 0.666 & 0.742
& 0.648 & 0.716 & \bfseries 0.880 \\
& ResNet50  &UNI2-H
& 0.514 & 0.530 & 0.551 & 0.652 & 0.545 & 0.452 & 0.727 \\

\midrule

\multirow{10}{*}{\textbf{LUSC}}
& CTransPath & Prov-GigaPath
& 0.646 & 0.705 & 0.706 & \bfseries 0.776
& 0.707 & 0.703 & 0.840 \\
& CTransPath & KEEP
& 0.555 & 0.569 & 0.479 & 0.596 & 0.630 & 0.504 & 0.754 \\
& CTransPath & ResNet50 
& 0.704 & 0.745 & 0.528 & 0.655
& \bfseries 0.726 & 0.746 & 0.865 \\
& CTransPath &UNI2-H
& 0.643 & 0.662 & \bfseries 0.708 & 0.753 & 0.643 & 0.591 & 0.814 \\
& Prov-GigaPath & KEEP
& 0.493 & 0.528 & 0.618 & 0.727 & 0.603 & 0.497 & 0.723 \\
& Prov-GigaPath & ResNet50 
& 0.715 & 0.720 & 0.537 & 0.673 & 0.624 & 0.663 & 0.850 \\
& Prov-GigaPath &UNI2-H
& 0.625 & 0.666 & 0.562 & 0.671 & 0.700 & 0.612 & 0.813 \\
& KEEP & ResNet50 
& 0.448 & 0.458 & 0.416 & 0.574 & 0.532 & 0.413 & 0.674 \\
& KEEP &UNI2-H
& \bfseries 0.787 & \bfseries 0.804 & 0.663 & 0.737
& 0.643 & \bfseries 0.758 & \bfseries 0.898 \\
& ResNet50  &UNI2-H
& 0.509 & 0.526 & 0.553 & 0.650 & 0.539 & 0.469 & 0.724 \\

\bottomrule
\end{tabular}
\end{table*}

\begin{table*}[t]
\centering
\caption{\textbf{Subcohort-level patch representational similarity on TCGA at 256 px and $10\times$.}
Pairwise similarity is reported separately for BRCA, LUAD, and LUSC.
Bold denotes the highest value within each subcohort and metric.}
\label{tab:tcga_patch_subcohort_256_10x}

\scriptsize
\setlength{\tabcolsep}{4.8pt}
\renewcommand{\arraystretch}{1.06}
\sisetup{table-format=1.3, detect-weight=true, detect-inline-weight=math}

\begin{tabular}{@{}lll*{7}{S}@{}}
\toprule
&&&
\multicolumn{2}{c}{\textbf{CKA}} &
\multicolumn{2}{c}{\textbf{CCA}} &
\multicolumn{3}{c}{\textbf{Relational / Shape}} \\
\cmidrule(lr){4-5}
\cmidrule(lr){6-7}
\cmidrule(l){8-10}

\textbf{Subcohort} &
\textbf{Encoder A} &
\textbf{Encoder B} &
{\textbf{Linear}} & {\textbf{Kernel}} &
{\textbf{SVCCA}} & {\textbf{PWCCA}} &
{\textbf{Proc.}} & {\textbf{Cos.\ RSA}} & {\textbf{dCor}} \\
\midrule

\multirow{15}{*}{\textbf{BRCA}}
& CONCH & CTransPath & 0.754 & 0.784 & 0.642 & 0.696 & 0.694 & 0.757 & 0.890 \\
& CONCH & Prov-GigaPath & 0.646 & 0.668 & 0.770 & 0.781 & 0.557 & 0.633 & 0.822 \\
& CONCH & KEEP
& \bfseries 0.825 & \bfseries 0.843 & 0.644 & 0.675
& 0.690 & \bfseries 0.831 & \bfseries 0.920 \\
& CONCH & ResNet50 
& 0.740 & 0.765 & 0.576 & 0.656 & 0.649 & 0.749 & 0.877 \\
& CONCH &UNI2-H
& 0.738 & 0.762 & \bfseries 0.795 & \bfseries 0.787
& 0.547 & 0.749 & 0.879 \\
& CTransPath & Prov-GigaPath
& 0.723 & 0.740 & 0.698 & 0.760 & 0.669 & 0.696 & 0.862 \\
& CTransPath & KEEP
& 0.731 & 0.750 & 0.493 & 0.610 & 0.676 & 0.718 & 0.868 \\
& CTransPath & ResNet50 
& 0.808 & 0.831 & 0.524 & 0.668
& \bfseries 0.755 & 0.823 & 0.912 \\
& CTransPath &UNI2-H
& 0.754 & 0.771 & 0.717 & 0.756 & 0.638 & 0.740 & 0.881 \\
& Prov-GigaPath & KEEP
& 0.655 & 0.668 & 0.644 & 0.736 & 0.616 & 0.622 & 0.818 \\
& Prov-GigaPath & ResNet50 
& 0.667 & 0.682 & 0.568 & 0.674 & 0.588 & 0.656 & 0.827 \\
& Prov-GigaPath &UNI2-H
& 0.730 & 0.744 & 0.566 & 0.675 & 0.705 & 0.683 & 0.861 \\
& KEEP & ResNet50 
& 0.704 & 0.711 & 0.415 & 0.591 & 0.607 & 0.695 & 0.843 \\
& KEEP &UNI2-H
& 0.770 & 0.786 & 0.689 & 0.752 & 0.631 & 0.756 & 0.889 \\
& ResNet50  &UNI2-H
& 0.672 & 0.684 & 0.584 & 0.662 & 0.552 & 0.673 & 0.829 \\

\midrule

\multirow{15}{*}{\textbf{LUAD}}
& CONCH & CTransPath
& 0.574 & 0.612 & 0.626 & 0.679 & 0.643 & 0.569 & 0.785 \\
& CONCH & Prov-GigaPath
& 0.572 & 0.594 & 0.757 & 0.772 & 0.537 & 0.524 & 0.772 \\
& CONCH & KEEP
& \bfseries 0.722 & \bfseries 0.740 & 0.624 & 0.665
& 0.672 & \bfseries 0.700 & \bfseries 0.861 \\
& CONCH & ResNet50 
& 0.570 & 0.584 & 0.562 & 0.640 & 0.582 & 0.528 & 0.763 \\
& CONCH &UNI2-H
& 0.640 & 0.675 & \bfseries 0.777 & \bfseries 0.777
& 0.551 & 0.614 & 0.825 \\
& CTransPath & Prov-GigaPath
& 0.645 & 0.678 & 0.687 & 0.755 & 0.654 & 0.619 & 0.824 \\
& CTransPath & KEEP
& 0.544 & 0.573 & 0.476 & 0.601 & 0.633 & 0.511 & 0.758 \\
& CTransPath & ResNet50 
& 0.599 & 0.652 & 0.522 & 0.655
& \bfseries 0.705 & 0.657 & 0.808 \\
& CTransPath &UNI2-H
& 0.636 & 0.662 & 0.700 & 0.749 & 0.631 & 0.587 & 0.816 \\
& Prov-GigaPath & KEEP
& 0.535 & 0.551 & 0.640 & 0.731 & 0.592 & 0.469 & 0.741 \\
& Prov-GigaPath & ResNet50 
& 0.583 & 0.612 & 0.552 & 0.665 & 0.562 & 0.538 & 0.781 \\
& Prov-GigaPath &UNI2-H
& 0.643 & 0.667 & 0.557 & 0.672 & 0.700 & 0.556 & 0.813 \\
& KEEP & ResNet50 
& 0.457 & 0.465 & 0.422 & 0.581 & 0.540 & 0.416 & 0.678 \\
& KEEP &UNI2-H
& 0.695 & 0.720 & 0.674 & 0.743 & 0.625 & 0.657 & 0.849 \\
& ResNet50  &UNI2-H
& 0.468 & 0.495 & 0.561 & 0.653 & 0.526 & 0.408 & 0.702 \\

\midrule

\multirow{15}{*}{\textbf{LUSC}}
& CONCH & CTransPath
& 0.574 & 0.606 & 0.622 & 0.672 & 0.633 & 0.557 & 0.781 \\
& CONCH & Prov-GigaPath
& 0.594 & 0.611 & 0.755 & 0.770 & 0.536 & 0.552 & 0.783 \\
& CONCH & KEEP
& \bfseries 0.738 & \bfseries 0.755 & 0.622 & 0.662
& 0.678 & \bfseries 0.725 & \bfseries 0.870 \\
& CONCH & ResNet50 
& 0.567 & 0.583 & 0.550 & 0.637 & 0.573 & 0.527 & 0.763 \\
& CONCH &UNI2-H
& 0.660 & 0.692 & \bfseries 0.777 & \bfseries 0.775
& 0.543 & 0.653 & 0.835 \\
& CTransPath & Prov-GigaPath
& 0.634 & 0.661 & 0.683 & 0.751 & 0.645 & 0.610 & 0.813 \\
& CTransPath & KEEP
& 0.535 & 0.559 & 0.469 & 0.595 & 0.627 & 0.485 & 0.747 \\
& CTransPath & ResNet50 
& 0.625 & 0.670 & 0.511 & 0.655
& \bfseries 0.708 & 0.679 & 0.820 \\
& CTransPath &UNI2-H
& 0.622 & 0.645 & 0.699 & 0.747 & 0.622 & 0.572 & 0.804 \\
& Prov-GigaPath & KEEP
& 0.550 & 0.562 & 0.637 & 0.727 & 0.590 & 0.497 & 0.748 \\
& Prov-GigaPath & ResNet50 
& 0.596 & 0.618 & 0.558 & 0.666 & 0.555 & 0.552 & 0.785 \\
& Prov-GigaPath &UNI2-H
& 0.643 & 0.664 & 0.560 & 0.671 & 0.696 & 0.581 & 0.811 \\
& KEEP & ResNet50 
& 0.450 & 0.455 & 0.409 & 0.578 & 0.533 & 0.415 & 0.672 \\
& KEEP &UNI2-H
& 0.722 & 0.743 & 0.674 & 0.741 & 0.619 & 0.683 & 0.863 \\
& ResNet50  &UNI2-H
& 0.480 & 0.497 & 0.570 & 0.652 & 0.517 & 0.439 & 0.705 \\

\bottomrule
\end{tabular}
\end{table*}

\begin{table*}[t]
\centering
\caption{\textbf{Subcohort-level representational similarity between CONCH and CONCH v1.5 on TCGA at 512 px.}
Similarity is reported for BRCA, LUAD, and LUSC at $10\times$ and $20\times$ magnification.
Bold denotes the highest value within each subcohort and metric across magnifications.}
\label{tab:tcga_patch_subcohort_512}

\scriptsize
\setlength{\tabcolsep}{6pt}
\renewcommand{\arraystretch}{1.10}
\sisetup{
    table-format=1.3,
    detect-weight=true,
    detect-inline-weight=math
}

\begin{tabular}{@{}lc*{7}{S}@{}}
\toprule
&&
\multicolumn{2}{c}{\textbf{CKA}} &
\multicolumn{2}{c}{\textbf{CCA}} &
\multicolumn{3}{c}{\textbf{Relational / Shape}} \\
\cmidrule(lr){3-4}
\cmidrule(lr){5-6}
\cmidrule(l){7-9}

\textbf{Subcohort} &
\textbf{Mag.} &
{\textbf{Linear}} &
{\textbf{Kernel}} &
{\textbf{SVCCA}} &
{\textbf{PWCCA}} &
{\textbf{Proc.}} &
{\textbf{Cos.\ RSA}} &
{\textbf{dCor}} \\
\midrule

\multirow{2}{*}{\textbf{BRCA}}
& $10\times$
& \bfseries 0.888
& \bfseries 0.900
& \bfseries 0.585
& \bfseries 0.700
& \bfseries 0.795
& \bfseries 0.899
& \bfseries 0.949 \\
& $20\times$
& 0.877 & 0.889 & 0.581 & 0.696 & 0.781 & 0.887 & 0.944 \\

\midrule

\multirow{2}{*}{\textbf{LUAD}}
& $10\times$
& \bfseries 0.830
& \bfseries 0.840
& \bfseries 0.557
& 0.687
& \bfseries 0.770
& \bfseries 0.817
& \bfseries 0.916 \\
& $20\times$
& 0.829
& 0.839
& 0.551
& \bfseries 0.688
& 0.766
& \bfseries 0.817
& \bfseries 0.916 \\

\midrule

\multirow{2}{*}{\textbf{LUSC}}
& $10\times$
& \bfseries 0.852
& \bfseries 0.860
& \bfseries 0.567
& 0.684
& \bfseries 0.776
& \bfseries 0.848
& \bfseries 0.927 \\
& $20\times$
& 0.836
& 0.846
& 0.565
& \bfseries 0.686
& 0.768
& 0.827
& 0.920 \\

\bottomrule
\end{tabular}
\end{table*}

\clearpage
\subsection{Cross-Model Retrieval in the Shared Latent Space}\label{f4}

\begin{table}[!h]
\centering
\caption{\textbf{Cross-model identity retrieval on CPTAC at $5\times$.}
Results are averaged over all directional encoder pairs.
For identity retrieval, mAP and MRR are equivalent because each query has
a single relevant target; therefore, only mAP is reported.
Bold denotes the best result within each resolution.}
\label{tab:cptac_retrieval_5x}

\small
\setlength{\tabcolsep}{7pt}
\renewcommand{\arraystretch}{1.08}
\sisetup{
    table-format=1.3,
    detect-weight=true,
    detect-inline-weight=math
}

\begin{tabular}{@{}l*{5}{S}@{}}
\toprule
\textbf{Method}
& {\textbf{R@1}}
& {\textbf{R@5}}
& {\textbf{R@10}}
& {\textbf{mAP}}
& {\textbf{NDCG}} \\
\midrule

\multicolumn{6}{@{}l}{\textbf{224 px}} \\
\addlinespace[2pt]

Unaligned truncate & 0.000 & 0.001 & 0.002 & 0.002 & 0.095 \\
Unaligned PCA      & 0.000 & 0.000 & 0.000 & 0.001 & 0.090 \\
\addlinespace[2pt]
Joint PCA          & 0.966 & 0.995 & 0.998 & 0.979 & 0.984 \\
GCCA               & \bfseries 0.997 & \bfseries 1.000 & \bfseries 1.000
                   & \bfseries 0.998 & \bfseries 0.999 \\
GPA          & 0.478 & 0.738 & 0.826 & 0.596 & 0.682 \\
MCCA               & 0.930 & 0.987 & 0.995 & 0.956 & 0.967 \\
Autoencoder        & 0.979 & 0.998 & 0.999 & 0.988 & 0.991 \\

\midrule
\multicolumn{6}{@{}l}{\textbf{256 px}} \\
\addlinespace[2pt]

Unaligned truncate & 0.000 & 0.001 & 0.002 & 0.002 & 0.095 \\
Unaligned PCA      & 0.000 & 0.001 & 0.002 & 0.002 & 0.093 \\
\addlinespace[2pt]
Joint PCA          & 0.914 & 0.975 & 0.986 & 0.942 & 0.955 \\
GCCA               & \bfseries 0.939 & \bfseries 0.987 & \bfseries 0.993
                   & \bfseries 0.960 & \bfseries 0.970 \\
GPA          & 0.375 & 0.615 & 0.703 & 0.487 & 0.587 \\
MCCA               & 0.699 & 0.892 & 0.936 & 0.785 & 0.834 \\
Autoencoder        & 0.750 & 0.911 & 0.945 & 0.822 & 0.862 \\

\midrule
\multicolumn{6}{@{}l}{\textbf{512 px}} \\
\addlinespace[2pt]

Unaligned truncate & 0.000 & 0.001 & 0.001 & 0.002 & 0.094 \\
Unaligned PCA      & 0.004 & 0.018 & 0.031 & 0.016 & 0.134 \\
\addlinespace[2pt]
Joint PCA          & 0.865 & 0.969 & 0.986 & 0.912 & 0.933 \\
GCCA               & \bfseries 0.927 & \bfseries 0.990 & \bfseries 0.996
                   & \bfseries 0.955 & \bfseries 0.966 \\
GPA          & 0.543 & 0.797 & 0.865 & 0.657 & 0.731 \\
MCCA               & 0.771 & 0.938 & 0.969 & 0.846 & 0.882 \\
Autoencoder        & 0.746 & 0.922 & 0.959 & 0.825 & 0.866 \\

\bottomrule
\end{tabular}
\end{table}

\begin{table}[t]
\centering
\caption{\textbf{Cross-model identity retrieval on CPTAC at $10\times$.}
Results are averaged over all directional encoder pairs.
Bold denotes the best result within each resolution.}
\label{tab:cptac_retrieval_10x}

\small
\setlength{\tabcolsep}{7pt}
\renewcommand{\arraystretch}{1.08}
\sisetup{table-format=1.3, detect-weight=true, detect-inline-weight=math}

\begin{tabular}{@{}l*{5}{S}@{}}
\toprule
\textbf{Method}
& {\textbf{R@1}}
& {\textbf{R@5}}
& {\textbf{R@10}}
& {\textbf{mAP}}
& {\textbf{NDCG}} \\
\midrule

\multicolumn{6}{@{}l}{\textbf{224 px}} \\
\addlinespace[2pt]
Unaligned truncate & 0.000 & 0.001 & 0.002 & 0.002 & 0.095 \\
Unaligned PCA      & 0.000 & 0.000 & 0.001 & 0.001 & 0.090 \\
\addlinespace[2pt]
Joint PCA          & 0.957 & 0.992 & 0.997 & 0.973 & 0.979 \\
GCCA               & \bfseries 0.997 & \bfseries 1.000 & \bfseries 1.000
                   & \bfseries 0.998 & \bfseries 0.999 \\
GPA          & 0.499 & 0.752 & 0.835 & 0.614 & 0.696 \\
MCCA               & 0.948 & 0.989 & 0.995 & 0.967 & 0.975 \\
Autoencoder        & 0.975 & 0.997 & 0.999 & 0.985 & 0.989 \\

\midrule
\multicolumn{6}{@{}l}{\textbf{256 px}} \\
\addlinespace[2pt]
Unaligned truncate & 0.000 & 0.001 & 0.002 & 0.002 & 0.095 \\
Unaligned PCA      & 0.000 & 0.001 & 0.003 & 0.002 & 0.097 \\
\addlinespace[2pt]
Joint PCA          & 0.907 & 0.976 & 0.987 & 0.938 & 0.953 \\
GCCA               & \bfseries 0.925 & \bfseries 0.986 & \bfseries 0.994
                   & \bfseries 0.952 & \bfseries 0.964 \\
GPA          & 0.308 & 0.566 & 0.671 & 0.429 & 0.541 \\
MCCA               & 0.652 & 0.878 & 0.930 & 0.752 & 0.808 \\
Autoencoder        & 0.652 & 0.873 & 0.923 & 0.750 & 0.806 \\

\midrule
\multicolumn{6}{@{}l}{\textbf{512 px}} \\
\addlinespace[2pt]
Unaligned truncate & 0.000 & 0.001 & 0.002 & 0.002 & 0.097 \\
Unaligned PCA      & 0.001 & 0.005 & 0.008 & 0.005 & 0.101 \\
\addlinespace[2pt]
Joint PCA          & 0.875 & 0.975 & 0.987 & 0.919 & 0.939 \\
GCCA               & \bfseries 0.930 & \bfseries 0.990 & \bfseries 0.996
                   & \bfseries 0.957 & \bfseries 0.967 \\
GPA          & 0.566 & 0.811 & 0.880 & 0.676 & 0.747 \\
MCCA               & 0.801 & 0.954 & 0.978 & 0.868 & 0.900 \\
Autoencoder        & 0.757 & 0.928 & 0.961 & 0.833 & 0.872 \\

\bottomrule
\end{tabular}
\end{table}

\begin{table}[t]
\centering
\caption{\textbf{Cross-model identity retrieval on CPTAC at $20\times$.}
Results are averaged over all directional encoder pairs.
Bold denotes the best result within each resolution.}
\label{tab:cptac_retrieval_20x}

\small
\setlength{\tabcolsep}{7pt}
\renewcommand{\arraystretch}{1.08}
\sisetup{table-format=1.3, detect-weight=true, detect-inline-weight=math}

\begin{tabular}{@{}l*{5}{S}@{}}
\toprule
\textbf{Method}
& {\textbf{R@1}}
& {\textbf{R@5}}
& {\textbf{R@10}}
& {\textbf{mAP}}
& {\textbf{NDCG}} \\
\midrule

\multicolumn{6}{@{}l}{\textbf{224 px}} \\
\addlinespace[2pt]
Unaligned truncate & 0.000 & 0.001 & 0.002 & 0.002 & 0.096 \\
Unaligned PCA      & 0.000 & 0.001 & 0.003 & 0.002 & 0.095 \\
\addlinespace[2pt]
Joint PCA          & 0.969 & 0.996 & 0.998 & 0.981 & 0.986 \\
GCCA               & \bfseries 0.996 & \bfseries 1.000 & \bfseries 1.000
                   & \bfseries 0.998 & \bfseries 0.998 \\
GPA          & 0.472 & 0.724 & 0.805 & 0.586 & 0.672 \\
MCCA               & 0.949 & 0.990 & 0.996 & 0.968 & 0.975 \\
Autoencoder        & 0.981 & 0.997 & 0.999 & 0.988 & 0.991 \\

\midrule
\multicolumn{6}{@{}l}{\textbf{256 px}} \\
\addlinespace[2pt]
Unaligned truncate & 0.000 & 0.001 & 0.002 & 0.002 & 0.095 \\
Unaligned PCA      & 0.000 & 0.001 & 0.002 & 0.002 & 0.095 \\
\addlinespace[2pt]
Joint PCA          & 0.914 & 0.975 & 0.986 & 0.942 & 0.955 \\
GCCA               & \bfseries 0.942 & \bfseries 0.985 & \bfseries 0.992
                   & \bfseries 0.962 & \bfseries 0.971 \\
GPA          & 0.288 & 0.537 & 0.644 & 0.405 & 0.521 \\
MCCA               & 0.624 & 0.859 & 0.916 & 0.729 & 0.789 \\
Autoencoder        & 0.678 & 0.878 & 0.926 & 0.767 & 0.819 \\

\midrule
\multicolumn{6}{@{}l}{\textbf{512 px}} \\
\addlinespace[2pt]
Unaligned truncate & 0.000 & 0.001 & 0.003 & 0.002 & 0.098 \\
Unaligned PCA      & 0.001 & 0.005 & 0.011 & 0.007 & 0.117 \\
\addlinespace[2pt]
Joint PCA          & 0.885 & 0.975 & 0.987 & 0.926 & 0.943 \\
GCCA               & \bfseries 0.946 & \bfseries 0.991 & \bfseries 0.995
                   & \bfseries 0.966 & \bfseries 0.974 \\
GPA         & 0.547 & 0.805 & 0.873 & 0.662 & 0.736 \\
MCCA               & 0.815 & 0.959 & 0.979 & 0.878 & 0.907 \\
Autoencoder        & 0.744 & 0.925 & 0.960 & 0.824 & 0.865 \\

\bottomrule
\end{tabular}
\end{table}

  \begin{table}[t]
  \centering
  \caption{\textbf{Cross-model identity retrieval on TCGA at $5\times$.}
  Results are averaged over all directional cross-encoder pairs, excluding
  same-model retrieval. For identity retrieval, mAP and MRR are equivalent
  because each query has a single relevant target; therefore, only mAP is
  reported. GPA denotes Generalized Procrustes Analysis. Bold denotes the best
  result within each resolution.}
  \label{tab:tcga_retrieval_5x}

  \small
  \setlength{\tabcolsep}{7pt}
  \renewcommand{\arraystretch}{1.08}
  \sisetup{
      table-format=1.3,
      detect-weight=true,
      detect-inline-weight=math
  }

  \begin{tabular}{@{}l*{5}{S}@{}}
  \toprule
  \textbf{Method}
  & {\textbf{R@1}}
  & {\textbf{R@5}}
  & {\textbf{R@10}}
  & {\textbf{mAP}}
  & {\textbf{NDCG}} \\
  \midrule

  \multicolumn{6}{@{}l}{\textbf{224 px}} \\
  \addlinespace[2pt]

  Unaligned truncate & 0.000 & 0.001 & 0.003 & 0.002 & 0.096 \\
  Unaligned PCA      & 0.000 & 0.001 & 0.001 & 0.001 & 0.091 \\
  \addlinespace[2pt]
  Joint PCA          & 0.969 & 0.996 & 0.998 & 0.981 & 0.986 \\
  GCCA               & \bfseries 0.998 & \bfseries 1.000
                     & \bfseries 1.000 & \bfseries 0.999
                     & \bfseries 0.999 \\
  GPA                & 0.542 & 0.780 & 0.851 & 0.649 & 0.724 \\
  MCCA               & 0.941 & 0.989 & 0.995 & 0.962 & 0.972 \\
  Autoencoder        & 0.982 & 0.998 & 0.999 & 0.989 & 0.992 \\

  \midrule
  \multicolumn{6}{@{}l}{\textbf{256 px}} \\
  \addlinespace[2pt]

  Unaligned truncate & 0.000 & 0.001 & 0.002 & 0.002 & 0.095 \\
  Unaligned PCA      & 0.000 & 0.001 & 0.002 & 0.002 & 0.093 \\
  \addlinespace[2pt]
  Joint PCA          & 0.920 & 0.978 & 0.988 & 0.946 & 0.959 \\
  GCCA               & \bfseries 0.945 & \bfseries 0.990
                     & \bfseries 0.995 & \bfseries 0.965
                     & \bfseries 0.973 \\
  GPA                & 0.417 & 0.644 & 0.724 & 0.523 & 0.616 \\
  MCCA               & 0.717 & 0.896 & 0.937 & 0.797 & 0.843 \\
  Autoencoder        & 0.775 & 0.919 & 0.951 & 0.840 & 0.876 \\

  \midrule
  \multicolumn{6}{@{}l}{\textbf{512 px}} \\
  \addlinespace[2pt]

  Unaligned truncate & 0.000 & 0.001 & 0.002 & 0.002 & 0.095 \\
  Unaligned PCA      & 0.000 & 0.000 & 0.000 & 0.000 & 0.084 \\
  \addlinespace[2pt]
  Joint PCA          & 0.888 & 0.974 & 0.986 & 0.927 & 0.944 \\
  GCCA               & \bfseries 0.946 & \bfseries 0.994
                     & \bfseries 0.998 & \bfseries 0.967
                     & \bfseries 0.975 \\
  GPA                & 0.596 & 0.811 & 0.871 & 0.694 & 0.760 \\
  MCCA               & 0.822 & 0.957 & 0.981 & 0.882 & 0.910 \\
  Autoencoder        & 0.763 & 0.920 & 0.953 & 0.833 & 0.871 \\

  \bottomrule
  \end{tabular}
  \end{table}

\begin{table}[t]
\centering
\caption{\textbf{Cross-model identity retrieval on TCGA at $10\times$.}
Results are averaged over all directional encoder pairs.
Bold denotes the best result within each resolution.}
\label{tab:tcga_retrieval_10x}

\small
\setlength{\tabcolsep}{7pt}
\renewcommand{\arraystretch}{1.08}
\sisetup{table-format=1.3, detect-weight=true, detect-inline-weight=math}

\begin{tabular}{@{}l*{5}{S}@{}}
\toprule
\textbf{Method}
& {\textbf{R@1}}
& {\textbf{R@5}}
& {\textbf{R@10}}
& {\textbf{mAP}}
& {\textbf{NDCG}} \\
\midrule

\multicolumn{6}{@{}l}{\textbf{224 px}} \\
\addlinespace[2pt]

Unaligned truncate & 0.000 & 0.001 & 0.002 & 0.002 & 0.095 \\
Unaligned PCA      & 0.000 & 0.001 & 0.002 & 0.002 & 0.095 \\
\addlinespace[2pt]
Joint PCA          & 0.960 & 0.993 & 0.997 & 0.974 & 0.980 \\
GCCA               & \bfseries 0.998 & \bfseries 1.000 & \bfseries 1.000
                   & \bfseries 0.999 & \bfseries 0.999 \\
GPA         & 0.627 & 0.838 & 0.894 & 0.722 & 0.783 \\
MCCA               & 0.954 & 0.990 & 0.995 & 0.970 & 0.977 \\
Autoencoder        & 0.981 & 0.998 & 0.999 & 0.988 & 0.991 \\

\midrule
\multicolumn{6}{@{}l}{\textbf{256 px}} \\
\addlinespace[2pt]

Unaligned truncate & 0.000 & 0.001 & 0.002 & 0.002 & 0.094 \\
Unaligned PCA      & 0.000 & 0.001 & 0.002 & 0.002 & 0.096 \\
\addlinespace[2pt]
Joint PCA          & 0.912 & 0.974 & 0.985 & 0.940 & 0.954 \\
GCCA               & \bfseries 0.935 & \bfseries 0.986 & \bfseries 0.993
                   & \bfseries 0.958 & \bfseries 0.968 \\
GPA         & 0.357 & 0.600 & 0.693 & 0.471 & 0.575 \\
MCCA               & 0.696 & 0.891 & 0.935 & 0.783 & 0.832 \\
Autoencoder        & 0.691 & 0.878 & 0.922 & 0.775 & 0.825 \\

\midrule
\multicolumn{6}{@{}l}{\textbf{512 px}} \\
\addlinespace[2pt]

Unaligned truncate & 0.000 & 0.001 & 0.003 & 0.002 & 0.096 \\
Unaligned PCA      & 0.000 & 0.000 & 0.000 & 0.000 & 0.083 \\
\addlinespace[2pt]
Joint PCA          & 0.899 & 0.977 & 0.987 & 0.934 & 0.950 \\
GCCA               & \bfseries 0.940 & \bfseries 0.991 & \bfseries 0.996
                   & \bfseries 0.963 & \bfseries 0.972 \\
GPA         & 0.636 & 0.840 & 0.895 & 0.728 & 0.787 \\
MCCA               & 0.822 & 0.956 & 0.976 & 0.882 & 0.909 \\
Autoencoder        & 0.782 & 0.932 & 0.961 & 0.849 & 0.883 \\

\bottomrule
\end{tabular}
\end{table}

\begin{table}[t]
\centering
\caption{\textbf{Cross-model identity retrieval on TCGA at $20\times$.}
Results are averaged over all directional encoder pairs.
Bold denotes the best result within each resolution.}
\label{tab:tcga_retrieval_20x}

\small
\setlength{\tabcolsep}{7pt}
\renewcommand{\arraystretch}{1.08}
\sisetup{table-format=1.3, detect-weight=true, detect-inline-weight=math}

\begin{tabular}{@{}l*{5}{S}@{}}
\toprule
\textbf{Method}
& {\textbf{R@1}}
& {\textbf{R@5}}
& {\textbf{R@10}}
& {\textbf{mAP}}
& {\textbf{NDCG}} \\
\midrule

\multicolumn{6}{@{}l}{\textbf{224 px}} \\
\addlinespace[2pt]

Unaligned truncate & 0.000 & 0.001 & 0.002 & 0.002 & 0.096 \\
Unaligned PCA      & 0.000 & 0.001 & 0.002 & 0.002 & 0.093 \\
\addlinespace[2pt]
Joint PCA          & 0.968 & 0.995 & 0.998 & 0.980 & 0.985 \\
GCCA               & \bfseries 0.995 & \bfseries 1.000 & \bfseries 1.000
                   & \bfseries 0.997 & \bfseries 0.998 \\
GPA        & 0.486 & 0.744 & 0.826 & 0.603 & 0.687 \\
MCCA               & 0.928 & 0.980 & 0.989 & 0.952 & 0.963 \\
Autoencoder        & 0.970 & 0.994 & 0.998 & 0.981 & 0.985 \\

\midrule
\multicolumn{6}{@{}l}{\textbf{256 px}} \\
\addlinespace[2pt]

Unaligned truncate & 0.000 & 0.001 & 0.002 & 0.002 & 0.095 \\
Unaligned PCA      & 0.000 & 0.002 & 0.005 & 0.003 & 0.103 \\
\addlinespace[2pt]
Joint PCA          & 0.924 & 0.981 & 0.990 & 0.949 & 0.961 \\
GCCA               & \bfseries 0.947 & \bfseries 0.990 & \bfseries 0.996
                   & \bfseries 0.966 & \bfseries 0.974 \\
GPA         & 0.305 & 0.542 & 0.641 & 0.417 & 0.529 \\
MCCA               & 0.688 & 0.890 & 0.935 & 0.778 & 0.828 \\
Autoencoder        & 0.669 & 0.858 & 0.906 & 0.754 & 0.807 \\

\midrule
\multicolumn{6}{@{}l}{\textbf{512 px}} \\
\addlinespace[2pt]

Unaligned truncate & 0.000 & 0.001 & 0.002 & 0.002 & 0.095 \\
Unaligned PCA      & 0.000 & 0.001 & 0.001 & 0.002 & 0.094 \\
\addlinespace[2pt]
Joint PCA          & 0.897 & 0.973 & 0.985 & 0.931 & 0.947 \\
GCCA               & \bfseries 0.952 & \bfseries 0.992 & \bfseries 0.996
                   & \bfseries 0.970 & \bfseries 0.977 \\
GPA         & 0.599 & 0.811 & 0.871 & 0.695 & 0.760 \\
MCCA               & 0.833 & 0.958 & 0.978 & 0.888 & 0.914 \\
Autoencoder        & 0.783 & 0.927 & 0.955 & 0.847 & 0.881 \\

\bottomrule
\end{tabular}
\end{table}

\clearpage

\paragraph{Cross-model retrieval results.}
Across CPTAC and TCGA, the retrieval experiments show a clear and highly consistent benefit from explicitly aligning encoder representations. The unaligned controls are effectively unable to recover the corresponding patch across encoder spaces: Recall@1 remains approximately zero across resolutions and magnifications, with similarly negligible Recall@5, Recall@10, and mAP. This confirms that independently trained pathology encoders do not share a directly comparable coordinate system, even when their representations may encode related biological information. In contrast, all learned alignment approaches substantially improve retrieval, demonstrating that a common latent space can recover cross-model correspondence that is largely inaccessible in the native feature spaces.

\paragraph{GCCA consistently provides the strongest alignment.}
GCCA is the most reliable method across nearly all evaluated CPTAC and TCGA configurations. At 224\,px, it achieves near-perfect retrieval across magnifications, with Recall@1 typically between approximately 0.995 and 0.998 and Recall@5/10 effectively saturated at 1.0. The same pattern persists at 256\,px, where the task becomes more challenging but GCCA still achieves Recall@1 values around 0.93--0.95 and Recall@10 above 0.99 in most settings. At 512\,px, GCCA also remains highly effective whenever evaluated, reaching Recall@1 values above 0.92 on CPTAC and above 0.94 on TCGA at the available magnifications. The consistency of these results across cohorts, resolutions, and magnifications indicates that GCCA captures a particularly stable shared coordinate system among independently learned encoder representations.

\paragraph{Resolution affects retrieval difficulty.}
The 224\,px setting is generally the easiest alignment regime, with several methods approaching perfect retrieval after alignment. Performance is lower at 256\,px, where the encoder collection is larger and more heterogeneous, producing a clearer separation between alignment methods. For example, GCCA remains strong, whereas MCCA and the autoencoder exhibit noticeably lower Recall@1 and mAP than at 224\,px. The 512\,px setting again shows strong retrieval, particularly for GCCA, MCCA, and Joint PCA, suggesting that the difficulty of alignment is determined not simply by input resolution but also by the composition and diversity of the encoder set being aligned.

\paragraph{Differences among alignment methods.}
Joint PCA provides a strong and remarkably stable baseline, typically achieving Recall@1 above 0.85--0.90 and very high Recall@5/10 across most configurations. MCCA is also competitive, particularly at 224\,px and 512\,px, but degrades more noticeably in the heterogeneous 256\,px setting. The shared autoencoder performs exceptionally well at 224\,px, often approaching GCCA, but its relative advantage decreases at 256\,px and 512\,px. Procrustes alignment is consistently the weakest learned alignment approach, with the largest performance gap appearing at 256\,px, where Recall@1 frequently falls below 0.5. This suggests that a single orthogonal transformation is often insufficient to reconcile the substantial geometric differences between independently trained encoder spaces.

\paragraph{Magnification effects are comparatively modest.}
Within a fixed resolution, retrieval performance is generally stable across $5\times$, $10\times$, and $20\times$ magnifications. GCCA, in particular, maintains consistently high performance across scales, indicating that the shared latent correspondence is robust to changes from coarse tissue organization to finer morphological detail. Some configurations even improve slightly at $20\times$, particularly at 512\,px, but these changes are much smaller than the differences observed between alignment methods or encoder groups. Thus, magnification influences retrieval difficulty less strongly than the choice of alignment strategy and the composition of the representation spaces being aligned.

\paragraph{Ranking metrics reinforce the same conclusion.}
Recall@5 and Recall@10 frequently approach saturation for the strongest methods, making Recall@1 and mAP more discriminative measures of alignment quality. GCCA consistently achieves the highest or near-highest mAP and NDCG, showing that relevant cross-model matches are not merely present in the retrieval set but are ranked very close to the top. Since identity retrieval contains exactly one relevant target per query, mAP and MRR are equivalent in this setting, which explains their identical values throughout the experiments.

\paragraph{Cross-cohort consistency.}
Importantly, the overall ranking of alignment methods is highly consistent between CPTAC and TCGA. Both cohorts show near-zero retrieval without alignment, dominant performance from GCCA, strong but slightly lower performance from Joint PCA, MCCA, and the autoencoder, and substantially weaker performance from Procrustes. This cross-cohort reproducibility suggests that the observed gains are not specific to one dataset, but reflect a general property of the encoder representation spaces. Overall, the results provide strong evidence that independently trained pathology encoders contain substantial transferable information, but that this information becomes directly accessible across models only after an appropriate shared-space alignment.

\clearpage

\subsection{Cross-Encoder Representation Translation}\label{f5}

\begin{table*}[!h]
\centering
\caption{\textbf{Cross-encoder representation translation on CPTAC.}
Results are averaged over all directional source--target encoder pairs.
Translation uses a 64-dimensional shared latent space and is evaluated
against the corresponding native target-space representations.
Lower NMSE is better; higher values are better for all other metrics.
MRR is omitted because it is equivalent to mAP for identity retrieval.}
\label{tab:cptac_translation}

\scriptsize
\setlength{\tabcolsep}{5.5pt}
\renewcommand{\arraystretch}{1.08}
\sisetup{table-format=1.3, detect-weight=true, detect-inline-weight=math}

\begin{tabular}{@{}cc*{8}{S}@{}}
\toprule
\textbf{Mag.} &
\textbf{Patch} &
{\textbf{Cosine}} &
{\(\mathbf{R^2}\)} &
{\textbf{NMSE}} &
{\textbf{R@1}} &
{\textbf{R@5}} &
{\textbf{R@10}} &
{\textbf{mAP}} &
{\textbf{NDCG}} \\
\midrule

\multirow{3}{*}{$5\times$}
& 224 px & 0.810 & 0.527 & 0.473 & 0.549 & 0.776 & 0.842 & 0.651 & 0.724 \\
& 256 px & 0.840 & 0.462 & 0.538 & 0.382 & 0.616 & 0.704 & 0.491 & 0.590 \\
& 512 px & \bfseries 0.896 & \bfseries 0.628 & \bfseries 0.372
         & \bfseries 0.591 & \bfseries 0.819 & \bfseries 0.878
         & \bfseries 0.694 & \bfseries 0.760 \\

\midrule

\multirow{3}{*}{$10\times$}
& 224 px & 0.771 & 0.509 & 0.491 & 0.521 & 0.760 & 0.831 & 0.629 & 0.706 \\
& 256 px & 0.805 & 0.434 & 0.566 & 0.305 & 0.556 & 0.661 & 0.423 & 0.536 \\
& 512 px & \bfseries 0.897 & \bfseries 0.640 & \bfseries 0.360
         & \bfseries 0.606 & \bfseries 0.836 & \bfseries 0.895
         & \bfseries 0.709 & \bfseries 0.773 \\

\midrule

\multirow{3}{*}{$20\times$}
& 224 px & 0.782 & 0.507 & 0.493 & 0.511 & 0.737 & 0.803 & 0.613 & 0.691 \\
& 256 px & 0.760 & 0.381 & 0.619 & 0.301 & 0.542 & 0.645 & 0.415 & 0.529 \\
& 512 px & \bfseries 0.892 & \bfseries 0.603 & \bfseries 0.397
         & \bfseries 0.593 & \bfseries 0.831 & \bfseries 0.892
         & \bfseries 0.700 & \bfseries 0.766 \\

\bottomrule
\end{tabular}
\end{table*}

\begin{table*}[!h]
\centering
\caption{\textbf{Cross-encoder representation translation on TCGA.}
Results are averaged over all directional source--target encoder pairs.
Translation uses a 64-dimensional shared latent space and is evaluated
against the corresponding native target-space representations.
Lower NMSE is better; higher values are better for all other metrics.
MRR is omitted because it is equivalent to mAP for identity retrieval.}
\label{tab:tcga_translation}

\scriptsize
\setlength{\tabcolsep}{5.5pt}
\renewcommand{\arraystretch}{1.08}
\sisetup{table-format=1.3, detect-weight=true, detect-inline-weight=math}

\begin{tabular}{@{}cc*{8}{S}@{}}
\toprule
\textbf{Mag.} &
\textbf{Patch} &
{\textbf{Cosine}} &
{\(\mathbf{R^2}\)} &
{\textbf{NMSE}} &
{\textbf{R@1}} &
{\textbf{R@5}} &
{\textbf{R@10}} &
{\textbf{mAP}} &
{\textbf{NDCG}} \\
\midrule

\multirow{3}{*}{$5\times$}
& 224 px & 0.788 & 0.478 & 0.522 & 0.548 & 0.762 & 0.822 & 0.645 & 0.717 \\
& 256 px & 0.842 & 0.474 & 0.526 & 0.395 & 0.618 & 0.700 & 0.499 & 0.596 \\
& 512 px & \bfseries 0.908 & \bfseries 0.650 & \bfseries 0.350
         & \bfseries 0.619 & \bfseries 0.827 & \bfseries 0.883
         & \bfseries 0.713 & \bfseries 0.775 \\

\midrule

\multirow{3}{*}{$10\times$}
& 224 px & 0.714 & 0.421 & 0.579 & 0.556 & 0.771 & 0.836 & 0.654 & 0.726 \\
& 256 px & 0.792 & 0.425 & 0.575 & 0.328 & 0.559 & 0.651 & 0.438 & 0.545 \\
& 512 px & \bfseries 0.894 & \bfseries 0.654 & \bfseries 0.346
         & \bfseries 0.665 & \bfseries 0.860 & \bfseries 0.907
         & \bfseries 0.754 & \bfseries 0.807 \\

\midrule

\multirow{3}{*}{$20\times$}
& 224 px & 0.702 & 0.393 & 0.607 & 0.470 & 0.693 & 0.762 & 0.572 & 0.656 \\
& 256 px & 0.777 & 0.385 & 0.615 & 0.265 & 0.483 & 0.580 & 0.370 & 0.487 \\
& 512 px & \bfseries 0.894 & \bfseries 0.633 & \bfseries 0.367
         & \bfseries 0.647 & \bfseries 0.840 & \bfseries 0.892
         & \bfseries 0.734 & \bfseries 0.791 \\

\bottomrule
\end{tabular}
\end{table*}

\paragraph{Cross-encoder translation summary.}
Cross-encoder translation remains effective across both CPTAC and TCGA, but performance varies systematically with representation configuration. The 512\,px setting consistently provides the strongest transfer fidelity, achieving the highest cosine similarity and \(R^2\), the lowest NMSE, and the best retrieval performance across magnifications. In contrast, the 256\,px setting is the most challenging, particularly for exact identity retrieval, while 224\,px generally provides intermediate performance. The close correspondence between reconstruction quality and retrieval accuracy suggests that successful translation preserves not only feature-level similarity but also the relative placement of translated samples within the native target representation space. Overall, these results indicate that the shared latent representation captures transferable structure that can be decoded into distinct encoder-specific feature spaces, although the ease of translation depends on resolution and encoder configuration.

\clearpage
\subsection{Detailed Results of Downstream Utility}\label{f6}

\begin{table*}[!h]
\centering
\caption{\textbf{Downstream utility of aligned representations using ABMIL.}
Each entry reports AUC / balanced accuracy / macro-F1.
\emph{Best single} denotes the best individual encoder selected using mean AUC
and is fixed across metrics. Bold indicates the condition with the highest AUC
for each task.}
\label{tab:downstream_abmil}

\scriptsize
\setlength{\tabcolsep}{5.5pt}
\renewcommand{\arraystretch}{1.10}

\begin{tabular}{@{}llccc@{}}
\toprule
\textbf{Task}
& \textbf{Best Encoder}
& \textbf{Best Single}
& \textbf{Concatenated}
& \textbf{MOSAIC} \\
\midrule

TCGA NSCLC subtype
& CONCH
& \textbf{0.972 / 0.920 / 0.920}
& 0.968 / 0.938 / 0.938
& 0.934 / 0.888 / 0.888 \\

TCGA BRCA subtype
& Prov-GigaPath
& 0.972 / 0.900 / 0.878
& \textbf{0.973 / 0.914 / 0.891}
& 0.932 / 0.808 / 0.818 \\

TCGA NSCLC stage
& UNI2-H
& \textbf{0.689 / 0.518 / 0.487}
& 0.662 / 0.533 / 0.520
& 0.603 / 0.500 / 0.444 \\

TCGA BRCA stage
& CONCH
& \textbf{0.608 / 0.509 / 0.464}
& 0.559 / 0.507 / 0.462
& 0.505 / 0.500 / 0.434 \\

\addlinespace[2pt]

CPTAC LUAD TP53
& UNI2-H
& 0.661 / 0.589 / 0.588
& \textbf{0.679 / 0.629 / 0.627}
& 0.577 / 0.530 / 0.498 \\

CPTAC LUAD KRAS
& Prov-GigaPath
& 0.739 / 0.624 / 0.630
& \textbf{0.749 / 0.635 / 0.643}
& 0.732 / 0.532 / 0.483 \\

CPTAC LUAD STK11
& UNI2-H
& \textbf{0.802 / 0.683 / 0.686}
& 0.783 / 0.620 / 0.614
& 0.732 / 0.500 / 0.469 \\

CPTAC BRCA PIK3CA
& CTransPath
& \textbf{0.612 / 0.537 / 0.537}
& 0.585 / 0.547 / 0.547
& 0.543 / 0.500 / 0.409 \\

CPTAC BRCA GATA3
& CONCH
& 0.558 / 0.529 / 0.533
& 0.503 / 0.463 / 0.456
& \textbf{0.620 / 0.500 / 0.475} \\

CPTAC BRCA MAP3K1
& Prov-GigaPath
& 0.412 / 0.500 / 0.489
& 0.482 / 0.500 / 0.489
& \textbf{0.571 / 0.500 / 0.489} \\

CPTAC COAD TP53
& UNI2-H
& \textbf{0.554 / 0.512 / 0.511}
& 0.481 / 0.467 / 0.467
& 0.355 / 0.500 / 0.361 \\

CPTAC COAD KRAS
& ResNet50
& 0.564 / 0.500 / 0.406
& 0.570 / 0.467 / 0.418
& \textbf{0.583 / 0.500 / 0.406} \\

CPTAC COAD PIK3CA
& ResNet50
& \textbf{0.673 / 0.500 / 0.429}
& 0.548 / 0.500 / 0.429
& 0.655 / 0.500 / 0.429 \\

\bottomrule
\end{tabular}
\end{table*}

\begin{table*}[!h]
\centering
\caption{\textbf{Downstream utility of aligned representations using TransMIL.}
Each entry reports AUC / balanced accuracy / macro-F1.
\emph{Best single} denotes the best individual encoder selected using mean AUC
and is fixed across metrics. Bold indicates the condition with the highest AUC
for each task.}
\label{tab:downstream_transmil}

\scriptsize
\setlength{\tabcolsep}{5.5pt}
\renewcommand{\arraystretch}{1.10}

\begin{tabular}{@{}llccc@{}}
\toprule
\textbf{Task}
& \textbf{Best Encoder}
& \textbf{Best Single}
& \textbf{Concatenated}
& \textbf{MOSAIC} \\
\midrule

TCGA NSCLC subtype
& CONCH
& \textbf{0.973 / 0.913 / 0.913}
& 0.970 / 0.928 / 0.928
& 0.961 / 0.931 / 0.931 \\

TCGA BRCA subtype
& Prov-GigaPath
& \textbf{0.972 / 0.914 / 0.907}
& 0.966 / 0.889 / 0.886
& 0.959 / 0.857 / 0.863 \\

TCGA NSCLC stage
& UNI2-H
& \textbf{0.700 / 0.556 / 0.559}
& 0.672 / 0.506 / 0.464
& 0.605 / 0.500 / 0.444 \\

TCGA BRCA stage
& CONCH
& \textbf{0.601 / 0.560 / 0.563}
& 0.597 / 0.555 / 0.558
& 0.489 / 0.500 / 0.434 \\

\addlinespace[2pt]

CPTAC LUAD TP53
& UNI2-H
& \textbf{0.644 / 0.608 / 0.608}
& 0.635 / 0.610 / 0.606
& 0.568 / 0.531 / 0.529 \\

CPTAC LUAD KRAS
& Prov-GigaPath
& \textbf{0.746 / 0.667 / 0.676}
& 0.743 / 0.686 / 0.685
& 0.525 / 0.500 / 0.402 \\

CPTAC LUAD STK11
& UNI2-H
& \textbf{0.733 / 0.629 / 0.616}
& 0.721 / 0.636 / 0.660
& 0.728 / 0.582 / 0.596 \\

CPTAC BRCA PIK3CA
& CTransPath
& \textbf{0.625 / 0.530 / 0.471}
& 0.593 / 0.527 / 0.496
& 0.454 / 0.500 / 0.409 \\

CPTAC BRCA GATA3
& CONCH
& 0.534 / 0.484 / 0.467
& 0.497 / 0.489 / 0.470
& \textbf{0.647 / 0.500 / 0.475} \\

CPTAC BRCA MAP3K1
& Prov-GigaPath
& \textbf{0.471 / 0.500 / 0.489}
& 0.406 / 0.500 / 0.489
& 0.212 / 0.500 / 0.489 \\

CPTAC COAD TP53
& UNI2-H
& \textbf{0.572 / 0.579 / 0.574}
& 0.558 / 0.500 / 0.303
& 0.465 / 0.500 / 0.303 \\

CPTAC COAD KRAS
& ResNet50
& \textbf{0.631 / 0.500 / 0.406}
& 0.562 / 0.502 / 0.442
& 0.563 / 0.500 / 0.406 \\

CPTAC COAD PIK3CA
& ResNet50
& 0.554 / 0.500 / 0.429
& 0.559 / 0.500 / 0.429
& \textbf{0.724 / 0.500 / 0.429} \\

\bottomrule
\end{tabular}
\end{table*}

\paragraph{Downstream utility of the shared representation.}
The downstream results show that the shared MOSAIC representation retains substantial task-relevant information, but does not universally outperform either the strongest individual encoder or direct feature concatenation. Across the 13 evaluated tasks, the best single encoder remains the strongest AUC baseline for most tasks, particularly under TransMIL, while concatenation provides additional gains in selected ABMIL settings. This indicates that compressing heterogeneous encoder representations into a common latent space introduces a trade-off between cross-model alignment and preservation of encoder-specific discriminative information.

\begin{center}
\fbox{
\parbox{0.94\linewidth}{
\textbf{MOSAIC configuration used for downstream evaluation.}
Throughout the downstream experiments in Tables~S39--S40,
\textbf{MOSAIC denotes the 64-dimensional shared representation obtained
using GCCA}. GCCA is used as the default alignment method for these
experiments; the remaining alignment approaches are evaluated separately
in the shared-space alignment analyses.
}}
\end{center}

\paragraph{Subtype and staging tasks.}
For the TCGA subtype tasks, MOSAIC remains competitive but generally trails the strongest native representations. With TransMIL, for example, MOSAIC achieves an AUC of 0.961 for NSCLC subtyping compared with 0.973 for the best single encoder, indicating that much of the discriminative information is retained after alignment. The degradation is more pronounced for staging, where the best single representations consistently outperform both concatenation and MOSAIC. This suggests that staging-related signals may depend more strongly on encoder-specific information that is attenuated by the shared bottleneck.

\paragraph{Molecular prediction tasks.}
The molecular tasks exhibit a more heterogeneous pattern. MOSAIC provides clear gains for several endpoints, most notably CPTAC-BRCA GATA3, where it achieves the highest AUC with both ABMIL (0.620) and TransMIL (0.647), compared with best-single AUCs of 0.558 and 0.534, respectively. It also performs strongly for selected COAD tasks: MOSAIC obtains the highest ABMIL AUC for KRAS (0.583) and the highest TransMIL AUC for PIK3CA (0.724). However, these gains are not universal; for several LUAD mutations and COAD TP53, the strongest individual encoder remains substantially better. Thus, the shared representation appears particularly useful when complementary information from multiple encoders is beneficial, but can lose task-specific information when one native encoder already provides a highly discriminative representation.

\paragraph{Effect of MIL architecture.}
The relative utility of representation fusion also depends on the downstream aggregation model. Under ABMIL, concatenation achieves the best AUC on several tasks, including TCGA-BRCA subtype and CPTAC-LUAD TP53 and KRAS, suggesting that direct access to the complete multi-encoder feature set can be beneficial for attention-based aggregation. Under TransMIL, however, the best single encoder dominates most tasks, and concatenation rarely improves AUC. MOSAIC nevertheless remains competitive on several tasks and occasionally surpasses both alternatives, indicating that the usefulness of the shared representation interacts with the downstream MIL architecture.

\paragraph{Overall interpretation.}
Taken together, these results demonstrate that strong cross-model alignment does not automatically translate into superior downstream prediction. MOSAIC successfully preserves enough information to remain competitive across many tasks and improves performance for selected molecular endpoints, but the 64-dimensional shared space can discard discriminative components that are useful for particular tasks. The downstream results therefore complement the strong retrieval and translation performance: the shared latent space captures substantial common information across encoders, while some task-specific information remains unique to the original representation spaces.

\subsection{Layerwise encoder similarity Analysis}\label{f7}

\begin{table}[!h]
\centering
\caption{\textbf{Encoders used for the layer-wise similarity analysis.}
The experiment uses 512 patches from four TCGA slides at $10\times$/256\,px
and evaluates intermediate representations using Linear CKA.}
\label{tab:layerwise_setup}

\small
\setlength{\tabcolsep}{10pt}
\renewcommand{\arraystretch}{1.08}

\begin{tabular}{@{}lc@{}}
\toprule
\textbf{Encoder} & \textbf{Architecture depth} \\
\midrule
UNI2-H        & 24 blocks \\
Prov-GigaPath & 40 blocks \\
CONCH         & 12 blocks \\
CTransPath    & 12 blocks \\
ResNet50      & 3 stages \\
\bottomrule
\end{tabular}
\end{table}

\begin{table*}[!h]
\centering
\caption{\textbf{Layer-wise cross-encoder similarity trajectories.}
For each layer of encoder A, the most similar layer in encoder B is identified
using Linear CKA. Early and late denote the average similarity near the
beginning and end of the resulting trajectory, respectively, with
$\Delta=\text{late}-\text{early}$. Positive $\Delta$ indicates increasing
cross-model similarity with depth. Divergence reports the relative depth at
which similarity falls below the CKA threshold of 0.5 and remains below it;
1.000 indicates no sustained terminal divergence.}
\label{tab:layerwise_cross_encoder}

\scriptsize
\setlength{\tabcolsep}{4.5pt}
\renewcommand{\arraystretch}{1.08}

\begin{tabular}{@{}lrrrrrrrr@{}}
\toprule
& \multicolumn{4}{c}{\textbf{CLS / First-token}}
& \multicolumn{4}{c}{\textbf{Mean pooling}} \\
\cmidrule(lr){2-5}
\cmidrule(l){6-9}

\textbf{A $\rightarrow$ B}
& \textbf{Early}
& \textbf{Late}
& \(\boldsymbol{\Delta}\)
& \textbf{Div.}
& \textbf{Early}
& \textbf{Late}
& \(\boldsymbol{\Delta}\)
& \textbf{Div.} \\
\midrule

UNI2-H $\rightarrow$ GigaPath
& 0.816 & 0.932 & +0.116 & 1.000
& 0.850 & 0.817 & -0.033 & 1.000 \\

UNI2-H $\rightarrow$ CONCH
& 0.836 & 0.946 & +0.110 & 1.000
& 0.967 & 0.866 & -0.101 & 1.000 \\

UNI2-H $\rightarrow$ CTransPath
& 0.438 & 0.428 & -0.010 & 0.000
& 0.927 & 0.886 & -0.041 & 1.000 \\

UNI2-H $\rightarrow$ ResNet50
& 0.623 & 0.737 & +0.114 & 1.000
& 0.939 & 0.802 & -0.137 & 1.000 \\

CONCH $\rightarrow$ GigaPath
& 0.809 & 0.956 & +0.147 & 1.000
& 0.840 & 0.800 & -0.040 & 1.000 \\

CONCH $\rightarrow$ CTransPath
& 0.497 & 0.444 & -0.053 & 0.455
& 0.956 & 0.902 & -0.054 & 1.000 \\

CONCH $\rightarrow$ ResNet50
& 0.818 & 0.759 & -0.059 & 1.000
& 0.966 & 0.896 & -0.070 & 1.000 \\

CTransPath $\rightarrow$ GigaPath
& 0.405 & 0.388 & -0.017 & 0.273
& 0.896 & 0.926 & +0.030 & 1.000 \\

CTransPath $\rightarrow$ ResNet50
& 0.421 & 0.424 & +0.003 & 0.273
& 0.898 & 0.921 & +0.022 & 1.000 \\

GigaPath $\rightarrow$ ResNet50
& 0.596 & 0.787 & \textbf{+0.191} & 1.000
& 0.754 & 0.852 & \textbf{+0.098} & 1.000 \\

\bottomrule
\end{tabular}
\end{table*}

\begin{table}[!h]
\centering
\caption{\textbf{Within-encoder representational change across depth.}
Linear CKA compares each encoder's final representation with its first
block or stage. Lower similarity indicates greater representational change
through the network.}
\label{tab:within_encoder_depth}

\small
\setlength{\tabcolsep}{6pt}
\renewcommand{\arraystretch}{1.08}

\begin{tabular}{@{}lcrr@{}}
\toprule
\textbf{Encoder}
& \textbf{Depth}
& \textbf{CLS / First-token}
& \textbf{Mean pooling} \\
\midrule
UNI2-H        & 24 & 0.359 & 0.498 \\
Prov-GigaPath & 40 & \textbf{0.168} & 0.780 \\
CONCH         & 12 & 0.467 & 0.602 \\
CTransPath    & 12 & 0.655 & 0.921 \\
ResNet50      & 3  & 0.872 & 0.872 \\
\bottomrule
\end{tabular}
\end{table}

\paragraph{Layer-wise representational similarity.}
The layer-wise analysis reveals that conclusions about cross-model
convergence depend strongly on how token representations are summarized.
Under CLS/first-token pooling, several pathology transformers become more
similar with depth: UNI2-H--GigaPath increases from 0.816 to 0.932,
UNI2-H--CONCH from 0.836 to 0.946, and CONCH--GigaPath from 0.809 to
0.956. In contrast, trajectories involving CTransPath remain substantially
weaker under this view, with UNI2-H--CTransPath remaining below the
0.5 CKA threshold throughout. Overall, four of ten CLS/first-token
trajectories exhibit terminal divergence, while the mean similarity
increases modestly from 0.626 to 0.680. The pattern changes markedly under mean pooling. Mean-pooled
representations exhibit substantially higher cross-model correspondence,
with average CKA remaining high from early (0.899) to late (0.867) depth
and no encoder pair showing sustained terminal divergence. This indicates
that apparent layer-wise convergence or divergence is sensitive to the
token summary used to represent each layer, rather than being a universal
property of the architectures themselves. Within individual encoders,
GigaPath undergoes the largest representational transformation across
depth under CLS pooling (first--final CKA = 0.168), followed by UNI2-H
(0.359) and CONCH (0.467), whereas CTransPath and ResNet50 retain more
similar representations from early to late stages.

\paragraph{CTransPath caveat.}
CTransPath is a Swin-based model without a semantic CLS token; consequently,
the CLS-mode analysis uses the first token of its token sequence and should
be interpreted as a \emph{first-token} trajectory rather than a true
CLS-token trajectory. Mean pooling therefore provides the more directly
comparable token-level summary when interpreting CTransPath relative to
the transformer encoders with explicit CLS representations.

%%%%%%%%%%%%%%%%%%%%%%%%%%%%%%%%%%%%%%%%%%%%%%%
% Bibliography
%%%%%%%%%%%%%%%%%%%%%%%%%%%%%%%%%%%%%%%%%%%%%%%%%%%%%%%%%%%%%%%%%%%%%%%%%%%%%%

\twocolumn
{
    \small
    \bibliographystyle{ieeenat_fullname}
    \bibliography{main}
}

%%%%%%%%%%%%%%%%%%%%%%%%%%%%%%%%%%%%%%%%%%%%%%%%%%%%%%%%%%%%%%%%%%%%%%%%%%%%%%